\documentclass{imsd2026_preprint}

\usepackage{cleveref}   

\usepackage{soul}

\begin{document}

\begin{center}
  \fontsize{12}{16}{\bf
    Learning Forced Multibody Dynamics on Lie Groups
    }
\end{center}

\begin{center}
\normalsize{
  \bf{
    M. Dyring Hansen$^{*\dag}$
    M. Ghirardelli$^\dag$,
    E. Celledoni$^\dag$,
    D. M. de Diego$^\#$,
    B. Owren$^\dag$}
}
\end{center}

\begin{center}
  \begin{tabular}{c}
    $^*$ Department of Mathematics and Cybernetics, SINTEF Digital               \\
    Forskningsveien 1, 0373 Oslo, Norway \\
    martine.dyring.hansen@sintef.no \\
  \end{tabular}

  \vspace{0.5ex}

  \begin{tabular}{c}
    $^\dag$ Department of Mathematical Sciences \\
    Norwegian University of Science and Technology (NTNU) \\
    Alfred Getz' vei 1, 7034 Trondheim, Norway  \\
    {[marta.ghirardelli, elena.celledoni, brynjulf.owren]@ntnu.no} \\
  \end{tabular}
  
  \vspace{0.5ex} 

  \begin{tabular}{c}
    $^\#$ Instituto de Ciencias Matematicas (CSIC) \\
    Calle Nicol\'as Cabrera 13-15, Cantoblanco, 28048 Madrid, Spain  \\
    david.martin@icmat.es     \\	
  \end{tabular} \\

\end{center}

\section*{ABSTRACT}
We introduce an architecture for learning the dynamics of forced
mechanical systems evolving on Lie groups, based on
discrete forced Euler-Lagrange equations. The proposed
method, termed Lie Group Discrete Forced
Euler-Lagrange Neural Network (LieDFLNN), learns
mechanical dynamics directly from discrete configuration
observations, without requiring velocity measurements.
This formulation allows the simultaneous identification
of unknown mechanical properties and forcing terms,
while ensuring that the predicted configurations remain
on the underlying Lie group. We develop explicit
formulations for rigid body dynamics on $SO(3)$ and
$SE(3)$, extend the framework to interacting multibody
systems, and introduce a method for learning
unknown control forces when the Lagrangian is known.
Finally, we evaluate the method through numerical experiments
involving rigid body dynamics, nonlinear external forces,
controlled systems, and real-world human motion capture
data. The numerical results confirm the advantages of incorporating
Lie group geometry and discrete variational structure
into data-driven modeling of forced mechanical systems.

\section{INTRODUCTION}
\paragraph{Motivation}
Many problems in learning dynamical systems from observational data involve states evolving on nonlinear configuration spaces. In particular, rotations and rigid body motions are naturally represented by Lie groups such as $SO(3)$, $SE(3)$ or product manifolds thereof \cite{barfoot2024state, bullo2005geometric, Holm-Schamah-Stoica, lee2013nonlinear, lee2018global, marsden1997introduction,  murray2017mathematical, selig2005geometric}. These representations are fundamental in a wide range of applications, including the attitude dynamics of spacecraft and drones, the pose of mobile robots, the motion of cameras and objects in computer vision, and the articulated motion of multibody systems \cite{barrau2016invariant, mahony2008nonlinear, zhou2019continuity}. In these settings, the available data typically consist of discrete observations that inherently satisfy geometric constraints, such as orthogonality and unit determinant. Encoding this geometry in the learning process yields models that respect the configuration manifold and fit the data better. 
As an example, rigid body orientations are in practice often parameterized by Euler angles. However, Euler angles provide only local coordinates on $SO(3)$: the map between angles
and rotations is a diffeomorphism only on an open subset of $SO(3)$ and breaks down at
coordinate singularities such as gimbal lock \cite{hemingway2018perspectives}. As a result, smooth rotational motions can appear discontinuous in Euler-angle coordinates, with abrupt jumps whenever the trajectory crosses a singularity. In contrast, matrix representations in $SO(3)$ remain globally well-defined and vary smoothly along any smooth trajectory on the group.

\paragraph{Related works}
Many existing approaches for learning dynamical systems from data assume that the state evolves in a Euclidean space. Even when the natural configuration manifold is a Lie group---such as rigid bodies or pendula on $S^1$---it is common to work with Euclidean representations rather than treating $SO(3)$ and $SE(3)$ as configuration manifolds \cite{cranmer2020lagrangian, greydanus2019hamiltonian, lutter2019deep, tong2021symplectic}. Such representations are of two kinds. \emph{Intrinsic} descriptions use local coordinate charts, e.g.\ Euler angles, in which the state has the dimension of the manifold. \emph{Extrinsic} descriptions embed the manifold in a Euclidean space, e.g.\ rotation matrices as vectors in $\mathbb R^9$ or unit quaternions in $\mathbb R^4$ (a double cover of $SO(3)$). These approaches typically enforce physical consistency through Hamiltonian or Lagrangian formalisms, often prioritizing the preservation of symplecticity while neglecting the intrinsic geometry of the configuration manifold. In many applications, however, sensor data are already group-valued (e.g.\ rotation matrices or poses estimated by SLAM/visual odometry) \cite{cadena2017past, engel2017direct, mur2015orb}. Both choices have drawbacks when applied to group-valued data. A single chart cannot cover $SO(3)$ without coordinate singularities (e.g.\ gimbal lock) and discontinuities, so one must either restrict to a region of the manifold or switch between charts, and representations of dimension four or less are discontinuous \cite{hemingway2018perspectives, zhou2019continuity}. In an extrinsic description, the coordinates are global, but a generic Euclidean model does not constrain its predictions to the embedded manifold: it can lose orthogonality or drift off the manifold, and often requires additional projection steps or specialized parameterizations to maintain consistency with the true configuration space \cite{barfoot2024state, kuffner2004effective, marsden1999introduction, walker1991estimating}.

A broader effort in geometric deep learning incorporates group and manifold structure into learning algorithms \cite{bonnabel2013stochastic, bronstein2021geometric, celledoni2021structure, offen2022symplectic}.
For mechanical systems, this has been done on Lie groups in continuous time by Hamiltonian and port-Hamiltonian neural ODEs on $SE(3)$ and on matrix Lie groups
\cite{duong2021hamiltonian, duong2024port}. These embed the group in a Euclidean space of matrices, parameterize the mass matrix, the potential energy and the input matrix (and, in \cite{duong2024port}, a dissipation term) by neural networks, and are trained by integrating the learned vector field between observation times. The training data include velocities, and the group constraint is satisfied only up to the accuracy of the ODE solver.

A different route, which we follow, is to learn discrete mechanics
\cite{marsden2001discrete}: a discrete Lagrangian on pairs of configurations replaces the continuous one, so that only positions are needed. Discrete Lagrangians have been learned from position data in Euclidean spaces
\cite{ober2023variational, offen2025machine, hansen2025learning, li2026learning}, and discrete variational formulations on Lie groups have been used for learning controlled systems on the rototranslation group in \cite{duruisseaux2023lie}.

\paragraph{Our contribution}
We propose LieDFLNN, a method for learning forced mechanical systems evolving on Lie groups from position data only. It combines two existing ingredients. The first is learning a discrete Lagrangian instead of a continuous one, which removes the need for velocities and hence for numerical differentiation of noisy positions \cite{ober2023variational, hansen2025learning, li2026learning}. This is in contrast to most Hamiltonian- and Lagrangian-based learning approaches, including those on Lie groups \cite{duong2021hamiltonian, duong2024port}, which build on the continuous formulation and typically require velocities, accelerations or momenta as training data \cite{cranmer2020lagrangian, greydanus2019hamiltonian, lutter2019deep}. The second ingredient is discrete variational mechanics on Lie groups \cite{lee2013nonlinear, leok2004foundations, marsden2001discrete}. The closest work is \cite{duruisseaux2023lie}, which considers controlled systems on the rototranslation group but not multibody systems.

The contributions specific to this paper are the following. (i)~We derive the trivialized discrete forced Euler-Lagrange equations on a general Lie group and give them explicitly for $SO(3)$, $SE(3)$ and their products (multibody systems), including external forces and control inputs (\Cref{sec:DFEL Lie group,sec:DFEL SO3 and SE3}). (ii)~We parameterize the discrete Lagrangian and forces so that physical structure is built into the model (e.g.\ positive definite inertia, dissipative damping) and add a penalty promoting regularity of the learned Lagrangian (\Cref{section: Lie group DFLNN}). (iii)~The framework is modular: it identifies mechanical properties and forces jointly, which is relevant in robotics, e.g.\ for learning mass matrices and friction forces from real data \cite{li2026learning}, or a single component such as an unknown control law when the Lagrangian is known. (iv)~We validate it on synthetic rigid bodies, on a learned controller, and on real motion-capture data of a multibody system, with baselines that isolate the choice of coordinates, of discretization, and of manifold constraint (\Cref{sec:experiments LieDFLNN}).

At each step the unknown increment is an element of the Lie algebra mapped to the group by the exponential map, so the predicted configurations remain on the manifold by construction (up to round-off) and no re-orthogonalization of the state is required. Local charts such as Euler angles, with their singularities and discontinuities \cite{hemingway2018perspectives, zhou2019continuity}, are avoided.

The paper is structured as follows. \Cref{sec:DFEL Lie group} derives the discrete forced Euler-Lagrange equations on Lie groups, which are then specialized to $SO(3)$, $SE(3)$ and their product manifolds in \Cref{sec:DFEL SO3 and SE3}. \Cref{section: Lie group DFLNN} details the model architecture, including the loss function and regularization strategies. Finally, \Cref{sec:experiments LieDFLNN} presents a series of experiments validating the proposed approach.

\section{DISCRETE FORCED EULER-LAGRANGE EQUATIONS ON LIE GROUPS}
\label{sec:DFEL Lie group}
In this section, we first recall the continuous forced Euler-Lagrange equations on Lie groups. We then introduce discrete approximations of both the Lagrangian and the external forces and derive the corresponding discrete forced Euler-Lagrange equations on Lie groups (see \cite{abraham, Holm-Schamah-Stoica, marsden1997introduction}). Importantly, the resulting discretization is expressed entirely in terms of Lie group elements (positions), and does not require explicit velocities in the Lie algebra.

\subsection{Trivialized forced Euler-Lagrange equations on Lie groups}

Let \(Q\) be a differentiable manifold  with $\dim Q=n$ with local coordinates $(q^i)$, $i=1,\dots,n$, and   \((q^i,\dot{q}^i)\) the corresponding induced coordinates on its tangent bundle \(TQ\) with canonical projection $\tau_Q: TQ\rightarrow Q$. Denote also by $T^*Q$ its cotangent bundle with induced coordinates $(q^i, p_i)$ and canonical projection $\pi_Q: T^*Q\rightarrow Q$.

Given a  Lagrangian function \(L\colon  TQ\to{\mathbb R}\), and  an external force \( F\colon  TQ\to T^*Q\) (a fibered map over \(Q\), that is, $\pi_Q\circ F=\tau_Q$). The Euler-Lagrange equations for the system \((L,F)\) are
\begin{equation}
    \label{eq:EL:group:continuous}
    \frac{d}{dt}\left(\frac{\partial L}{\partial \dot{q}}\right) - \frac{\partial L}{\partial q} = F.
\end{equation}

In this paper, we are interested in the case when $Q=G$ is a Lie group. In this case, we   can alternatively represent the Euler-Lagrange equations using  left or right trivialization, \(TG\cong G\times\mathfrak{g}\) where \( \mathfrak{g}=T_eG\) is the Lie algebra of \(G\). Let $g,e\in G$, $e$ being the identity element. We denote $\mathcal{L}_g : G \to G$ the left multiplication by $g$, $ T_e \mathcal{L}_g : \mathfrak{g} \to T_g G$ its tangent lift at the identity, and $\mathcal{L}_g^* : T^*_g G \to \mathfrak{g}^* $ its cotangent lift. We can then rewrite \(L\colon TG\to {\mathbb R}\) and \(F\colon  TG\to T^*G\), using left-trivialization as  \(\bar{L}\colon  G\times\mathfrak{g}\to {\mathbb R}\) and $\bar{F}\colon G\times\mathfrak{g}\to\mathfrak{g}^*$ by
\begin{equation*}
    \bar {L}(g,\xi) = L(g, g\xi), \qquad
    \bar{F}(g,\xi) ={\mathcal L}_g^*\big(F(g,g\xi)\big),
\end{equation*}
where $\dot g=T_e {\mathcal L}_g\xi=g\xi$ and $\langle {\mathcal L}_g^*\big(F(g,g\xi)\big), \eta\rangle=\langle F(g, g\xi), T_e{\mathcal L}_g (\eta)\rangle$ for $\eta\in {\mathfrak g}$.

The left-trivialized Euler-Lagrange equations for \((\bar{L},\bar{F})\) are
\begin{equation}\label{eq:EL:group:}
    \frac{d}{dt}\left(\frac{\partial \bar{L}}{\partial \xi}\right)
    - \ad_{\xi}^*\left(\frac{\partial \bar{L}}{\partial \xi}\right)
    - {\mathcal L}_g^*\left(\frac{\partial \bar{L}}{\partial g}\right) = \bar{F} ,
\end{equation}
which, together with the \emph{reconstruction equation} \(\dot{g}=g\xi\), are equivalent to Eq. \eqref{eq:EL:group:continuous}. Here $\ad_{\xi}^*: {\mathfrak g}^*\rightarrow {\mathfrak g}^*$ denotes  the dual map of the adjoint operator $\ad_{\xi}(\eta)=[\xi, \eta]$, where $[\; ,\;]$ is the Lie bracket of the Lie algebra ${\mathfrak g}$.

An interesting case is when the Lagrangian $\bar{L}$ only depends on elements of ${\mathfrak g}$ (it is left-invariant), that is, $\bar{L}(g, \xi)=l(\xi)$ for all $g\in G$ and $\xi\in {\mathfrak g}$. In this case, Equations \eqref{eq:EL:group:} are known as the \emph{forced Euler-Poincar\'e equations}: 

\begin{equation}\label{eq:ELP:group}
    \frac{d}{dt}\left(\frac{\partial l}{\partial \xi}\right)
    - \ad_{\xi}^*\left(\frac{\partial l}{\partial \xi}\right)
     = \bar{F} .
\end{equation}

\subsection{Discrete forced Euler-Lagrange equations on Lie groups}
Since we aim at using only positions, we replace the velocity phase space \(TQ\)  by \(Q\times Q\) \cite{marsden2001discrete}. 
Consider a discrete Lagrangian $L_d: Q\times Q\rightarrow {\mathbb R}$  and two families of discrete external forces \(F^\pm_k\colon Q\times Q\to T^*Q\) satisfying
\[
F^-_k (q_k, q_{k+1})\in T^*_{q_k}Q,\qquad 
F^+_k (q_k, q_{k+1})\in T^*_{q_{k+1}}Q.
\]
Then the discrete forced Euler-Lagrange equations are
\begin{equation}
    \label{eq:DEL}
    D_1 L_d(q_k,q_{k+1}) + D_2 L_d(q_{k-1},q_k) + F^-_k(q_k,q_{k+1}) + F^+_{k-1}(q_{k-1},q_k) = 0. 
\end{equation}
 Given two initial points \((q_0,q_1)\), Eq. \eqref{eq:DEL} determines iteratively \(q_{k+1}\) from the two previous points \((q_{k-1},q_k)\) for \(k\geq 1\) under suitable regularity conditions \cite{marsden2001discrete}.\\
In the particular case when \(Q\) is a Lie group \(G\), in a similar way to the continuous case, we can consider trivialized expressions. That is, instead of working with pairs \((g_k,g_{k+1})\) of consecutive points in a trajectory, we work with pairs of the form \emph{source-arrow} \((g_k,W_k)\) pointing towards a \emph{target} \(g_kW_k=g_{k+1}\). Hence $W_{k}=g_{k}^{-1}g_{k+1}$. We define the ``trivialized'' discrete Lagrangian and forces as follows
\begin{align*}
    \bar{L}_d(g,W) =& L_d(g,gW) ,
    \qquad {\bar F}^-_k(g,W)= {\mathcal L}_{g}^*F^-_k(g,gW) ,
    \qquad {\bar F}^+_k(g,W) ={\mathcal L}_{gW}^*F^+_k(g,gW).
\end{align*}
Observe that $\bar{F}^-_k(g,W)$ and $ {\bar F}^+_k(g,W)$ are elements of ${\mathfrak g}^*$, that is, ${\bar F}^{\pm}_k: G\times G\rightarrow {\mathfrak g}^*$. \\
The discrete Euler-Lagrange equations are now rewritten as 
\begin{align}\label{eq:DEL2}
    {\mathcal L}_{g_k}^*
    \frac{\partial \bar{L}_d}{\partial g}(g_k, W_k) - {\mathcal R}_{W_k}^*\frac{\partial \bar{L}_d}{\partial W}(g_k, W_k) & + {\mathcal L}_{W_{k-1}}^*\frac{\partial \bar{L}_d}{\partial W}(g_{k-1}, W_{k-1}) \nonumber \\ 
    &+ {\bar F}^-_k (g_{k}, W_{k}) + {\bar F}^+_{k-1}(g_{k-1}, W_{k-1}) = 0,
\end{align}
with the reconstruction equation \(g_{k+1}=g_kW_k\). Here, $T{\mathcal L}_{g}$ and $T{\mathcal R}_{g}$ are the tangent lifts at the identity of the left and right multiplication by $g$ in $G$, and ${\mathcal L}_{g}^*$ and ${\mathcal R}_{g}^*$ denote the corresponding dual maps.

\section{DISCRETE FORCED EULER-LAGRANGE EQUATIONS ON $SO(3)$ AND $SE(3)$}
\label{sec:DFEL SO3 and SE3}
In this section, we derive the previous discrete forced Euler-Lagrange equations on two specific Lie groups, $SO(3)$ and $SE(3)$, and use rigid body dynamics in both spaces as concrete examples. We then show how the approach generalizes to multibody systems and how control inputs can be incorporated. In all cases, the discrete approximations of the Lagrangian and the external forces are obtained using a mid-point approximation scheme.

\subsection{Discrete forced Euler-Lagrange equations on $SO(3)$.}
Consider the 3D rotation group $SO(3) = \{ R\in\mathbb{R}^{3\times 3} : R^\top R = R R^\top = I, \, \det(R) =1\} $. Its Lie algebra $\mathfrak{so}(3)=\{\Omega \in \mathbb{R}^{3\times 3} : \Omega^\top = -\Omega  \}$ can be identified with $\mathbb{R}^3$ via the hat map 
\begin{equation*}
    \hat{}:\mathbb{R}^3 \to \mathfrak{so}(3),\qquad
    \widehat{(a,b,c)}=\begin{pmatrix}
    0&-c&b\\
    c&0&-a\\
    -b&a&0
    \end{pmatrix},
\end{equation*}
and we let $\vex: \mathfrak{so}(3) \to \mathbb{R}^3 $ be its inverse. 
Consider a Lagrangian $\bar{L} : SO(3) \times \mathfrak{so}(3) \to \mathbb{R}$, $\bar{L} = \bar{L}(R,\omega)$. The reconstruction equation is $\dot{R}=R\hat{\omega}$ for $\omega \in \mathbb{R}^3$, hence $\hat{\omega} = R^{-1} \dot{R}$ and we consider the approximation (see Moser and Veselov \cite{moser91dvo})
\begin{eqnarray*}
    \hat{\omega}_k = R_k^{-1}\dot{R}_k=R_k^\top\dot{R}_k\approx R_k^\top\left( \frac{R_{k+1}-R_k}{h}\right)
    = \frac{R_k^\top R_{k+1}-I}{h}= \frac{W_k-I}{h}, 
\end{eqnarray*}
with $W_k=R_k^{-1}R_{k+1}=R_k^\top R_{k+1} \in SO(3)$ and stepsize $h>0$. Consider the   discrete Lagrangian $\bar{L}_d : SO(3) \times SO(3) \to \mathbb{R}$:
\begin{align*}
    \bar{L}_{d,k} =\bar{L}_{d} (R_k,W_k) 
    = h \bar{L} \left(R_k, \frac{W_k - I}{h} \right). 
\end{align*}
\begin{remark}
Observe that, following \cite{moser91dvo}, we have used the
first-order approximation
\[
\hat{\omega}_k \approx \frac{W_k-I}{h},
\]
where $(W_k-I)/h$ belongs to $\mathfrak{gl}(3,\mathbb{R})$,
the space of all $3\times 3$ real matrices.
The reason is that in the definition defining $\bar{L}_{d,k}$, we have assumed that the Lagrangian
$\bar{L}$ is defined on the entire space
$\mathfrak{gl}(3,\mathbb{R})$, rather than being restricted
to $\mathfrak{so}(3)$. This assumption is satisfied in many
standard examples. In more general situations, however,
retraction maps \cite{BouMa} provide a natural
alternative for constructing an appropriate discretization
of the Lagrangian. In practice, we parametrize the matrix $W_k$ with an element of $\mathfrak{so}(3) \simeq \mathbb{R}^3$, which is mapped to the group $SO(3)$ via the exponential map (see \Cref{subsection: inference routine} and \Cref{appendix:basic notions about SO3 and SE3}).
\end{remark}
Let $\tau_\mathrm{ext}:  SO(3) \times \mathfrak{so}(3) \to \mathbb{R}^3\equiv \mathfrak{so}^*(3)$ be an external torque, a discretization is (see \Cref{remark: approximation of log map in SE3})
\begin{equation*}
    {\tau_{\mathrm{ext}}}_{d,k}^\pm =
    \frac{h}{2} \tau_{\mathrm{ext}} \left(R_k, \mathrm{skew} \left(\frac{W_k - I}{h} \right) \right) =
    \frac{h}{2} \tau_{\mathrm{ext}} \left(R_k, \frac{W_k - W_k^\top}{2h} \right).
\end{equation*}
The discrete forced Euler-Lagrange equations \eqref{eq:DEL2} rewrite (see \Cref{appendix: DEL SO3} and \cite{moser91dvo})
\begin{equation}
    \label{eq: DEL SO3 version with A}
    A - A^\top =0
\end{equation}
where 
\begin{equation*}
    A:= R_k^\top \frac{\partial \bar{L}_{d,k}}{\partial R_k} -  \frac{\partial \bar{L}_{d,k}}{\partial W_k} W_k + W_{k-1} \frac{\partial \bar{L}_{d,k}}{\partial W_{k-1}} 
    + \frac{1}{2} (\widehat{{\tau_{\mathrm{ext}}}^-_{d,k}}+ \widehat{{\tau_{\mathrm{ext}}}^+_{d,k-1}}).
\end{equation*}
Given $R_k$ and $W_{k-1}$, solving \eqref{eq: DEL SO3 version with A} gives $W_k$. Then $R_{k+1}=R_k W_k$. 

The above formulation is rather general. The derivatives in \eqref{eq: DEL SO3 version with A} can be written out explicitly for specific Lagrangians, or computed with automatic differentiation. Imposing more structure on the Lagrangian, as in the following example, removes the need for automatic differentiation. This lowers the computational cost and simplifies the learning process.

\begin{example}[Rigid body]
    \label{example: L=T-V in SO3}
    Let us consider the Lagrangian of a rigid body
    \begin{equation*}
        \bar{L}(R,\omega) = T(\omega) - V(R) = \frac{1}{2} \omega^\top J \omega - V(R) \nonumber
    \end{equation*}
    where $J$ is the inertia matrix for vector representation. Then 
    \begin{align*}
        \bar{L}_{d,k} = \bar{L}_{d} (R_k,W_k) = \frac{1}{h}(\trace(J_d)-\trace(W_k J_d)) - h V(R_k),
    \end{align*}
    where $J_d$ is the inertia matrix for matrix representation satisfying $J = \trace({J_d}) I - J_d$, as noted in \cite{lee2005lie}. Equation \eqref{eq: DEL SO3 version with A} then rewrites
    \begin{equation}
    \label{eq: discrete EL eq in SO3}
        - h  \left( R_k^\top  \frac{\partial V_{k}}{\partial R_k} - \frac{\partial V_{k}}{\partial R_k}^\top R_k \right)
        + \frac{1}{h} \left( J_d W_k^\top + J_d W_{k-1} \right) - \frac{1}{h} \left( W_k J_d +  W_{k-1}^\top J_d \right) +  (\widehat{{\tau_{\mathrm{ext}}}^-_{d,k}}+ \widehat{{\tau_{\mathrm{ext}}}^+_{d,k-1}}) = 0
    \end{equation}
\end{example}

\subsection{Discrete forced Euler-Lagrange equations on $SE(3)$.}
Consider the group of rigid transformations  $SE(3) \simeq SO(3) \times \mathbb{R}^3$ and its Lie algebra $\mathfrak{se}(3) \simeq \mathbb{R}^3 \times \mathbb{R}^3$. For $g\in SE(3)$ and $\xi\in \mathfrak{se}(3)$ we have
\[ 
g=\begin{bmatrix}
    R & p \\ 0 & 1
\end{bmatrix}, \qquad\xi=\begin{bmatrix}
    \hat{\omega} & v \\ 0 & 0
\end{bmatrix}, \qquad R\in SO(3), \, \hat{\omega} \in \mathfrak{se}(3), \, p, v\in\mathbb{R}^3.
\]
Consider a Lagrangian $ \bar{L} : SE(3) \times \mathfrak{se}(3) \to \mathbb{R}$, $\bar{L}=\bar{L}(R, p, \omega, v)$.
The reconstruction equation is $\Dot{g} = g\xi$, hence $\xi = g^{-1} \Dot{g} = (R^\top \Dot{R}, R^\top \dot{p})$ and we consider the approximations 
\begin{align*}
    \hat{\omega}_k &=  R^{\top}_k \dot{R}_k \approx R_k^\top \frac{R_{k+1} - R_k}{h} = \frac{W_k - I}{h}\\
    v_k &= R^{\top}_k \dot{p}_k \approx R_k^\top \frac{p_{k+1} - p_k}{h} = \frac{z_k}{h}
\end{align*}
with $W_k := R^{-1}_k R_{k+1} = R^{\top}_k R_{k+1} \in SO(3)$, $z_k := R_k^\top (p_{k+1} - p_k) \in \mathbb{R}^3$ and stepsize $h>0$. The discrete Lagrangian $\bar{L}_d : SE(3) \times SE(3) \to \mathbb{R}$ is then
\begin{equation*}
    \label{eq:general_discrete_lagrangian_with_potential}
    \bar{L}_{d,k} = \bar{L}_d (R_k, p_k, W_k, z_k) = \, h \bar{L} \left(R_k, p_k, \frac{W_k - I}{h}, \frac{z_k}{h} \right)  
\end{equation*}
Let $F: SE(3) \times \mathfrak{se}(3) \to \mathbb{R}^6\equiv \mathfrak{se}^*(3)$ be an external force, and let us split $F$ into external torque and external translational force, $F=[\tau_\mathrm{ext}, f_\mathrm{ext}] \in \mathbb{R}^3 \times \mathbb{R}^3$. For its discretization $( {\tau_{\mathrm{ext}}}_{d}^\pm, {f_{\mathrm{ext}}}_{d}^\pm )$, we consider the following approximation of the logarithmic map (see \Cref{remark: approximation of log map in SE3})
\begin{equation*}
    (\tilde{\omega}_k, \tilde{v}_k )
    = \left( \vex \left( \mathrm{skew} \left(\frac{W_k - I}{h} \right) \right), \left( I - \frac{1}{2} h \tilde{\omega}_k \right) z_k \right) \approx \frac{\log^{SE}(W_k, z_k)}{h},
\end{equation*}
so that
\begin{equation*}
    \left( {\tau_{\mathrm{ext}}}_{d,k}^\pm, {f_{\mathrm{ext}}}_{d,k}^\pm \right) 
    = \frac{h}{2}  \left( {\tau_{\mathrm{ext}}} \left(R_k, p_k, \tilde{\omega}_k, \tilde{v}_k \right), {f_{\mathrm{ext}}}\left(R_k, p_k, \tilde{\omega}_k, \tilde{v}_k \right) \right).
\end{equation*}
The discrete forced Euler-Lagrange equation \eqref{eq:DEL2} rewrites (see \Cref{appendix: DEL SE3})
\begin{align}
    A - A^\top & = 0  \label{eq: DEL SE3 version with A}\\
    b & = 0 \label{eq: DEL SE3 version with b}
\end{align}
where 
\begin{align*}
    A &:=  R_k^\top \frac{\partial \bar{L}_{d,k}}{\partial R_k} - \frac{\partial \bar{L}_{d,k}}{\partial W_k} W_k^\top + \frac{1}{2} \left( \frac{\partial \bar{L}_{d,k}}{\partial z_k} \times z_k \right)^\wedge + W_{k-1}^\top \frac{\partial \bar{L}_{d,k-1}}{\partial W_{k-1}} 
    + \frac{1}{2} (\widehat{{\tau_\mathrm{ext}}^-_{d,k}}+ \widehat{{\tau_\mathrm{ext}}^+_{d, k-1}}), \\
    b &:= R_k^\top \frac{\partial \bar{L}_{d,k}}{\partial p_k} - \frac{\partial \bar{L}_{d,k}}{\partial z_{k}} + W_{k-1}^\top \frac{\partial \bar{L}_{d,k-1}}{\partial z_{k-1}} 
    + ({f_\mathrm{ext}}^-_{d,k}+{f_\mathrm{ext}}^+_{d,k-1}).
\end{align*}
Given $R_k$, $p_k$, $W_{k-1}$ and $z_{k-1}$, solving \eqref{eq: DEL SE3 version with A}-\eqref{eq: DEL SE3 version with b} gives $W_k$ and $z_k$. Then $R_{k+1} = R_k W_k$ and $p_{k+1} = p_k + R_k z_k $. 

As in the $SO(3)$-case, imposing more structure on the Lagrangian may allow to replace (some of) the derivatives in \eqref{eq: DEL SE3 version with A}-\eqref{eq: DEL SE3 version with b} (see the rigid body \Cref{example: L=T-V in SE3}).

\subsection{Generalization to multibody systems}
\label{section: Generalization to multibody systems}
The previous framework extends naturally to the product Lie group consisting of finitely many copies of $SE(3)$ and $SO(3)$,
\begin{equation*}
    G = \underbrace{SE(3) \times\dots\times SE(3)}_m \times \underbrace{SO(3) \times\dots\times SO(3)}_n,
    \qquad m,n \ge 0,
\end{equation*}
which provides an appropriate configuration space for a multibody system. An element $g_k = (g_{1,k}, \dots, g_{m+n,k}) \in G$ encodes the states of each body at timestep $k$, while $\xi_k = (\xi_{1,k}, \dots, \xi_{m+n,k}) \in \mathfrak{g}$ encodes their corresponding velocities. 
The Lagrangian $\bar{L}:G\times\mathfrak{g} \to \mathbb{R}$ is given by
\begin{equation*}
    \bar{L}(g,\xi) = \sum_i^{m+n} \bar{L}_i(g_i,\xi_i) - V_{\mathrm{int}}(g,\xi)
\end{equation*}
where $\bar{L}_i(g_i,\xi_i) : SE(3) \times \mathfrak{se}(3) \to\mathbb{R}$ for $i=1,\dots,m$,  $\bar{L}_i(g_i,\xi_i) : SO(3) \times \mathfrak{so}(3) \to\mathbb{R}$ for $i=m+1,\dots,m+n$ are the individual Lagrangians for each body, and $V_\mathrm{int}:G\times\mathfrak{g}\to\mathbb{R}$ is the interaction potential (e.g. a spring-damper connection, a gravitational attraction between the bodies, or a penalty function if a joint is to be modeled in that way; explicit examples
are given in Appendix~\ref{appendix: examples of interaction potentials}). The discrete Lagrangian $\bar{L}_d : G \times G \to \mathbb{R}$ can be constructed as in the previous sections, along with the associated reconstruction equations and the inclusion of external forces. The discrete forced Euler-Lagrange equation \eqref{eq:DEL2} becomes a system of $2m$ equations of type \eqref{eq: DEL SE3 version with A}-\eqref{eq: DEL SE3 version with b} and $n$ equations of type \eqref{eq: DEL SO3 version with A}. Each body satisfies its own discrete Euler-Lagrange equation, but the equations are coupled through the partial derivatives of the interaction potential (see \Cref{appendix: DEL SE3 x SO3}).

\subsection{Generalization to controlled systems}
\label{section: Generalization to controlled systems}
The external forces described in the previous sections can also be incorporated into mechanical systems actuated by control forces or control inputs (see, for instance, \cite{bullo2005geometric, duruisseaux2023lie, lee2005lie, lee2013nonlinear} and references therein). Typically, these forces are described by a map
\[
F_c:TQ\times U\rightarrow T^*Q,
\]
where \(U\) denotes the set of admissible controls. Their discretization gives rise to the discrete control forces
\[
F_k^\pm:Q\times Q\times U\rightarrow T^*Q.
\]
In the Lie group setting, after left trivialization, we obtain
\[
\bar F_k^\pm:G\times G\times U\rightarrow\mathfrak g^*.
\]
The discrete controlled Euler-Lagrange equations then become
\begin{align}\label{eq:DEL2-control}
    {\mathcal L}_{g_k}^*
    \frac{\partial \bar{L}_d}{\partial g}(g_k, W_k)
    -{\mathcal R}_{W_k}^*
    \frac{\partial \bar{L}_d}{\partial W}(g_k, W_k)
    +{\mathcal L}_{W_{k-1}}^*
    \frac{\partial \bar{L}_d}{\partial W}(g_{k-1}, W_{k-1})
    \nonumber\\
    +\,{\bar F}^-_k(g_k,W_k,u_k)
    +{\bar F}^+_{k-1}(g_{k-1},W_{k-1},u_{k-1})
    =0,
\end{align}
where \(u_{k-1},u_k\in U\) are the discrete control inputs.

\begin{example}
    [Controlled system in $SO(3)$]
    \label{example: control SO3}
    Let us consider the rigid body system in \Cref{example: L=T-V in SO3}, with a control torque of the type
    $\bar{\tau}_c: SO(3) \times  SO(3) \times U \rightarrow {\mathbb R}^3\equiv \mathfrak{so}(3)^*$, $ U\subset {\mathbb R}^m$,
    given by 
    \[
    \bar{\tau}_c(R, W, u)= \bar{\tau}_c( u) = C u
    \]
    where $C\in {\mathbb R}^{3\times m}$.
    Then a simple discretization of this force is given by
    \[
    {\bar \tau^-_k}(u_k)=Cu_k\; ,\qquad {\bar \tau^+_k}(u_{k})=0
    \]
    and equation \eqref{eq:DEL2-control} rewrites
    \begin{equation}
        \label{eq:DEL2-4-bis}
        - h  \left( R_k^\top  \frac{\partial V_{k}}{\partial R_k} - \frac{\partial V_{k}}{\partial R_k}^\top R_k \right)
        + \frac{1}{h} \left( J_d W_k^\top + J_d W_{k-1} \right) - \frac{1}{h} \left( W_k J_d +  W_{k-1}^\top J_d \right) - h\widehat{Cu_k} = 0
    \end{equation}
    with $R_{k+1}=R_kW_k$.
\end{example}

\section{LIE GROUP DISCRETE FORCED EULER-LAGRANGE NEURAL NETWORK (LieDFLNN)}
\label{section: Lie group DFLNN}
The proposed method LieDFLNN learns the dynamics of an observed system by approximating the discrete Lagrangian and the discrete forces with neural networks
\begin{equation*}
    L^\theta\approx \bar{L}_d \quad L^\theta : G\times G \to \mathbb{R}, \qquad
    F^\theta\approx \bar F^\pm \quad F^\theta : G\times G \to T^*G
\end{equation*}
where $L^\theta$ and $F^\theta$ are parametrized by learnable parameters $\theta$.

\subsection{Loss function and regularization}
\paragraph{Loss function.}
During training, we minimize 
\begin{equation}
    \label{eq: loss general}
    \mathrm{loss} = \frac{\omega_\mathrm{DEL}}{|\mathcal{T}| (N-1)} \sum_\mathcal{T} \sum_{n=1}^N
     \mathrm{loss}_\mathrm{DEL}(L^\theta, F^\theta) (g_{n-1}, g_n, g_{n+1})
    + \frac{\omega_\mathrm{reg}}{N_R} \sum_{r=1}^{N_R}
     \mathrm{loss}_\mathrm{reg}(L^\theta) (g_{r}, g_{r+1}),
\end{equation}
where, $\mathrm{loss}_\mathrm{DEL}$ is the residual of the discrete forced Euler-Lagrange equations~\eqref{eq:DEL2}, and $\mathrm{loss}_\mathrm{reg}$ enforces regularity of $L^\theta$. 
The two terms are weighted by hyperparameters $\omega_\mathrm{DEL}$ and $\omega_\mathrm{reg}$, respectively. We consider a set $\mathcal{T}$ of $|\mathcal{T}|$ trajectories, each with $N$ points, and use $N_R$ pairs $(g_n, g_{n+1})$ for regularization. One may choose $N_R=N$ and regularize over the entire dataset, but in practice we find that $N_R<N$ is sufficient to obtain a regular Lagrangian. 

\paragraph{Evaluating the residual on data.}
For each observed triplet $(g_{n-1}, g_n, g_{n+1})$, the group increments are computed directly from the data, for example $(W_n, z_n) = g_n^{-1} g_{n+1}$ on $SE(3)$. They are inserted into \eqref{eq:DEL2} to compute the residual, evaluated with the learned $L^\theta$ and $F^\theta$. The explicit expressions of $\mathrm{loss}_\mathrm{DEL}(L^\theta, F^\theta)$ for $SO(3)$, $SE(3)$ and multibody systems are given in Appendix~\ref{appendix: detailed derivation of DEL}.

The proposed framework thus learns a discrete Lagrangian and forces directly from observed trajectories by minimizing the residual of the discrete forced Euler-Lagrange equations. However, minimizing the residual alone may lead to trivial or degenerate solutions---for instance, a nearly constant Lagrangian whose derivatives vanish. The regularization term $\mathrm{loss}_\mathrm{reg}$ is therefore crucial to ensure that the learned Lagrangian remains regular and avoid such pathological cases. 

\paragraph{Regularity of $L$.}
A continuous Lagrangian $\bar L:G\times\mathfrak g\to\mathbb R$ is regular if its Hessian with respect to the left-trivialized velocity variables is non-singular at every point \cite{lee2005lie, offen2025machine}. This is the second derivative of $\bar L(g,\cdot)$ on the vector space $\mathfrak g$, hence a well-defined symmetric bilinear form at every $g\in G$. In a basis $\{e_i\}_{i=1}^m$ of $\mathfrak g$, where $m$ is the dimension of the Lie group $G$, $\bar L$ is regular if
\begin{equation*}
    \left[ \frac{\partial^2 \bar L}{\partial \xi^i \partial \xi^j} (g,\xi) \right]_{i,j=1}^m
\end{equation*}
is invertible for all fixed $(g,\xi)\in G\times\mathfrak g$.

In our discrete setting, $\bar L$ is replaced by a position-only discrete Lagrangian $L_d:G\times G \to \mathbb{R}$, in which the role of the velocity is played by the second
configuration $f$ (e.g.\ the next configuration in time). Following
\cite{marsden2001discrete}, $L_d$ is regular if its mixed second derivative $D_1D_2L_d(g,f)$
is non-degenerate at every $(g,f)\in G\times G$. This is a bilinear form
$T_gG\times T_fG\to\mathbb R$ obtained by differentiating once with respect to each
argument, so it is well defined without a connection or a metric. 
Intrinsically, the regularity condition can be characterized
in terms of the discrete Legendre transformations.
More precisely, $L_d$ is regular if and only if the
discrete Legendre transformation
\[
\begin{aligned}
\mathbb{F}^{-}L_d:G\times G&\longrightarrow T^*G,\\
(g,f)&\longmapsto \bigl(g,-D_1L_d(g,f)\bigr)
\end{aligned}
\]
is a local diffeomorphism. Equivalently, regularity
holds if and only if the discrete Legendre transformation
\[
\begin{aligned}
\mathbb{F}^{+}L_d:G\times G&\longrightarrow T^*G,\\
(g,f)&\longmapsto \bigl(f,D_2L_d(g,f)\bigr)
\end{aligned}
\]
is a local diffeomorphism. Alternatively, the regularity condition is equivalent to
the non-degeneracy of the discrete Poincar\'e-Cartan
2-form
\[
\omega_{L_d}
=
(\mathbb{F}^-L_d)^*\omega_G
=
(\mathbb{F}^+L_d)^*\omega_G,
\]
where $\omega_G$ denotes the canonical symplectic
form on $T^*G$. Since $\omega_{L_d}$ is closed,
regularity is therefore equivalent to
$\omega_{L_d}$ being a symplectic form on $G\times G$ (see \cite{marsden2001discrete}).
Using the left
trivializations $T_gG\cong T_fG\cong\mathfrak g$, the condition of regularity is  expressed,  for a given basis the basis
$\{e_i\}$, by the regularty of the matrix:
\begin{equation*}
    \mathbb{H}(L_d)(g,f)_{ij} = \left.\frac{\partial^2}{\partial s\,\partial t}\,
    L_d\big(g\exp^G (s e_i),\, f\exp^G (t e_j)\big)\right|_{s=t=0},
\end{equation*}
where $\exp^G: \mathfrak{g} \to G$ is the exponential map. In general, $\mathbb H$ is not symmetric, since its two arguments live in different tangent spaces. Non-degeneracy of $\mathbb H$ does not depend on the choice of basis, but the value of its determinant does: under a change of basis $P$, $\det(P^\top\mathbb H P)=\det(P)^2\det\mathbb H$. To give a meaning to a threshold, we fix an inner product on $\mathfrak g$ and take $\{e_i\}$ orthonormal (the Euclidean inner product on $\mathbb R^3$ for $\mathfrak{so}(3)$ via the hat map, and on $\mathbb R^6$ for $\mathfrak{se}(3)$). Equivalently, we raise one index of $\mathbb H$ with this inner product and consider the resulting endomorphism of $\mathfrak g$, whose determinant is invariant under orthonormal changes of basis.
Regularity then reads $\det\mathbb H(L_d)(g,f)\neq 0$. 

For the learned Lagrangian, which approximates $\bar L_d$, we write $L_d^\theta(g,f):=L^\theta(g,g^{-1}f)$, so that $\mathbb H(L^\theta)(g,f)$ denotes $\mathbb H(L_d^\theta)(g,f)$, evaluated at pairs of consecutive configurations. To encourage regularity during learning, we penalize Hessians that are close to singular by enforcing that $|\det\mathbb H(L^\theta)(g,f)|$ stays above a positive threshold $e^\epsilon$ (relative to the chosen inner product). Two possible choices of the regularizing term are:
\begin{equation*}
    \mathrm{loss}_\mathrm{reg1} (L^\theta)(g,f) = \mathrm{ReLU} \left(e^\epsilon - \left|\det \mathbb{H}(L^\theta)(g,f) \right| \right),
\end{equation*}
or a smooth (gaussian-type) penalty:
\begin{equation*}
    \mathrm{loss}_\mathrm{reg2} (L^\theta)(g,f) = \frac{1}{e^\epsilon\sqrt{2\pi}} \exp\left( -\frac{ |\det\mathbb{H}(L^\theta)(g,f)|^2 }{2e^{2\epsilon}} \right),
\end{equation*}
where $\epsilon$ is a hyperparameter, and $e^\epsilon$ sets the desired threshold away from singularity.
A concrete example with detailed calculations, including a simplification for computing $\det\mathbb{H}$, is provided in \Cref{Appendix: regular Lagrangian}.

\subsection{Example (rigid body on $SE(3)$)}
Let us consider the discrete Lagrangian of a rigid body evolving on $SE(3)$
\begin{equation*}
    \bar{L}_d(R,p,W,z) = 
    \frac{1}{h} (\trace(J_d) - \trace(W J_d)) + \frac{1}{2h} m (z + (W - I)\rho_{\mathrm{COM}})^\top  (z + (W - I)\rho_{\mathrm{COM}})
    - h V(R, p),
\end{equation*}
where $J_d$ is the (symmetric positive definite) inertia matrix for matrix representation, 
$m>0$ the mass and $\rho_{COM}\in\mathbb{R}^3$ the center of mass vector in the body frame (see \Cref{example: L=T-V in SE3}). Assume a linear external damping force, $F(\omega,v) = -D \begin{pmatrix} \omega & z \end{pmatrix}^\top$ where $D\in\mathbb{R}^{6\times 6}$ is a symmetric positive semidefinite matrix.

Our objective is to learn the discrete Lagrangian and the external forces through the components $J_d$, $D$, $\rho_{COM}$ and $V$.
We parameterize each quantity by a set of trainable parameters \( \theta \), yielding 
\begin{equation*}
    J_d^{\theta}\in\mathbb{R}^{3 \times 3}, \qquad
    D^{\theta} \in \mathbb{R}^{6\times 6}, \qquad
    \rho_{COM}^{\theta} \in \mathbb{R}^{3}, \qquad
     V^{\theta}(R_k, p_k, \rho_{COM}^\theta) : \mathbb{R}^{3\times 3} \times \mathbb{R}^3 \times \mathbb{R}^3 \to \mathbb{R}.
\end{equation*}
To ensure $J_d^\theta$ is symmetric positive definite (SPD), we learn vectors $r_\theta\in \mathbb{R}^3$ and $i_\theta = (a_\theta, b_\theta, c_\theta) \in \mathbb{R}^3$, and define
\begin{equation*}
    R_\theta = \exp^{SO}(\widehat{r_\theta}), \qquad
    I_\theta = \begin{bmatrix}
            e^{a_\theta} + e^{b_\theta} & 0 & 0 \\
            0 & e^{a_\theta} + e^{c_\theta} & 0 \\
            0 & 0 & e^{b_\theta} + e^{c_\theta}
        \end{bmatrix}
\end{equation*}
where $\exp^{SO}:\mathfrak{so}(3) \to SO(3)$ is the exponential map (see \Cref{appendix:basic notions about SO3 and SE3}).
Setting $J_\theta = R_\theta I_\theta {R_\theta}^\top $ and $J_d^\theta = \frac{1}{2} \trace(J_\theta) I - J_\theta$ guarantees that $J_d^\theta$ is SPD. 

To promote $D^\theta$ being symmetric positive semi-definite (SPSD), we learn a lower triangular matrix $L_\theta \in \mathbb{R}^{3 \times 3}$ and a vector $\Lambda_\theta = (x_\theta,y_\theta,z_\theta) \in \mathbb{R}^3$ and set
\begin{equation*}
    D^\theta = {L}_\theta {{L}_\theta}^\top + \diag({x_\theta}^2, {y_\theta}^2, {z_\theta}^2)
\end{equation*}
The learning task is to identify $J_d^\theta$, $D^\theta$, $\rho_{COM}^{\theta}$ and $V^\theta$ such that the discrete forced Euler-Lagrange residual specified in Eq.~\eqref{eq: discrete EL eq in SO3} is minimized, together with the regularity penalty described above.

\paragraph{Rigid body in $SO(3)$.}
For a rigid body in $SO(3)$, the discrete Lagrangian reduces to that in \Cref{example: L=T-V in SO3} The quantities we parametrize with trainable parameters are
\begin{equation*}
    J_d^{\theta}\in\mathbb{R}^{3 \times 3}, \qquad
    D^{\theta} \in \mathbb{R}^{3\times 3}, \qquad
    V^{\theta}(R_k) : \mathbb{R}^{3\times 3} \to \mathbb{R},
\end{equation*}
and we impose on $J_d^\theta$ and $D^\theta$ the same structure as in the $SE(3)$ case, ensuring that they are SPD and SPSD, respectively.

\paragraph{Multibody systems.} In a multibody system with $m$ bodies evolving on $SE(3)$ and $n$ bodies on $SO(3)$, we write a discrete Lagrangian for each body (see \Cref{section: Generalization to multibody systems}) and learn all the components as if they were single rigid bodies in $SE(3)$ or $SO(3)$. To capture interactions between bodies, we additionally parameterize an interaction potential $V_\mathrm{int}^\theta$ as a function of the configurations of all bodies.

\paragraph{Nonlinear external forces.}
The dissipative load consider above assumes dissipation linearly proportional to the velocity, with a constant damping matrix. A more general formulation---still with linear dependence on velocity---allows the damping matrix to vary with the configuration
\begin{equation*}
    F^\theta(R, p, \omega, v) = -D^\theta(R, p) \begin{pmatrix} \omega \\ v \end{pmatrix},
\end{equation*}
where $D^\theta(R, p) \in \mathbb{R}^{6 \times 6}$. Factorizing $D^\theta(R,p) = L_\theta(R,p)\, L_\theta(R,p)^\top$, with $L^\theta : SE(3) \to \mathbb{R}^{6 \times 6}$, guarantees that $D^\theta(R, p)$ is SPSD for all configurations.
This allows damping intensity and principal directions to vary with body pose, while preserving the dissipative structure.
 
For fully general external forces, one may use an unconstrained neural network
\begin{equation*}
    F^\theta(R, p, \omega, v) = \mathrm{NN}^\theta(R, p, \omega, v) \in \mathbb{R}^6,
\end{equation*}
which can represent any arbitrary nonlinear, non-dissipative, or energy-injecting forces (e.g. Coulomb friction, velocity-dependent inputs). The gain in expressiveness comes at the cost of the dissipative guarantee provided by the structured parameterizations above.

\subsection{Inference method}
\label{subsection: inference routine}
After training, we perform forward prediction by integrating the learned discrete system with the discrete forced Euler-Lagrange equations. Rather than reconstructing a continuous-time model, we evolve the system directly in discrete time using the learned discrete Lagrangian and forces. In the $SE(3)$ case, for example, given $(R_{k-1}, R_k, p_{k-1}, p_k)$ we solve the discrete forced Euler-Lagrange equations at each step for the group increments $(W_k, z_k)$ (e.g. via a root-finding method), and then update $(R_{k+1}, p_{k+1})$ accordingly. Repeating this recursion yields the predicted trajectory.
To ensure that the updated matrix $W_k$ lies in $SO(3)$, we solve for a vector $w_k\in \mathbb{R}^3$ such that  $\exp^{SO}(\hat{w}_k)=W_k$
(see \Cref{appendix:basic notions about SO3 and SE3}).

\section{EXPERIMENTS} 
\label{sec:experiments LieDFLNN}
The purpose of the experiments is to evaluate the importance of (i) representing the data using a Lie group and (ii) adopting a model formulation that preserves the underlying Lie group structure such that the learned dynamics remain on the appropriate configuration manifold. We concentrate on systems evolving on $SO(3)$ and $SE(3)$. We investigate each component of the proposed method one at a time, comparing it with a number of different baselines.

We consider five experimental settings, each isolating one aspect of the proposed method. First, we investigate the choice of coordinates by comparing the proposed Lie group-based method with an analogous Euclidean model using Euler angles. Second, we evaluate the discrete Lie group formulation against baselines derived from the continuous forced Euler-Lagrange equations, assessing whether the discrete variational structure matters when only discrete observations are available. Third, we compare our geometry-aware approach with models that operate on matrix entries of the group elements but do not constrain predictions to the manifold, showing that the data representation alone is insufficient. Fourth, we apply the proposed model to a multibody human-motion dataset on the product manifold $SE(3) \times SO(3)^{B-1}$, focusing on a cartwheel sequence with large rotations and coordinated multibody interactions. Lastly, we fix the Lagrangian and learn only an unknown feedback control from position data.

In all experiments, the observations lie on the respective group $G$ up to numerical precision: the synthetic data are integrated with high accuracy, and the real-world human-motion data are given in Euler angles, which map exactly to $SO(3)$. For observations given in an ambient representation, such as noisy rotation matrices in the space of $3\times 3$ matrices, this property does not hold in general. In that case, a possible remedy is to project each observation onto $G$ before the increments are computed. 

Architecture and training
hyperparameters for every experiment are reported in \Cref{appendix: implementation details} (\Cref{tab:hyperparams-arch} and~\Cref{tab:hyperparams-train}).

\subsection{Experiment 1: Euler angles vs.\ matrix representation on $SO(3)$}
\label{subsection: experiment 1}
\begin{figure}[ht]
    \centering
    \begin{subfigure}[c]{0.35\linewidth}
        \includegraphics[width=\linewidth]{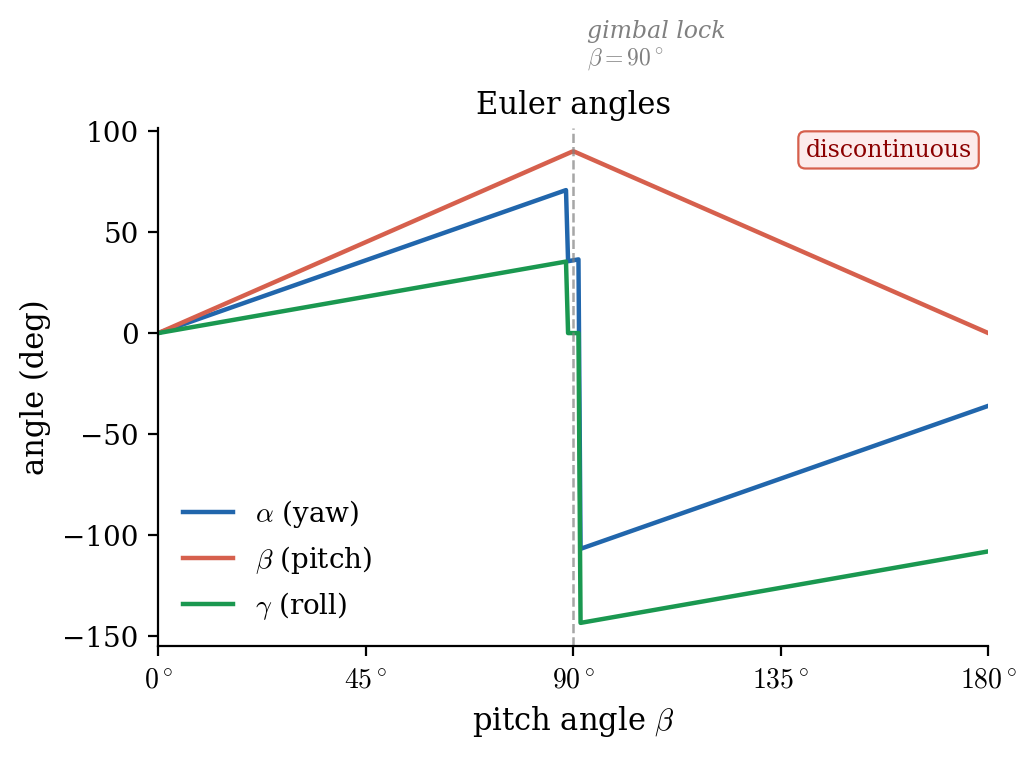}
        \caption{ }
        \label{fig:euler_discontinuity}
    \end{subfigure}
    \hspace{1cm}
    \begin{subfigure}[c]{0.35\linewidth}
        \includegraphics[width=\linewidth]{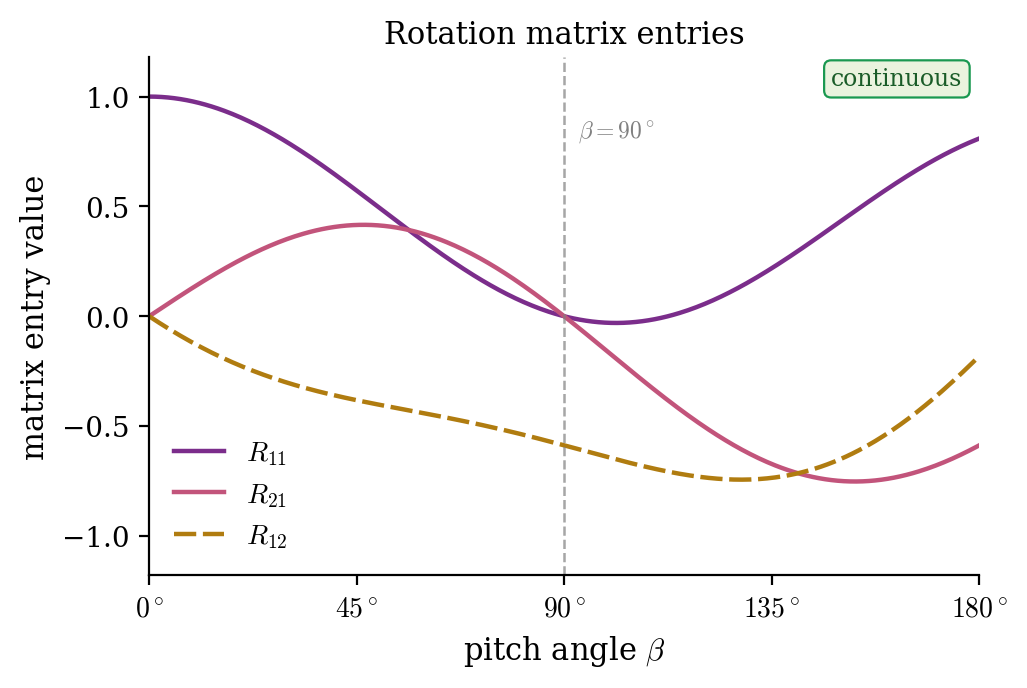}
        \caption{}
        \label{fig:rotmat_continuity}
    \end{subfigure}
    \caption{
        Discontinuity of Euler angles vs.\ continuity of rotation matrices. A smooth rotation trajectory (pitch $\beta$ sweeping from $0^\circ$ to $180^\circ$) is parametrized by $ZYX$ Euler angles~\subref{fig:euler_discontinuity} and rotation matrix entries~\subref{fig:rotmat_continuity}. The Euler-angle representation exhibits a discontinuity at the gimbal-lock point $(\beta = 90^\circ)$, leading to abrupt jumps in $\alpha$ and $\gamma$. In contrast, the rotation matrix entries remain smooth throughout.
    }
    \label{fig:rotation_representations}
\end{figure}

Euler angles provide a compact three-dimensional parameterization of rotations but constitute only a local coordinate representation of $SO(3)$: the map between Euler angles and rotation matrices is not globally invertible. In particular, Euler angles suffer from angle wrapping and from kinematic singularities such as gimbal lock, where one degree of freedom is lost and the remaining angles exhibit discontinuous jumps. As a result, a smooth trajectory on $SO(3)$ may appear discontinuous when expressed in Euler angles, as illustrated in \Cref{fig:rotation_representations}. This forces a model trained on Euler angles to fit artificial coordinate discontinuities that are not present in the physical motion.

This experiment isolates the effect of the configuration-space representation. We compare the proposed LieDFLNN, which operates directly on rotation matrices $R \in SO(3)$, against a discrete Euler-Lagrange baseline that uses Euler angles as generalized coordinates. To stress-test both representations, we systematically vary the magnitude of angular displacements, ranging from small rotations where Euler angles remain well-behaved to large rotations where singularities and wrapping artifacts are expected to degrade the baseline.
The baseline uses a single chart, which is a common practice in many Euclidean learning approaches~\cite{cranmer2020lagrangian,greydanus2019hamiltonian,lutter2019deep}.

\paragraph{Setup and data generation.} 
We consider a single rigid body evolving on $SO(3)$ subject to an external gravitational field and a frictional damping torque (see \Cref{example: L=T-V in SO3} and \Cref{appendix: Rigid body on SO3 in a gravitational field with Rayleigh dissipation}). The governing equations are
\begin{align}
    \label{eq:rigid_body_ode_SO3_potencial omega}
    \dot{\omega} 
    &= J^{-1} \left(- \omega \times (J\omega) + \tau \right), \\
    \label{eq:rigid_body_ode_SO3_potencial R}
    \dot{R} &= R\hat{\omega},
\end{align}
where $J\in\mathbb{R}^{3\times3}$ is the symmetric positive definite inertia matrix for vector representation, and $\tau = \tau_\mathrm{cons} + \tau_\mathrm{ext} $ is the sum of the conservative gravitational torque and the external Rayleigh dissipation torque.

Training data are generated by integrating Eq. \eqref{eq:rigid_body_ode_SO3_potencial omega}-\eqref{eq:rigid_body_ode_SO3_potencial R} for $256$ distinct initial conditions ($R_0$, $\omega_0$), each over $100$ timesteps with stepsize $h = 0.05$, using the explicit Runge-Kutta RK45 solver from \\ \texttt{scipy.integrate.solve\_ivp}. The rigid body has mass $m = 0.5$ and center of mass offset $\rho_\mathrm{com} = (0,\, 0.5,\, 1.0)^\top$, and diagonal inertia tensor
$J = \mathrm{diag}(1.521,\, 1.362,\, 1.211)$, under gravitational acceleration $g = 9.81$. The system includes linear dissipation with diagonal damping matrices $D_{\omega\omega} = 0.1\,I$. The generated trajectories are split into a training set ($90\%$) and a test set ($10\%$).

To probe the effect of rotation magnitude in a controlled manner, all trajectories are initialized at the identity $R_0=I$, and three datasets are produced by scaling the initial angular velocity $\omega_0$ by factors of $0.1$, $0.5$ and $1.0$. This yields motions ranging from near-identity rotations to large-angle motions that traverse regions of $SO(3)$ where Euler-angle singularities are encountered.

The same underlying physical trajectories are represented in two ways:
\begin{equation*}
    R_k \in SO(3)
\end{equation*}
for the proposed model, and
\begin{equation*}
    q_k = (\phi_k, \theta_k, \psi_k) \in \mathbb{R}^{3}
\end{equation*}
for the baseline, where $(\phi_k, \theta_k, \psi_k)$ are ZYX Euler angles
extracted from the same rotation matrices. The lie group model is trained directly on sequences $\{R_k\}$, while the Euler-angle baseline is trained on the corresponding angle sequences.

\paragraph{Results and discussion.} 
\begin{figure}[ht]
    \centering
    \begin{subfigure}[c]{0.5\textwidth}       
        \includegraphics[width=\linewidth]{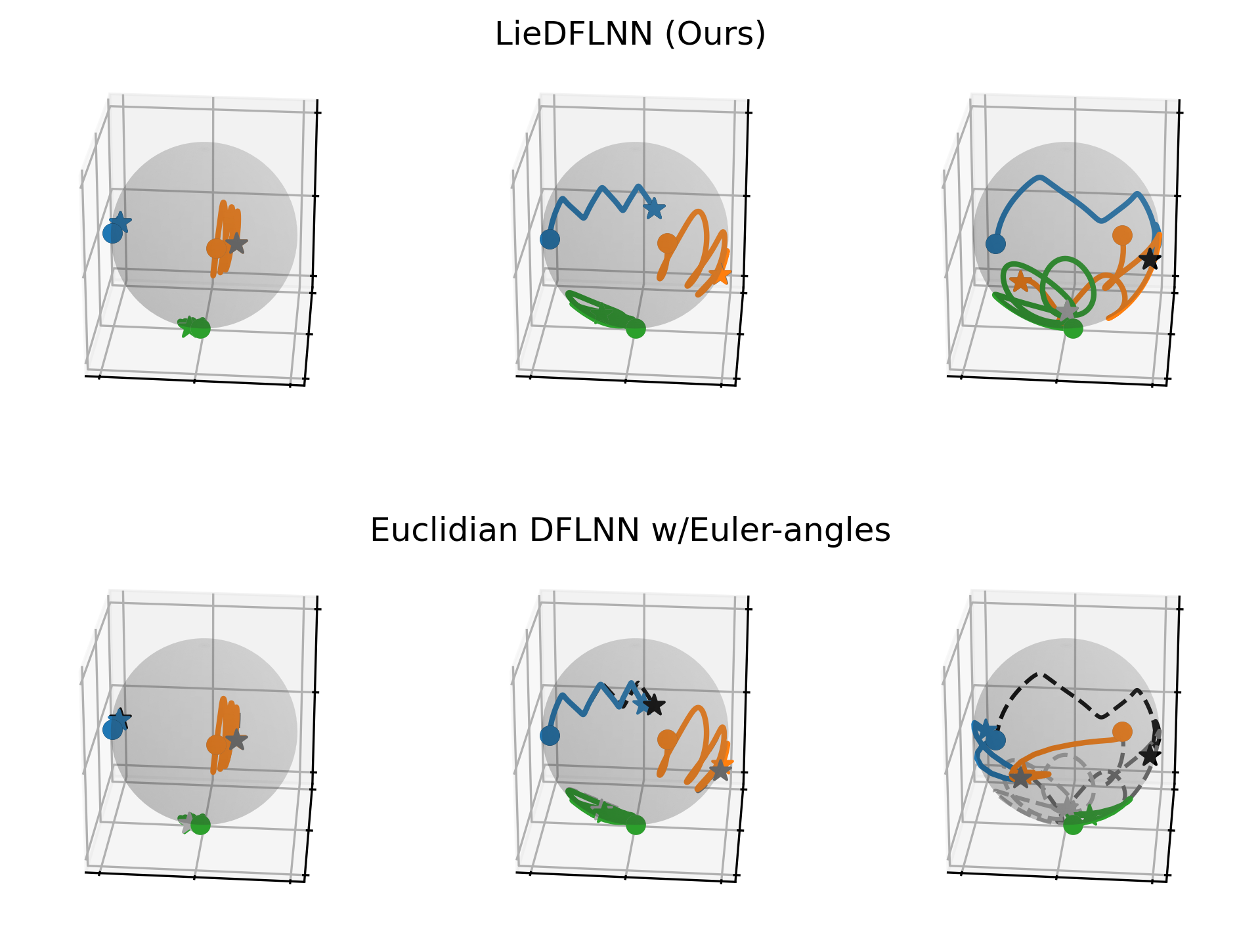} \\
        \vspace{-0.2em}
        \begin{minipage}{0.0\linewidth}
            \centering
        \end{minipage}
        \hfill
        \begin{minipage}{0.2\linewidth}
            \centering
            \scriptsize  0.1
        \end{minipage}
        \hfill
        \begin{minipage}{0.30\linewidth}
            \centering
            \scriptsize 0.5
        \end{minipage}
        \hfill
        \begin{minipage}{0.25\linewidth}
            \centering
            \scriptsize 1.0
        \end{minipage}
        \hfill
        \begin{minipage}{0.15\linewidth}
            \centering
        \end{minipage}\\
        \vspace{-0.2em}
        \begin{minipage}{\linewidth}
            \centering
            \footnotesize Initial Angular Velocity scaling factor
        \end{minipage}
        \caption{}
        \label{fig:euler_vs_lie_sphere}
    \end{subfigure}
    \hfill
    \begin{subfigure}[c]{0.48\textwidth}
        \centering
        \includegraphics[width=0.95\linewidth]{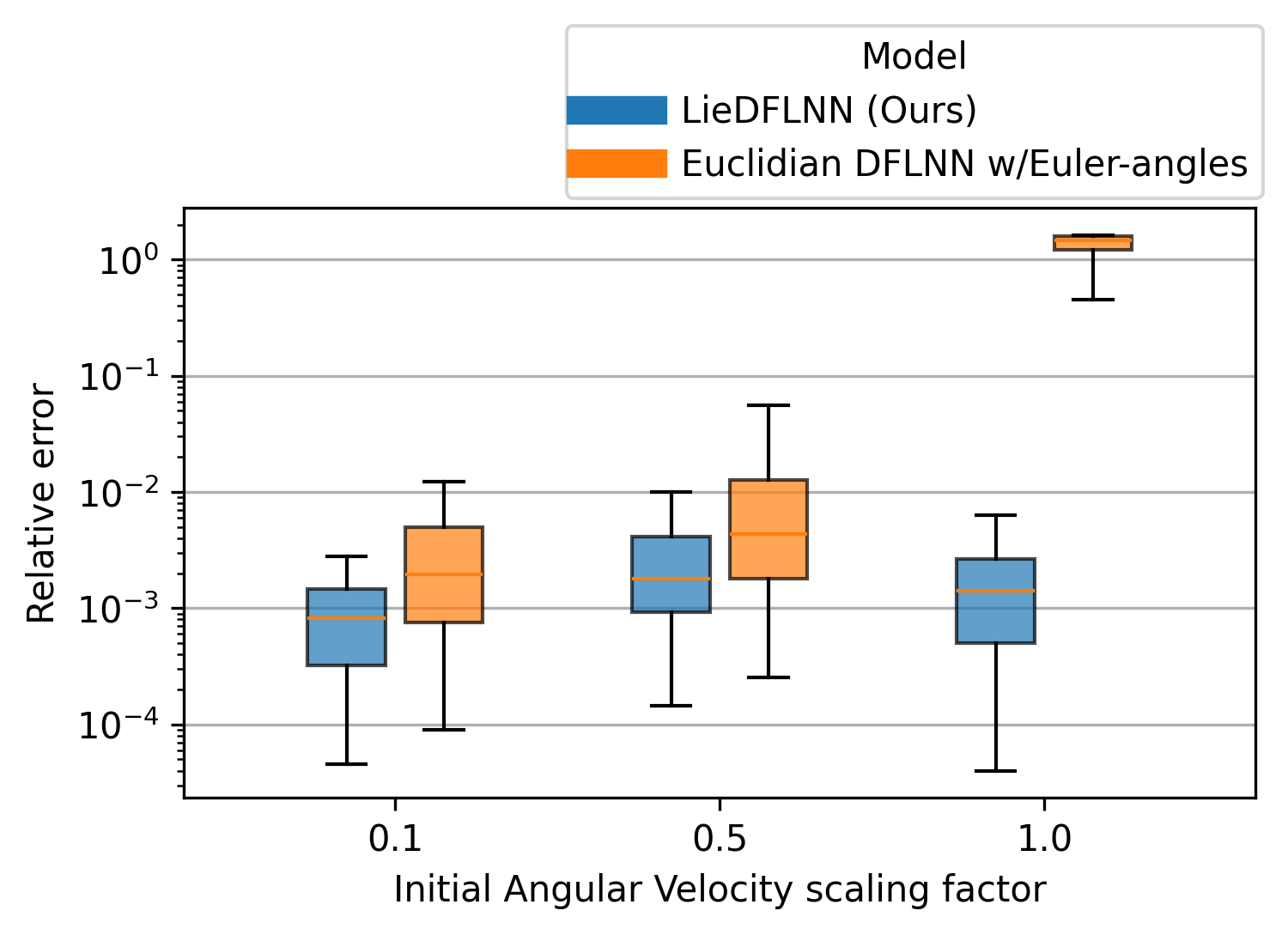}
        \caption{}
        \label{fig:euler_vs_lie_relative-error}
    \end{subfigure}
    \caption{
    Effect of the configuration-space representation under increasing rotation magnitude ($0.1\times$, $0.5\times$, $1.0\times$ initial angular velocity). \Cref{fig:euler_vs_lie_sphere}: Predicted rollouts on the unit sphere (ground truth: black). \Cref{fig:euler_vs_lie_relative-error}: Relative rotation errors, with Euler-angle predictions converted to $SO(3)$ before evaluation to ensure a fair comparison. At small scales both methods track the true trajectory closely. As the rotation magnitude increases, the Euler-angle baseline degrades as trajectories approach singular regions, while the LieDFLNN maintains low and stable error across all scales.
    }
    \label{fig:euler_vs_lie}
\end{figure}

\Cref{fig:euler_vs_lie} summarizes the performance comparison. Qualitatively (\Cref{fig:euler_vs_lie_sphere}), both models track the ground truth closely at the smallest scale $(0.1\times)$. However, as rotation magnitude increases, the Euler-angle baseline produces visibly inaccurate paths, while the LieDFLNN remains coincident with the ground-truth curves.

Quantitatively, \Cref{fig:euler_vs_lie_relative-error} reports the relative rotation error across scales. To ensure a fair comparison, Euler-angle predictions are converted back to $SO(3)$ before evaluation. At small magnitudes, the performance is comparable. However, at scale $(1.0\times)$, where trajectories pass through or near singular regions, the Euler-angle baseline degrades substantially due to the coordinate discontinuities shown in \Cref{fig:euler_discontinuity}. In contrast, the LieDFLNN maintains consistently low error across all scales, as rotation matrices remain smooth and non-singular everywhere on the manifold.

\subsection{Experiment 2: Discrete vs.\ continuous Euler-Lagrange equations on $SO(3)$}

Continuous-time forced Euler-Lagrange models can also be trained from discretely sampled
data, by integrating the learned vector field between the observation times and comparing
the predicted and observed states
\cite{chen2018neuralode, duong2021hamiltonian, duong2024port}. Compared with the discrete
formulation, this has two practical drawbacks. First, the numerical integration must
respect the group structure: classical solvers do not keep the trajectory on $SO(3)$ and
lead to geometric drift, while geometrically exact integrators are often implicit, costly
to unroll and differentiate through, and usually require a time grid finer than the data
sampling. Second, a second-order continuous model requires an initial angular velocity
which, when only positions are observed, has to be estimated by finite differences,
introducing an error in the initial condition.

This experiment compares our discrete Lie group formulation against continuous-time
Euler-Lagrange baselines that share the same $SO(3)$ structure and are trained by
unrolling the learned dynamics with different integrators. By varying the training
trajectory length, we investigate whether the discrete formulation offers superior accuracy and robustness when learning from limited temporal context.

\paragraph{Setup and data generation.} We train and evaluate on a single rigid body evolving on $SO(3)$ subject to an external torque $\tau$. The governing equations are \eqref{eq:rigid_body_ode_SO3_potencial omega}-\eqref{eq:rigid_body_ode_SO3_potencial R} as in Experiment~1, but here $\tau$ has no gravitational component and consists only of the external torque (see \Cref{appendix: Rigid body on SO3 in a gravitational field with Rayleigh dissipation}).

The training dataset consists of $128$ trajectories, each generated from a
random initial configuration $(R_0, \omega_0)$ and integrated for $100$
timesteps with stepsize $h = 0.1$, using the explicit Runge-Kutta RK45 solver from \texttt{scipy.integrate.solve\_ivp}. The rigid body has mass $m = 0.5$ and center of mass offset $\rho_\mathrm{com} = (1.2,\, 0.5,\, 1.0)^\top$, and diagonal inertia tensor
$J = \mathrm{diag}(1.521,\, 1.362,\, 1.211)$ in the absence of gravitational acceleration ($g = 0$). The system includes linear dissipation acting only on
the angular velocity, with $D_{\omega\omega} = 0.05\,I$ and vanishing velocity and cross-damping terms $D_{vv} = D_{v\omega} = D_{\omega v} = 0$. The generated trajectories are split into a training set ($90\%$) and a test set ($10\%$). To examine how sensitive each model is to the length of the trajectory, we train on four nested subsets of this dataset, using only the first $10$, $40$, $70$, and $100$ timesteps of each trajectory respectively.

We compare the LieDFLNN against three baselines based on Geometric Lagrangian Neural Networks GLNN~\cite{xiao2024generalized},  which differ in their integration schemes:
\begin{description}
    \item[GLNN w/RK4.] The continuous equations are integrated with the adaptive RK4(5) solver provided by \texttt{torchdiffeq}~\cite{chen2018neuralode}. A generic solver of this kind does not respect the geometry of $SO(3)$ and may accumulate drift away from the rotation manifold over long trajectories.
    \item[GLNN w/Lie-Midpoint.] To preserve the $SO(3)$ structure, we consider a second order implicit Lie group integrator. Let us denote the map in \eqref{eq:rigid_body_ode_SO3_potencial omega} with $\dot\omega =: A(\omega)$. Given $(R_k, \omega_k)$, the next angular velocity $\omega_{k+1}$ is found by solving the implicit equation
    \begin{equation*}
        \omega_{k+1} = \omega_k + h\,A\!\left(\frac{\omega_k + \omega_{k+1}}{2} \right),
    \end{equation*}
    and then the rotation is updated by 
    \begin{equation*}
        R_{k+1} = R_k \exp^{SO} \left(h \, \left(\frac{\widehat{\omega_k + \omega_{k+1}}}{2} \right) \right ).
    \end{equation*}
    \item[GLNN w/Lie-Heun.] As an explicit alternative to the Lie-Midpoint, we use the Lie-Heun method, a second-order Runge-Kutta scheme adapted to Lie groups. Given $(R_k, \omega_k)$, the update proceeds as
    \begin{align*}
        \omega_{k+1} &= \omega_k + \frac{h}{2} \left( A(\omega_k) + A(\omega^\star) \right ), \\
        R_{k+1}  &= R_k\exp^{SO}\!\left(h\,\widehat{\left(\frac{\omega_k + \omega^\star}{2}\right)}\right),
    \end{align*}
    where $\omega^\star = \omega_k + h\,A(\omega_k)$.
\end{description}

The three baselines differ in how the learned dynamics are unrolled during
training. For \textit{GLNN w/RK4} and \textit{GLNN w/Lie-Heun}, the model is integrated along the full trajectory and the predicted sequence is compared to the ground truth before backpropagation. For \textit{GLNN w/Lie-Midpoint}, which is implicit and therefore expensive to unroll, we instead perform a single two-step update per training iteration. This matches the local context available to the proposed model, which likewise uses only three consecutive poses $(R_{k-1}, R_k, R_{k+1})$ at a time. For all methods, the initial angular velocity is recovered from the first two observed rotations $\hat{\omega}_0 = \frac{1}{h} \log^{SO}(R_0^\top R_1)$.

\paragraph{Results and discussion.}

At test time, all models are evaluated by integrating $10$ unseen trajectories over $500$ timesteps---five times longer than the longest training sequence.

\begin{figure}[ht]
    \centering
    \begin{subfigure}[b]{0.5\textwidth}
        \includegraphics[width=\linewidth]{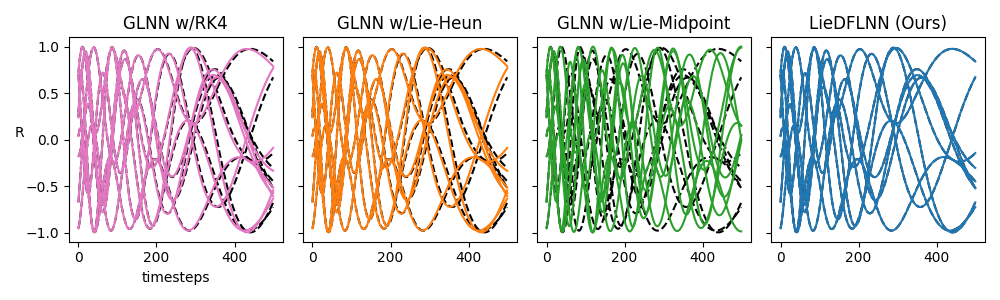} \\
        \includegraphics[width=\linewidth]{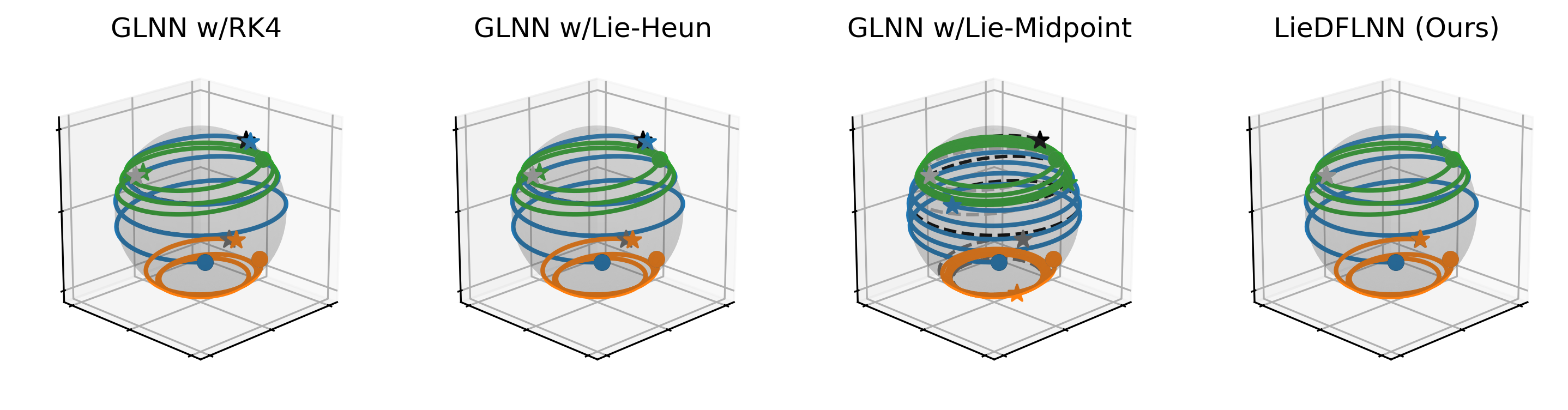}
        \caption{}
        \label{fig:free_rigid_body prediction sample}
    \end{subfigure}
    \hfill
    \begin{subfigure}[b]{0.48\textwidth}
        \includegraphics[width=\linewidth]{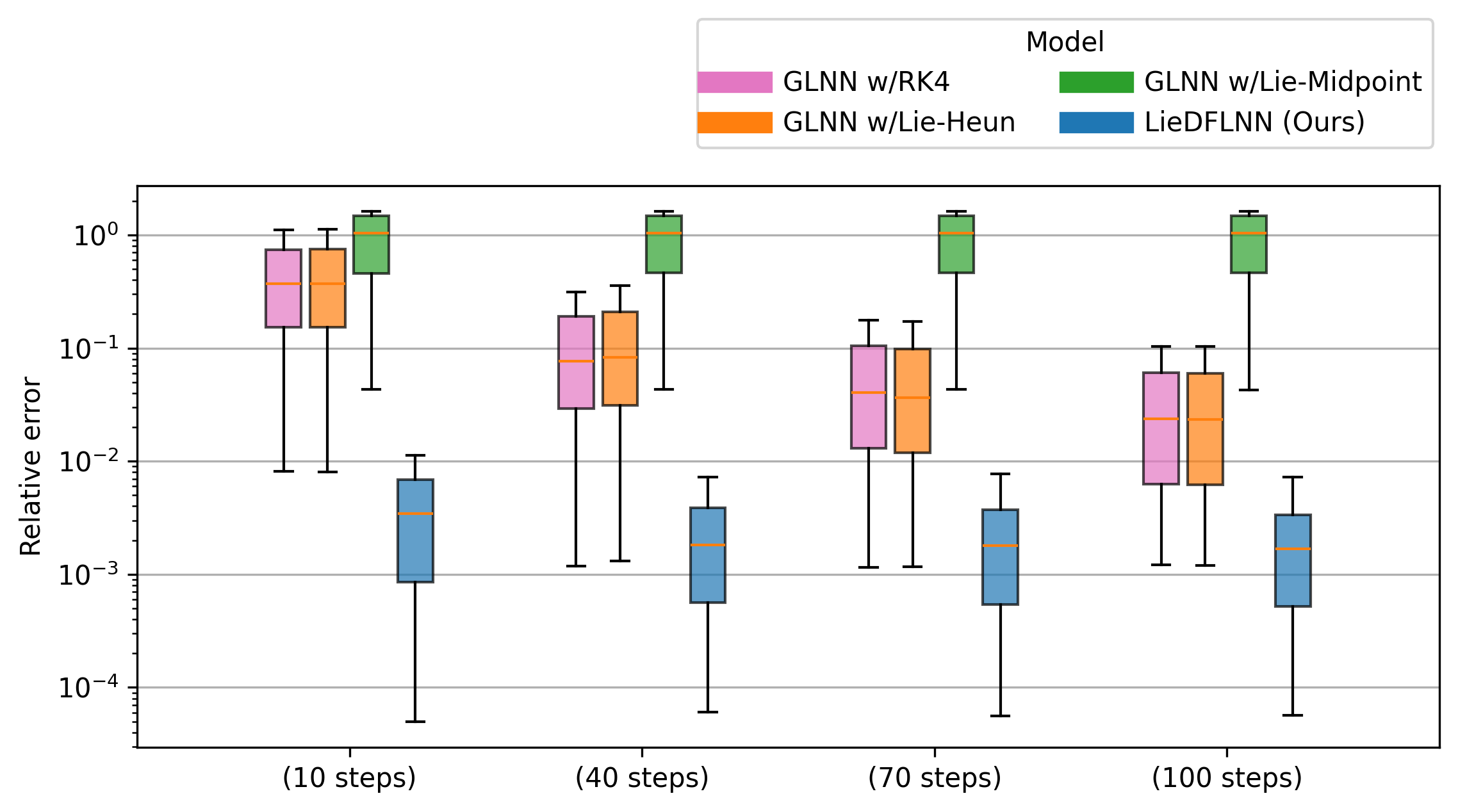}
        \caption{}
        \label{fig:free_rigid_body relative-error}
    \end{subfigure}
    \caption{Discrete vs.\ continuous Euler-Lagrange formulations on $SO(3)$. \Cref{fig:free_rigid_body prediction sample}: Qualitative 500-step test rollout (ground truth: black). \Cref{fig:free_rigid_body relative-error}: Mean test error relative to training trajectory length. The proposed discrete formulation (blue) achieves high accuracy even with the shortest training windows (10 steps). Continuous baselines (RK4, Lie-Heun) require longer trajectories to improve, yet remain less accurate than the proposed method. The failure of the Lie-Midpoint baseline highlights the necessity of the discrete variational structure for learning from discrete pose triplets.}
    \label{fig:free_rigid_body}
\end{figure}

\Cref{fig:free_rigid_body prediction sample} provides a qualitative illustration of a single test trajectory integrated over $500$ timesteps. 
\Cref{fig:free_rigid_body relative-error} shows the test error (average over the 10 trajectories) for all models as a function of training trajectory length. The proposed LieDFLNN performs consistently well across all lengths; its error is nearly invariant to the number of training timesteps. This suggests that the discrete variational structure allows the model to extract maximal dynamical information from just three successive poses, making long training rollouts unnecessary.

Among the baselines, \textit{GLNN w/Lie-Midpoint} performs significantly worse.  
When restricted to a local training window, the continuous-time model must estimate $\omega_{k-1}$ from the pair $(R_{k-1},R_k)$ through the log map. This first-order estimate enters every residual, and there is no long rollout to average out its error. The discrete formulation takes the pair $(R_k, W_k)$ as its state and needs no velocity estimate at all. That our proposed model succeeds under the identical "local" supervision confirms that the discrete variational formulation is the key to learning effectively from discrete observations.

The \textit{RK4} and \textit{Lie-Heun} baselines improve as the training window increases, benefiting from the richer signal provided by long-term unrolling. However, even with the maximum training context (100 steps), they remain less accurate than our proposed model trained on only 10 steps. This demonstrates that the geometric consistency of the discrete Euler-Lagrange equations provides a fundamental advantage in both data efficiency and long-term predictive accuracy.

\subsection{Experiment 3: Constraining the model to remain on the data manifold}
Experiment 1 demonstrated the benefits of globally valid parameterizations; we now investigate whether explicitly enforcing the manifold constraint is necessary for long-term stability. Without such constraints, a model may treat rotation matrix entries as independent coordinates. While potentially accurate on the training distribution, such models can accumulate errors that violate the orthogonality constraint $R^\top R=I$. Crucially, these off-manifold predictions lie entirely outside the training distribution, leading to unpredictable behavior as the model operates on inputs it has never encountered.
This experiment isolates that effect by comparing models that share the same data representation but differ in whether the manifold structure is enforced by construction.

\paragraph{Setup and data generation.}
We consider a single rigid body evolving on $SE(3)$ subject to an external gravitational field and a frictional damping torque (see \Cref{example: L=T-V in SE3} and \Cref{appendix: Rigid body on SE3 in a gravitational field with Rayleigh dissipation}). 
The governing equations are
\begin{align}
    \label{eq:rigid_body_ode_SE3_gravitational_potential omega}
    \dot \omega & = J^{-1}\!\left( \tau
    - \omega \times \big(J\omega + \rho_{\mathrm{COM}} \times P\big)
    - v \times P - \rho_{\mathrm{COM}} \times \big(f - \omega \times P\big) \right) \\
    \label{eq:rigid_body_ode_SE3_gravitational_potential v}
    \dot v &= \frac{1}{m}\big(f - \omega \times P\big)
    - \dot{\omega} \times \rho_{\mathrm{COM}} \\
    \label{eq:rigid_body_ode_SE3_gravitational_potential R}
    \dot R &=R\hat{\omega} \\
    \label{eq:rigid_body_ode_SE3_gravitational_potential p}
    \dot p &= Rv
\end{align}
where $P := m\big(v + \omega \times \rho_{\mathrm{COM}}\big)$, $J\in\mathbb{R}^{3\times 3}$ is the symmetric positive definite inertia matrix for vector representation, $m>0$ is the mass, $\rho_{COM}\in\mathbb{R}^3$ is the center of mass vector in the body frame, $\tau = \tau_\mathrm{cons} + \tau_\mathrm{ext}$ is the sum of the conservative gravitational torque and the external Rayleigh dissipation torque, and $f = f_\mathrm{cons} + f_\mathrm{ext}$ is the sum of the conservative gravitational translational force and the external Rayleigh dissipation translational force. 

Training data are generated by integrating Eq. \eqref{eq:rigid_body_ode_SE3_gravitational_potential omega}-\eqref{eq:rigid_body_ode_SE3_gravitational_potential p} for $512$ distinct initial conditions ($R_0$, $p_0$, $\omega_0$, $v_0$), each over $32$ timesteps with stepsize $h = 0.05$, using the explicit Runge-Kutta RK45 solver from \texttt{scipy.integrate.solve\_ivp}. The rigid body has mass $m = 0.5$ and center of mass offset $\rho_\mathrm{com} = (0,\, 0.5,\, 1.0)^\top$, and diagonal inertia tensor
$J = \mathrm{diag}(1.521,\, 1.362,\, 1.211)$, under gravitational acceleration $g = 9.81$. The system includes linear dissipation with diagonal damping matrices $D_{\omega\omega} = 0.1\,I$. The generated trajectories are split into a training set ($90\%$) and a test set ($10\%$).

We evaluate the proposed LieDFLNN against two baselines trained on the same data.
\begin{description}
    \item[Neural ODE~\cite{chen2018neuralode}:] models a generic unconstrained vector field in $\mathbb{R}^{12}$, with no geometric structure imposed on either the configuration or the velocity.
    \item[Euclidean DFLNN:] uses the discrete Euler-Lagrange formulation, but treats the twelve entries of $g=(R, p)$ as free generalized coordinates, ignoring the $SO(3)$ constraints.
\end{description}

\paragraph{Results and discussion.}

\begin{figure}[ht]
    \centering
    \begin{subfigure}[b]{0.54\textwidth}
        \includegraphics[width=\linewidth]{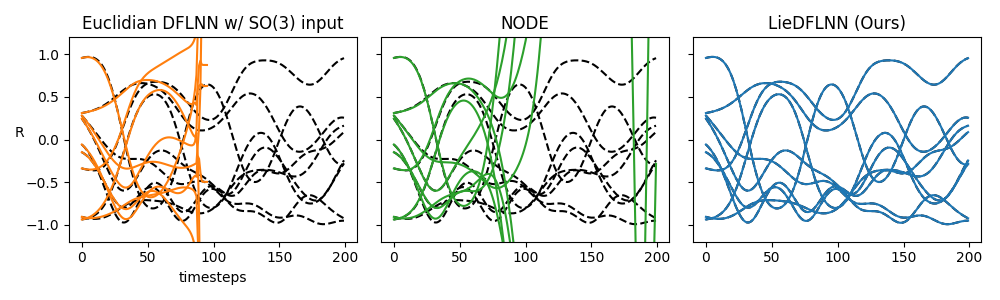} \\
        \includegraphics[width=\linewidth]{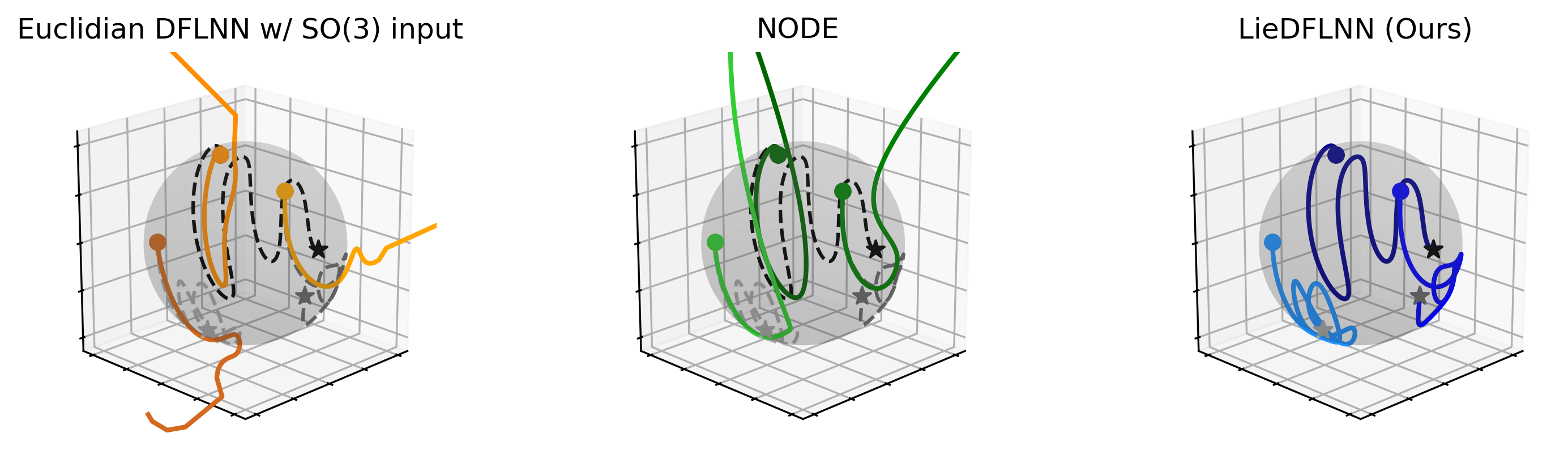}
        \caption{}
    \label{fig:rigid_body_manifold_violation prediction sample}
    \end{subfigure}
    \hfill
    \begin{subfigure}[b]{0.45\textwidth}
        \includegraphics[width=\linewidth]{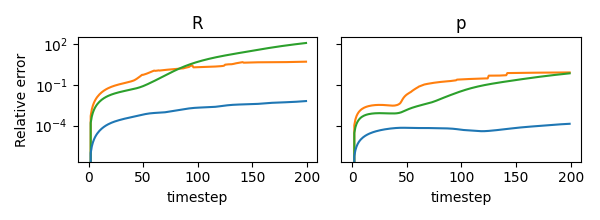} \\
        \includegraphics[width=\linewidth]{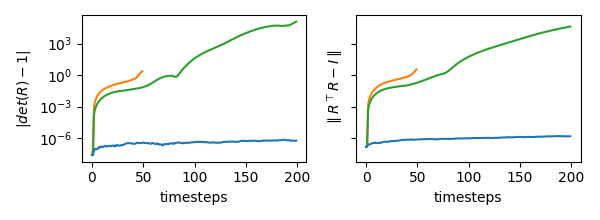}
        \caption{        }
    \label{fig:rigid_body_manifold_violation prediction error}
    \end{subfigure}
    \caption{
    Comparison of the LieDFLNN against Neural ODE and Euclidean DFLNN baselines over a 200-step rollout. \Cref{fig:rigid_body_manifold_violation prediction sample}: Qualitative rotation trajectories (ground truth: black): only the LieDFLNN remains on the unit sphere. \Cref{fig:rigid_body_manifold_violation prediction error}: Relative prediction error (top) and manifold violation measured by determinant drift $|\det(R) - 1|$ (bottom left) and orthogonality drift $\|R^\top R - I\|_F$ (bottom right). The proposed model maintains geometric consistency by construction, whereas baselines drift into non-physical state spaces. 
    }
    \label{fig:rigid_body_manifold_violation}
\end{figure}

\Cref{fig:rigid_body_manifold_violation} compares the performance of the three models over a 200-step test rollout. Qualitatively, the LieDFLNN tracks the ground-truth trajectory on the sphere, while both baselines progressively drift off the manifold, leading to physically implausible configurations (\Cref{fig:rigid_body_manifold_violation prediction sample}).

Quantitatively, the relative prediction error for both baselines grows steadily as they diverge from $SE(3)$ (\Cref{fig:rigid_body_manifold_violation prediction error}, top). 
The bottom panel in \Cref{fig:rigid_body_manifold_violation prediction error} shows the manifold violation of the predicted rollout, measured by the determinant drift $|\det(R) - 1|$ and the orthogonality drift $\|R^\top R - I\|_F$. By construction, the LieDFLNN predictions
remain on $SO(3)$ to machine precision throughout the rollout, whereas both baselines accumulate manifold violation over time.

Notably, the Euclidean DFLNN drifts at a rate comparable to the Neural ODE, despite possessing the same underlying discrete variational structure as our proposed model. This confirms that a variational structure alone is insufficient for geometric consistency; it must be coupled with an explicit Lie group formulation to maintain stability and physical validity.

\subsection{Experiment 4: Multibody real human motion capture}
Having validated the method on synthetic single-body systems, we now consider real-world observations of a multibody articulated system with unknown dynamics. We model human cartwheel motions as a kinematic chain where the root joint evolves on $SE(3)$, and the remaining $B-1$ joints evolve on $SO(3)$, resulting in the product configuration space  $SE(3) \times SO(3)^{B-1}$. These motions are particularly challenging due to simultaneous $360^\circ$ rotations across multiple coupled joints.

\paragraph{What can be expected from a shared force.}

\begin{figure}[ht]
    \centering
    \begin{subfigure}[c]{0.3\textwidth}
        \includegraphics[width=\linewidth]{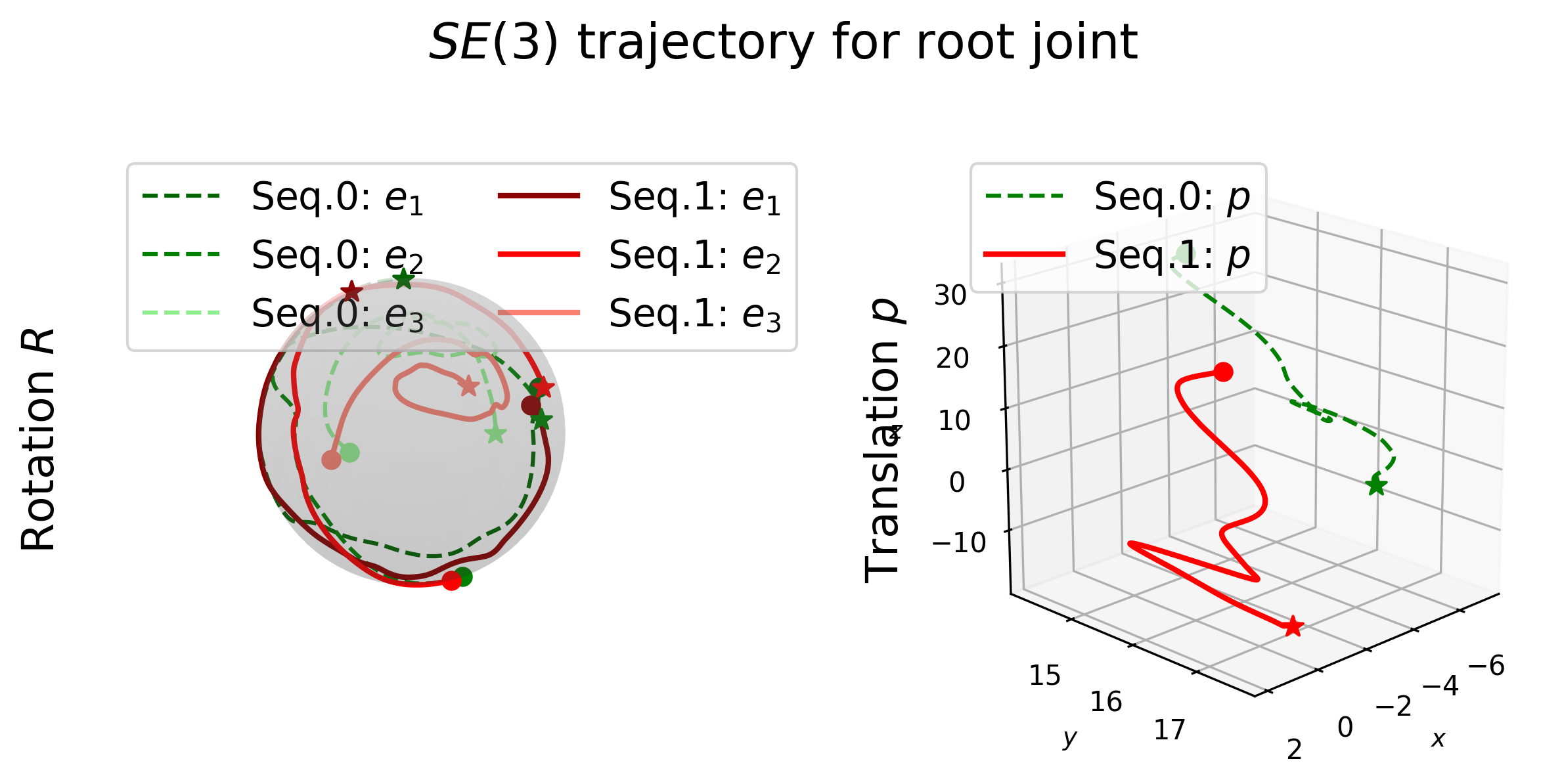}
        \caption{Root joint in SE(3).}
        \label{fig:mocap_root_traj0}
    \end{subfigure}
    \hfill
    \begin{subfigure}[c]{0.69\textwidth}
        \includegraphics[width=\linewidth]{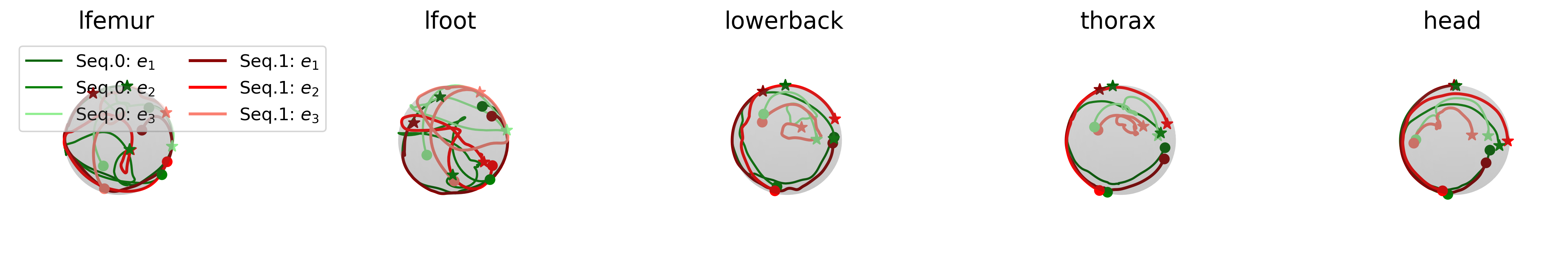}
        \caption{Selection of body joints in SO(3).}
        \label{fig:mocap_root_traj1}
    \end{subfigure}\\
    \begin{subfigure}[c]{0.8\textwidth}
        \includegraphics[width=\linewidth, trim={0cm 2cm 0 3cm}, clip]{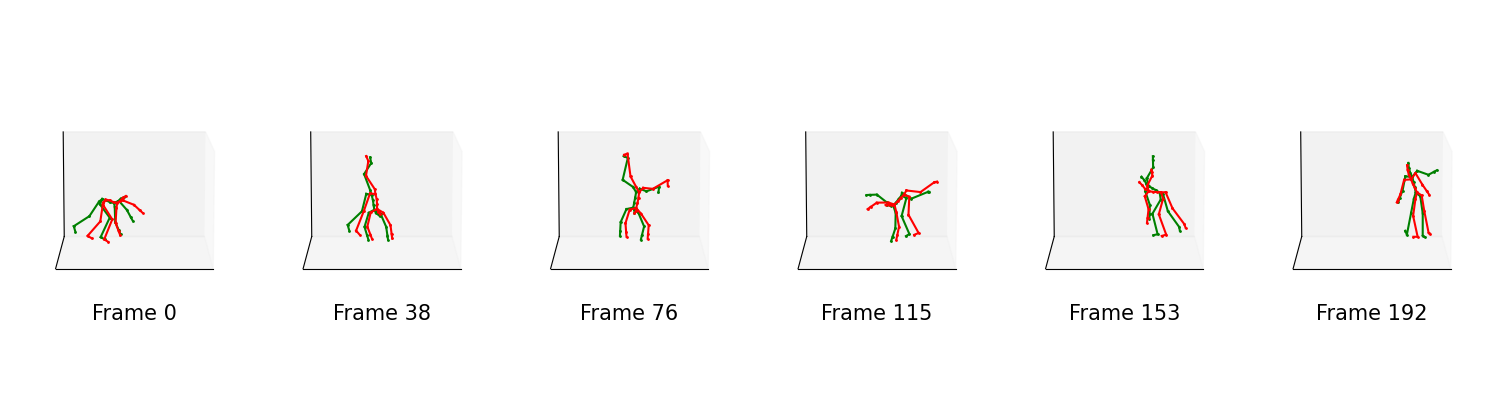}
        \caption{Skeleton renderings.}
        \label{fig:mocap_root_traj3}
    \end{subfigure}
    \caption{An overlay of Sequence 1 (green) and Sequence 2 (red) illustrates the natural variability in execution for the same subject. These visible differences in global paths and local joint rotations establish the baseline scale for evaluating reproduction errors. }
    \label{fig:mocap_data_comparison}
\end{figure}

The learned pair $(L^\theta, F^\theta)$ defines a single \emph{autonomous} discrete dynamical system. Consequently, two observed sequences can only be reproduced exactly if they are solutions to the same state-feedback law. Human motion, however, is internally driven by time-varying muscle actuation rather than a pure function of mechanical state. Since two repetitions of the same movement involve different actuation histories, a shared $F^\theta$ must necessarily compromise where the sequences diverge (\Cref{fig:mocap_data_comparison}).
The learned force should therefore be viewed as an effective state-feedback approximation of the actuated dynamics. Accordingly, our objective is not exact reproduction, but rather to assess: (i) if the model faithfully captures the global skeletal movement, and (ii) if the per-sequence reproduction error remains below the inherent discrepancy between the two ground-truth recordings. This inter-sequence variability provides the natural benchmark for evaluating the model's performance.

\paragraph{Setup and data curation.}
We use human motion capture recordings of cartwheel movements from the Carnegie Mellon University Motion Capture Database (CMU MoCap)~\cite{CMU-Database}, a publicly available collection of human motion sequences recorded at $120$\,Hz using a marker-based optical capture system. The skeleton consists of $B=23$ segments: a root joint in $SE(3)$ (global pose) and $22$ body joints in $SO(3)$, namely the left and right hip, femur, tibia, foot, clavicle, humerus, radius and wrist, lower and upper back, lower and upper neck, thorax and head.
We select Trials 6 and 7 from Subject 49 (\Cref{fig:mocap_data_comparison}), converting the original Euler angles into rotation matrices. Prior to training, the raw motion capture data undergoes preprocessing to remove noise and edge effects. Each joint trajectory is smoothed along the time dimension using a Savitzky--Golay \cite{savitzky_smoothing_1964} filter with a window size of $50$ frames and polynomial order $6$.

A single model $(L^\theta, F^\theta)$ is trained jointly on both sequences, and each sequence is then reproduced by rolling the learned dynamics forward from its true initial configuration pair. The relative motion between the joints is captured by the learned interaction potential $V_{\mathrm{int}}^\theta$ in $L^\theta$, which takes the configurations of all joints as input (see \Cref{section: Generalization to multibody systems}). Training on only two sequences intentionally tests the model's ability to capture complex, high-dimensional coupled dynamics from extremely sparse data. We model the external force as a sum of two components,
$F^\theta = F^\theta_\mathrm{NL} + F^\theta_\mathrm{NN}$. The first, $F^\theta_\mathrm{NL}$, is a nonlinear Rayleigh dissipation term in which the damping matrix $D^\theta(R, p)$ varies with configuration. The second, $F^\theta_\mathrm{NN}$, is an unconstrained neural network that imposes no structural constraints and can represent arbitrary nonlinear, non-dissipative, or energy-injecting effects from the human.

\paragraph{Results and Discussion.}

\begin{table}[ht]
    \centering
    \caption{
        Reproduction errors for cartwheel sequences, with
        discrepancy between the two ground-truth recordings as reference
        scale. Rotation and position errors are reported as mean geodesic distances (rad) and mean Euclidean translation errors, respectively, averaged over the rollout.
    }
    \label{tab:mocap_errors}
    \vspace{1mm}
    \begin{tabular}{*{4}{|c}|}
        \hline
        & Root rot.\ [rad] & Mean joint rot.\ [rad] & Root pos. \\
        \hline
        Sequence 1 (prediction error) & 0.193 & 0.186 & 0.887 \\
        Sequence 2 (prediction error) & 0.188 & 0.233 & 1.409 \\
        Ground truth discrepancy      & 0.373 & 0.416 & 6.624 \\
        \hline
    \end{tabular}
\end{table}

\begin{figure}[ht]
    \centering
    \begin{subfigure}[c]{0.33\textwidth}
        \includegraphics[width=\linewidth]{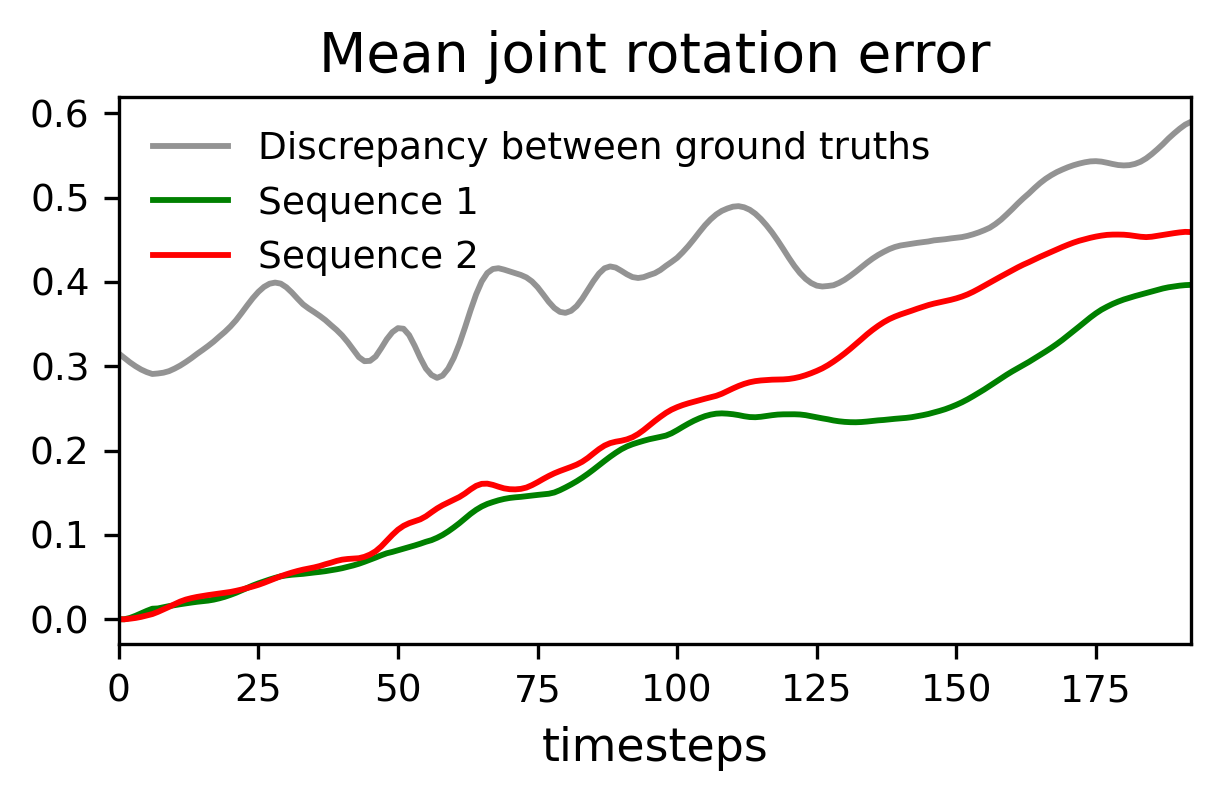}
        \caption{        }
        \label{fig:mocap_error_rotation}
    \end{subfigure}
    \hfill
    \begin{subfigure}[c]{0.66\textwidth}
        \includegraphics[width=\linewidth]{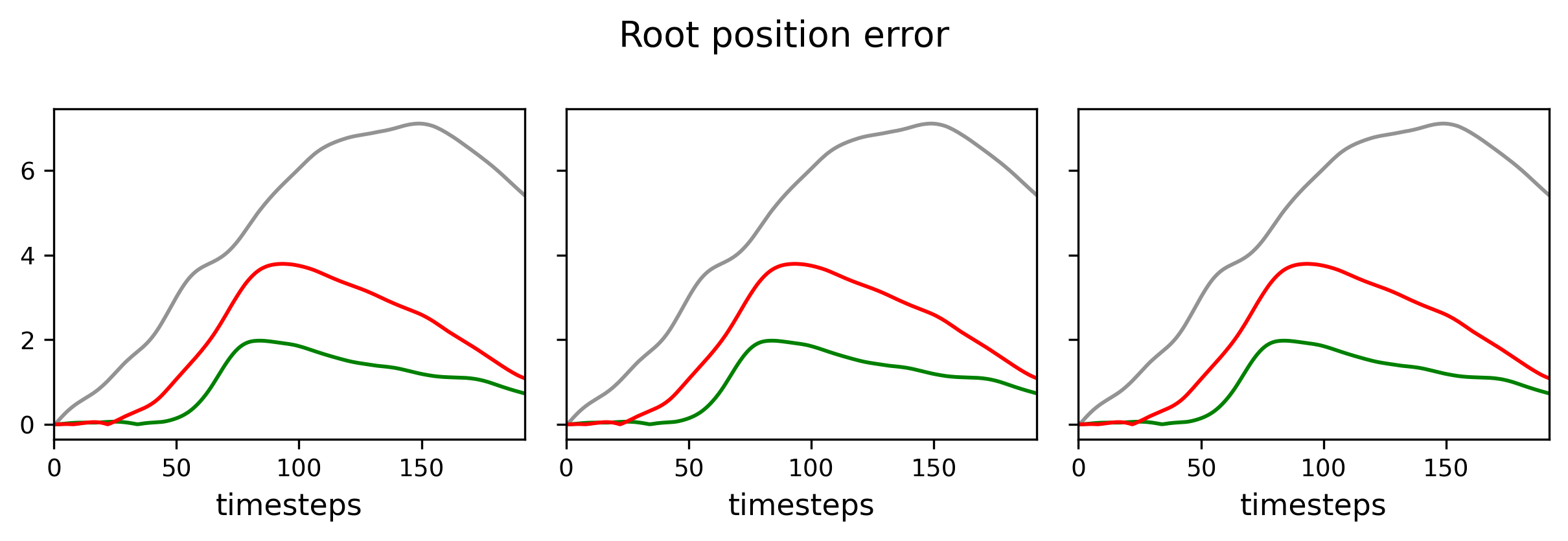}
        \caption{}
        \label{fig:mocap_error_position}
    \end{subfigure}
    \caption{%
        Per-timestep reproduction errors of cartwheel sequences. The gray baseline indicates the discrepancy between the two ground truth recordings. Reproduction errors remain consistently below the ground-truth variability. Error peaks align with phases of high recording divergence, where the shared state-feedback force must mediate between the two reference trajectories.
    }
    \label{fig:mocap_reproduction_error}
\end{figure}

\begin{figure}[ht]
    \centering
    \begin{subfigure}[c]{0.95\textwidth}
        \includegraphics[width=\linewidth, trim={0cm 2cm 0 3cm}, clip]{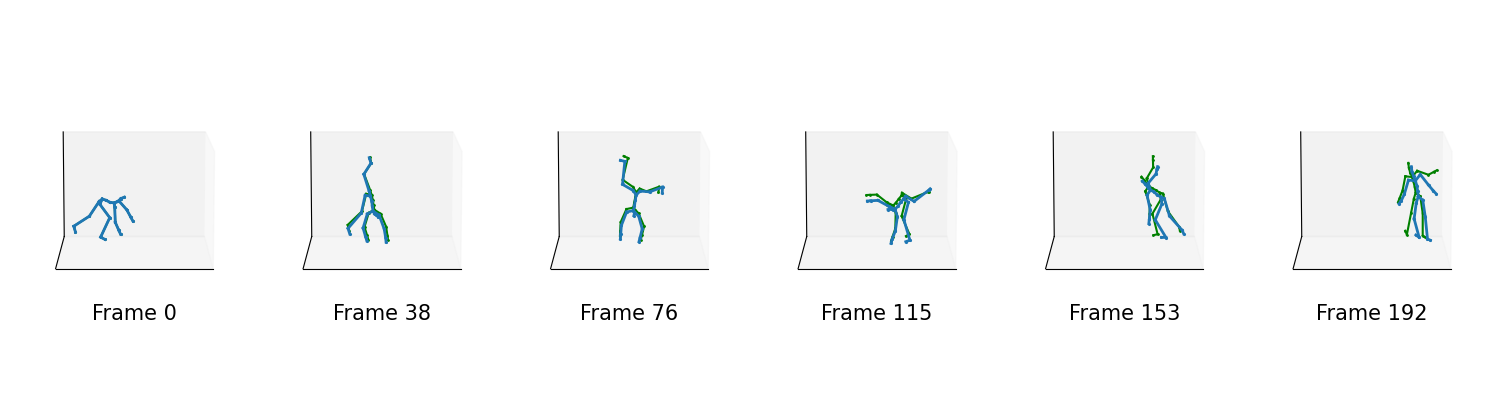} \\
        \includegraphics[width=0.3\linewidth]{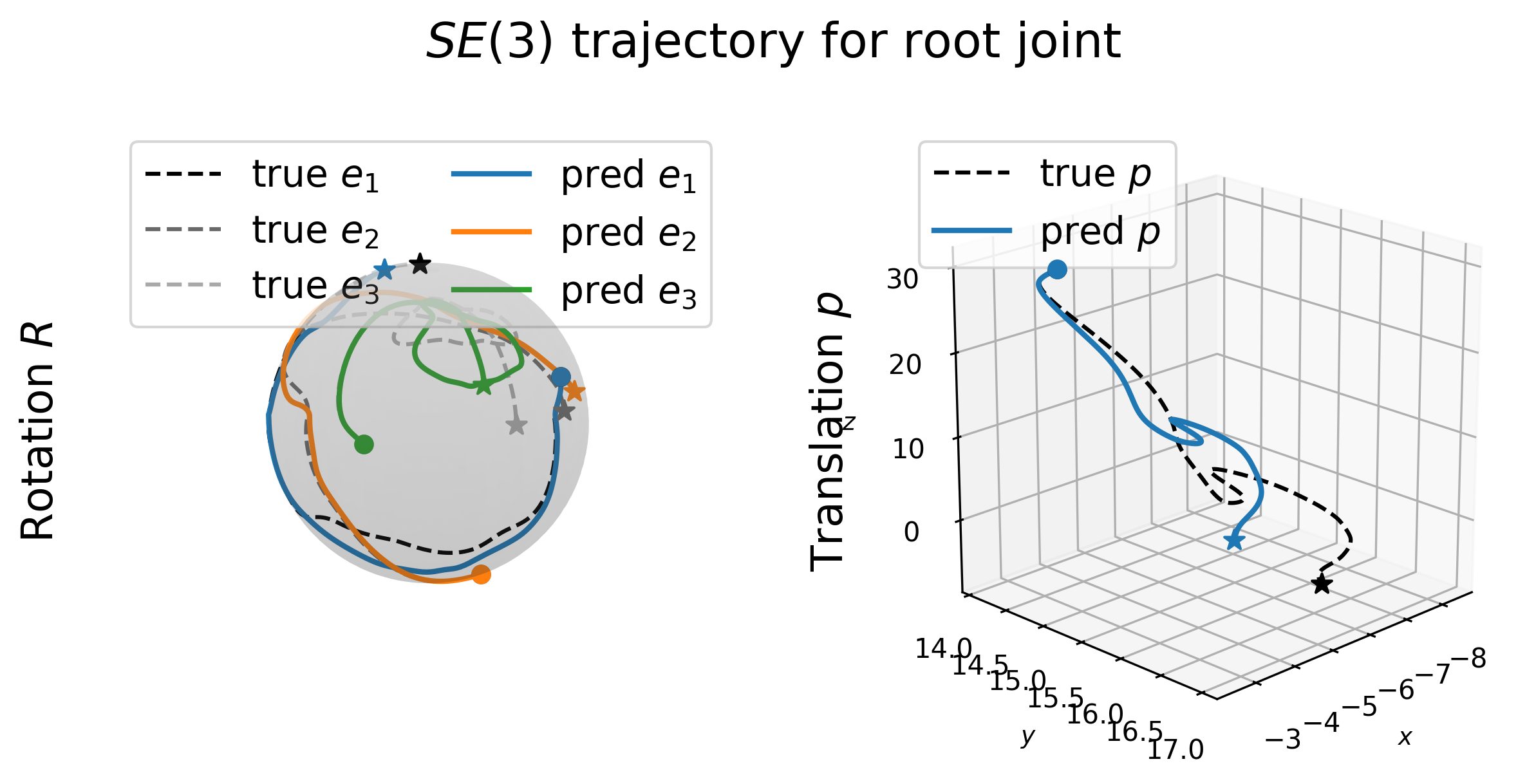}
        \hfill
        \includegraphics[width=0.73\linewidth]{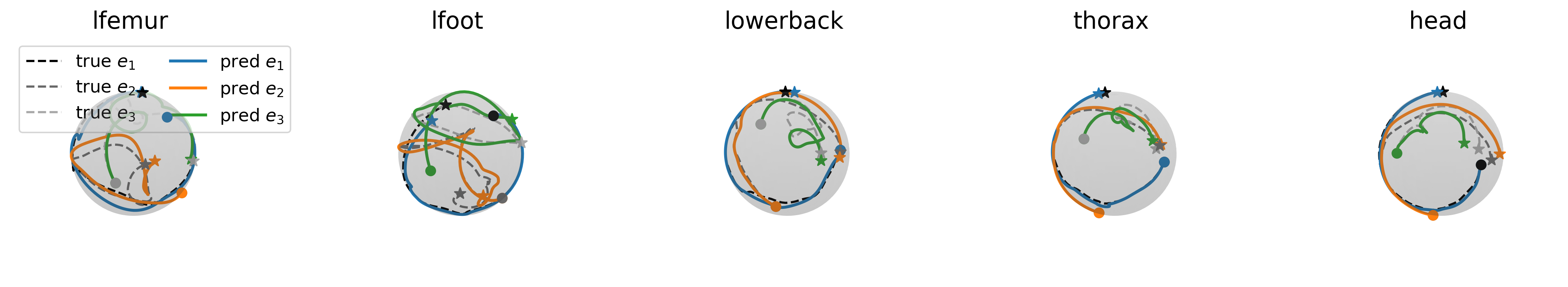}
        \caption{Sequence 1 (Trial~6).}
        \label{fig:mocap_traj0}
    \end{subfigure}\\
    \begin{subfigure}[c]{0.95\textwidth}
        \includegraphics[width=\linewidth, trim={0cm 2cm 0 3cm}, clip]{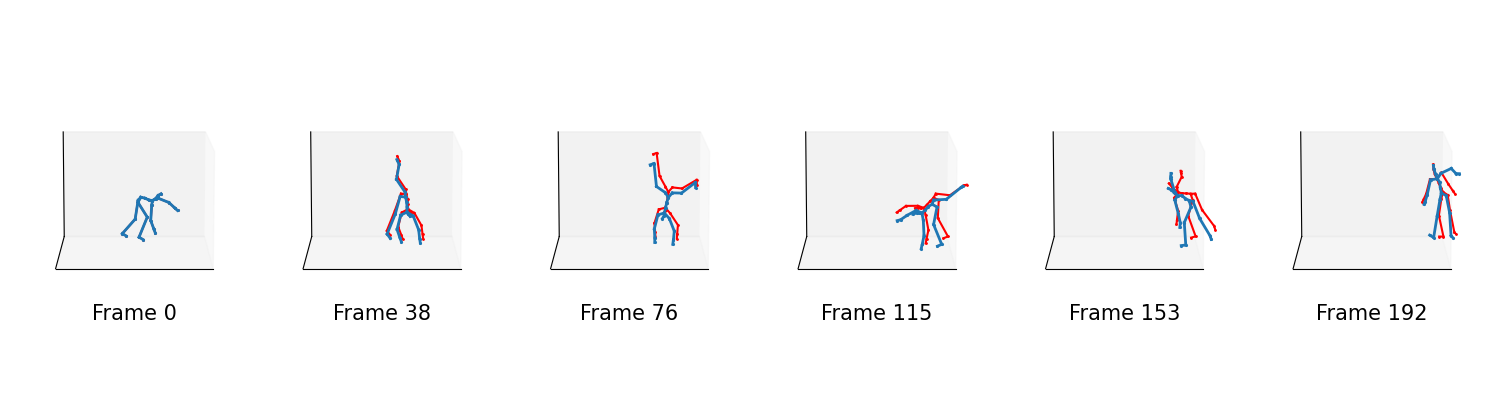} \\
        \includegraphics[width=0.3\linewidth]{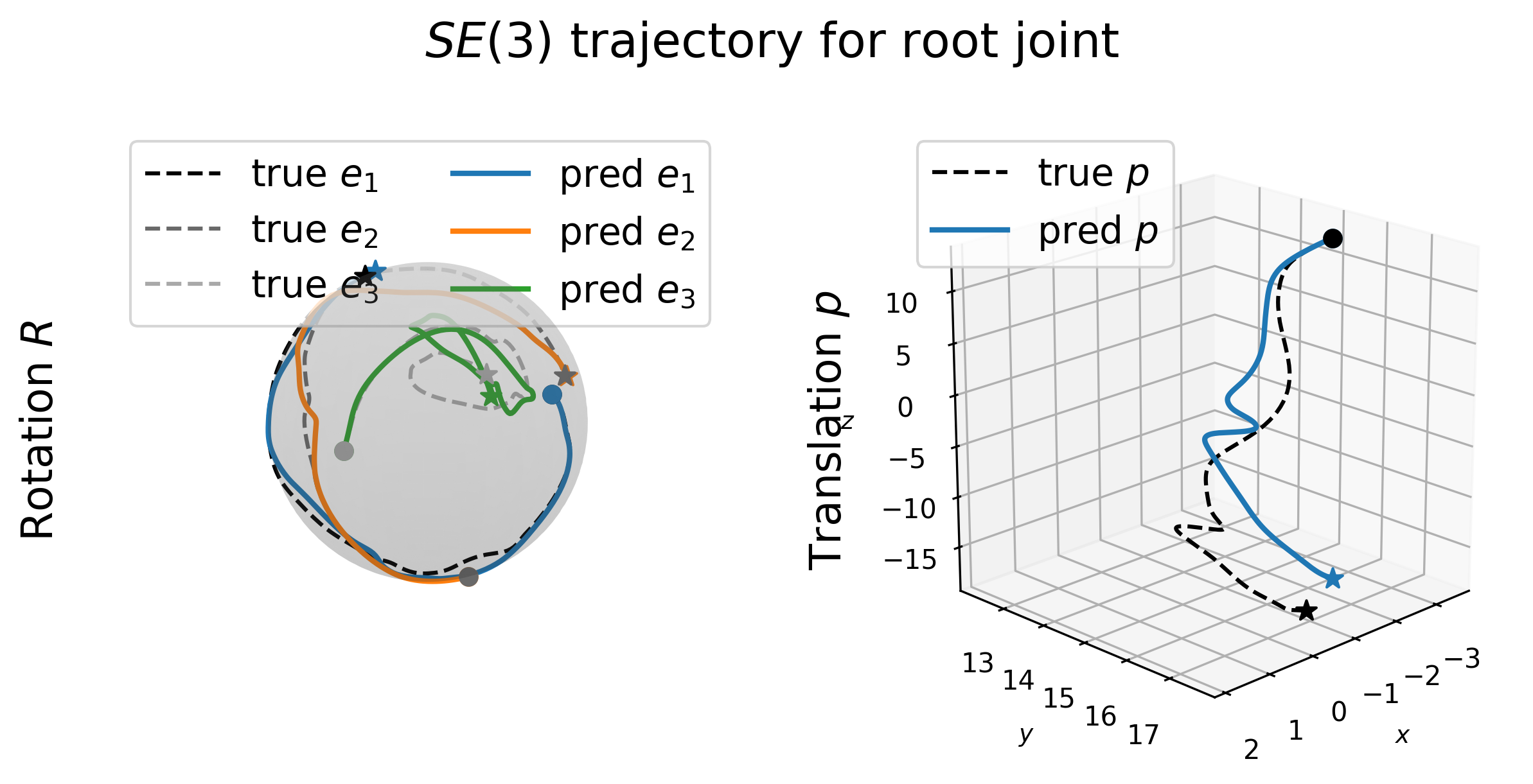}
        \hfill
        \includegraphics[width=0.73\linewidth]{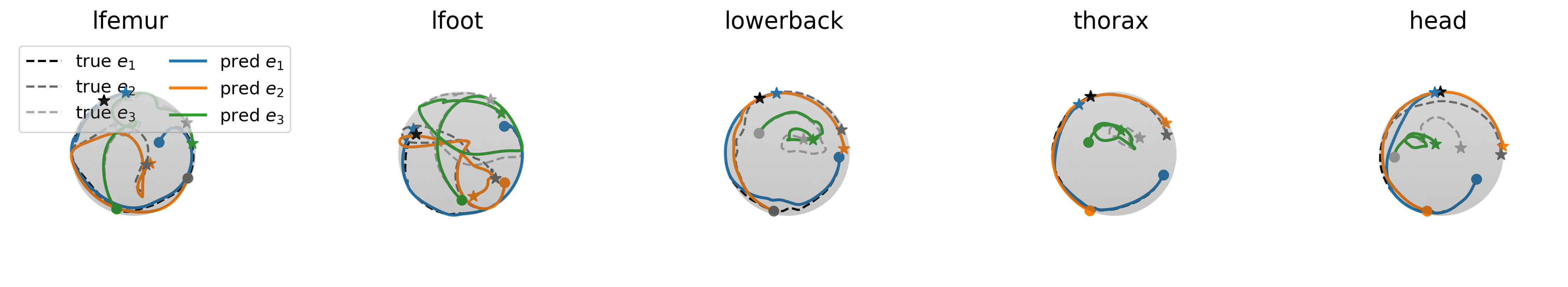}
        \caption{Sequence 2 (Trial~7).}
        \label{fig:mocap_traj1}
    \end{subfigure}
    \caption{Qualitative comparison for cartwheel sequences. 
    \Cref{fig:mocap_traj0}-\Cref{fig:mocap_traj1} Top: skeletal renderings at uniform intervals (ground truth: green/red, prediction: blue), showing nearly coincident poses throughout the $360^\circ$ rotation. \Cref{fig:mocap_traj0}-\Cref{fig:mocap_traj1} Bottom: reconstructed root trajectories in $SE(3)$ (left) and trajectories of selected joint rotation axes on the unit sphere, illustrating the high-fidelity reconstruction of $SO(3)$ dynamics (right). 
    }
    \label{fig:mocap_pred_trajectories}
\end{figure}

\Cref{tab:mocap_errors} and \Cref{fig:mocap_reproduction_error} report the quantitative reproduction errors. Across all metrics, per-sequence errors remain consistently below the inter-sequence ground-truth discrepancy. This confirms that the model captures the specific characteristics of each motion individually rather than a "blurred" average. As expected, the error peaks align with phases where the two demonstrations diverge most, marking the regions where the shared state-feedback force must mediate between the two actuation histories.

Qualitative results in \Cref{fig:mocap_pred_trajectories} show that the predicted and ground-truth skeletons remain nearly coincident throughout the $360^\circ$ rotation. The reconstructed joint trajectories on the unit sphere closely follow the recorded paths, with both sequences reproduced with comparable fidelity.

These results demonstrate that the LieDFLNN generalizes to real-world multibody systems without modifications to the model architecture. The discrete Lagrangian formulation naturally handles the product manifold $SE(3) \times SO(3)^{B-1}$, treating each joint group element independently while capturing coupled skeletal dynamics through a shared interaction potential. By operating directly on Lie groups, the model avoids the singularities and wrapping artifacts of Euler angles identified in Experiment~1. The close qualitative agreement of the predictions despite the very limited training data suggests that the inductive bias introduced by the discrete Lagrangian formulation together with the product Lie group structure serves as an effective regularizer, compensating for the small amount of training data. 

\subsection{Experiment 5: Learning an unknown control on $SO(3)$ with known Lagrangian}

The preceding experiments learned the discrete Lagrangian and the external forces jointly. In this experiment, we consider the complementary identification problem: the Lagrangian of the system is assumed known, and only the control input is learned. This scenario is relevant in practice when the mechanical properties of a system (inertia, potential) are available from design specifications, but the system is driven by an unknown or unmodeled feedback controller whose behavior one wishes to identify from observed motion. This setting also validates the framework's ability to isolate and recover individual dynamical components when others are fixed.

\paragraph{Setup and data generation.}
We consider a single rigid body evolving on $SO(3)$ in a gravitational field, using the Lagrangian from \Cref{example: L=T-V in SO3}, governing equations as in Experiment 1, and known parameters $J$, $J_d$ and $V$. The body is driven by a PD attitude controller \cite{lee2012robust} (see also \Cref{appendix: pd control in SO3})
\begin{equation}
    \label{eq: exp5 true PD control}
    \tau_{\mathrm{PD}}(R, \omega) = -K_R\, e_R(R, R_{\mathrm{f}}) - K_\omega\, \omega,
    \qquad
    e_R(R, R_{\mathrm{f}}) = \frac{1}{2} \vex\!\left(R_{\mathrm{f}}^\top R - R^\top R_{\mathrm{f}}\right),
\end{equation}
which drives the attitude toward a fixed target $R_{\mathrm{f}} \in SO(3)$. We set $K_R = k_r I$ and $K_\omega = k_\omega I$, scaling the gains by a
scalar representative inertia $\bar{J} = \tfrac{1}{N}\sum_{i,j} J_{ij}$
(the mean of the entries of $J$), with $k_r = 5.0\,\bar{J}$ and
$k_\omega = 1.0\,\bar{J}$. The training data consist of a \emph{single} trajectory of $N = 500$ timesteps with stepsize $h = 0.05$, generated by integrating Eq.~\eqref{eq:rigid_body_ode_SO3_potencial omega}--\eqref{eq:rigid_body_ode_SO3_potencial R} with $\tau = \tau_{\mathrm{cons}} + \tau_{\mathrm{PD}}$ from an initial condition $(R_0, \omega_0)$ using the RK45 solver from \texttt{scipy.integrate.solve\_ivp}. The rigid body has mass $m = 0.5$ and center of mass offset $\rho_\mathrm{com} = (0,\, 0.5,\, 1.0)^\top$, and diagonal inertia tensor
$J = \mathrm{diag}(1.521,\, 1.362,\, 1.211)$, under gravitational acceleration $g = 9.81$.
We retain only the rotation sequence $\{R_k\}_{k=0}^{N}$, neither angular velocities nor applied control inputs are observed.

We emphasize that learning from a single trajectory is a deliberately restricted setting. A state-feedback control can only be identified on the region of the state space visited by the observed motion. Accordingly, as in Experiment~4, the objective is faithful reproduction of the observed dynamics and recovery of the control along the trajectory, rather than generalization to unseen initial conditions.

\paragraph{Control parameterization and learning problem.}
Following the structure of Eq.~\eqref{eq:DEL2-4-bis}, we parameterize the unknown control as
\begin{equation}
    \label{eq: exp5 learned control}
    \tau^\theta(R, \omega) = C^\theta\, u^\theta(R, \omega),
\end{equation}
where $C^\theta \in \mathbb{R}^{3 \times m}$ is a learnable input matrix and $u^\theta : SO(3) \times \mathbb{R}^3 \to \mathbb{R}^m$ is a fully connected neural network taking the (flattened) rotation matrix and the angular velocity as inputs. The control is discretized in the same way as the external forces in \Cref{appendix: DEL SO3}, 
\begin{equation*}
    \tau^{\theta, \pm}_{d,k}
    = \frac{h}{2}\, \tau^\theta\!\left(R_k,\; \frac{\vex(W_k - W_k^\top)}{2h}\right),
\end{equation*}
where the angular velocity argument is recovered from the observed increment $W_k = R_k^\top R_{k+1}$, so that the model remains a function of positions only. Since the discrete Lagrangian is fixed, the loss \eqref{eq: loss general} reduces to the residual of the discrete controlled Euler-Lagrange equations \eqref{eq:DEL2-control} alone, evaluated on the observed triplets $(R_{k-1}, R_k, R_{k+1})$. No regularization is required, as the regularity of the Lagrangian is guaranteed by construction.
 
\begin{remark}[Identifiability of the factorization]
    \label{remark: exp5 identifiability}
    The factorization $\tau^\theta = C^\theta u^\theta$ is not unique; for any invertible $S \in \mathbb{R}^{m \times m}$, the pair $(C^\theta S,\, S^{-1} u^\theta)$ yields the same torque.
    Unlike settings where the control input $u$ is observed \cite{zhong2020symplectic}, here only the product $\tau^\theta$ is identifiable. We therefore evaluate the accuracy of the total torque $\tau^\theta$ rather than its individual factors.
    To remove the scale ambiguity during training, we normalize the columns of $C^\theta$ to have unit norm. 
\end{remark}

\begin{figure}[ht]
    \centering
    \hfill
    \begin{subfigure}[c]{0.30\textwidth}
        \includegraphics[width=0.8\linewidth]{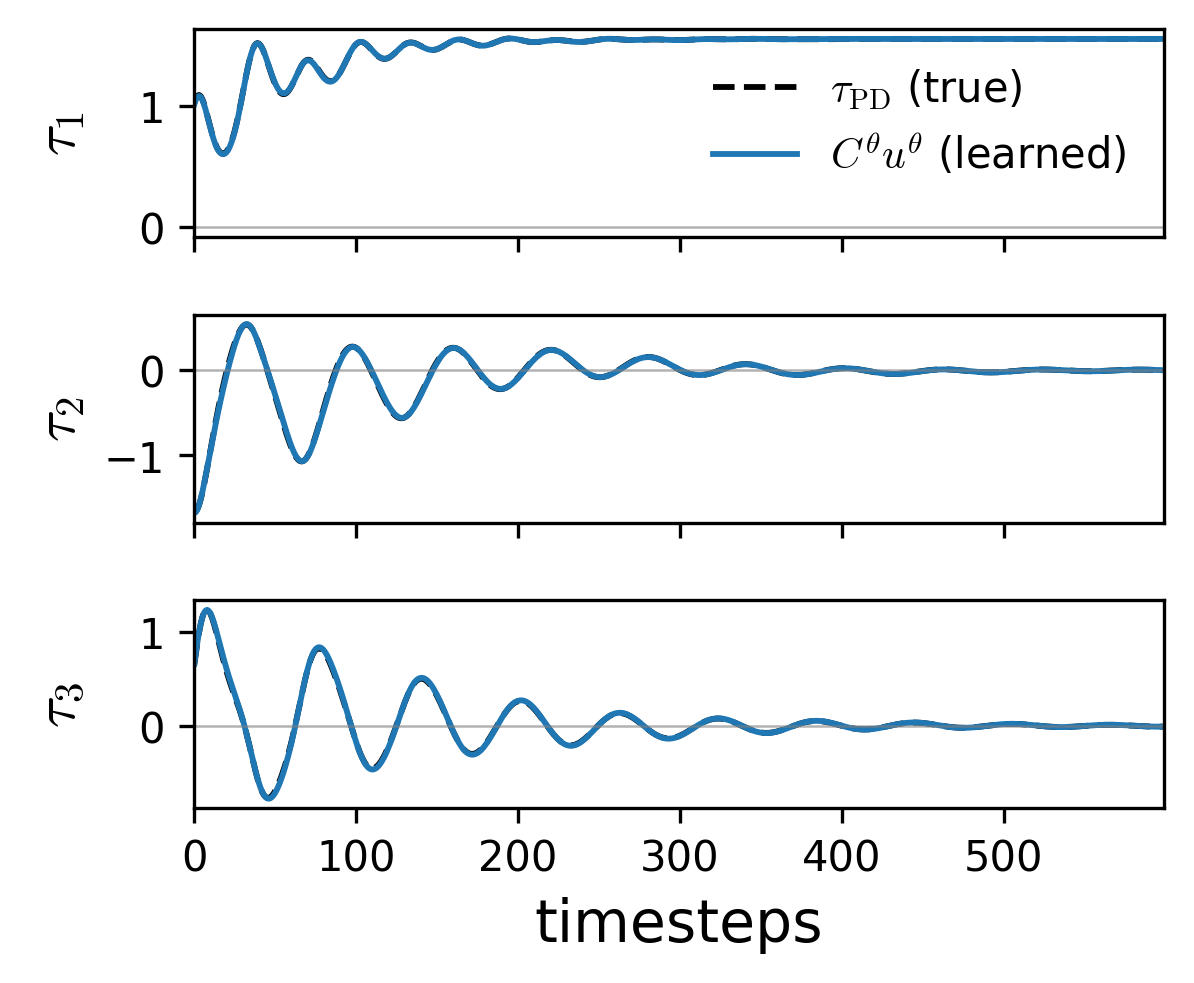}
        \caption{}
        \label{fig:control_learning components}
    \end{subfigure}
    \hfill
    \begin{subfigure}[c]{0.25\textwidth}
        \includegraphics[width=0.8\linewidth, trim={0.9cm 0cm 1cm 0cm}, clip]{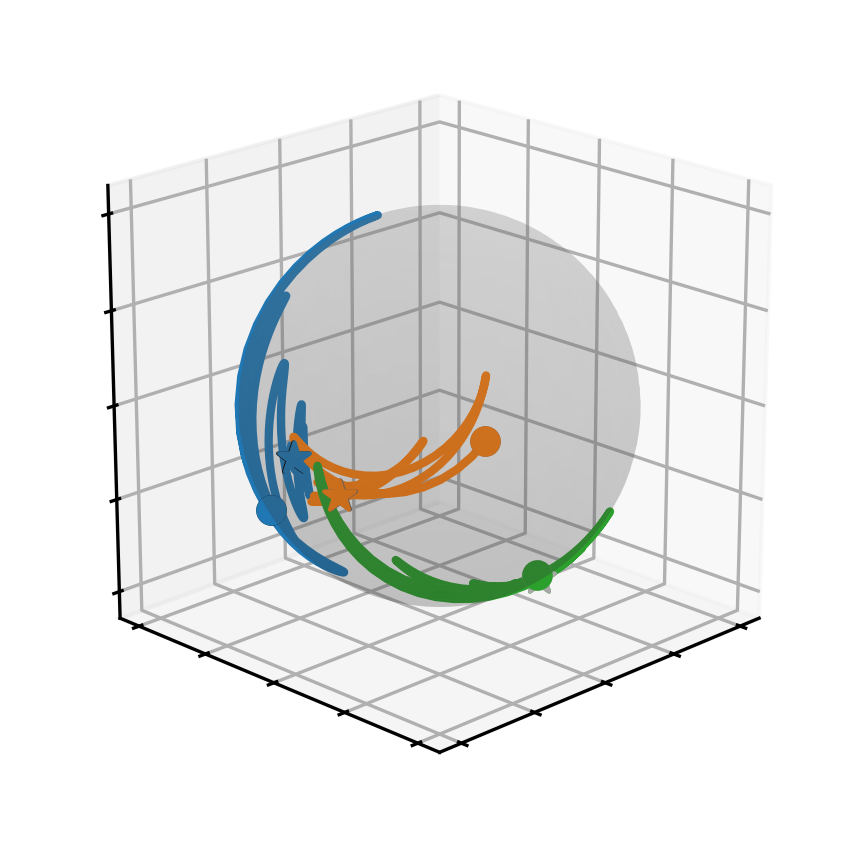}
        \caption{}
        \label{fig:control_learning rollout}
    \end{subfigure}
    \hfill
    \begin{subfigure}[c]{0.20\textwidth}
        \includegraphics[width=\linewidth]{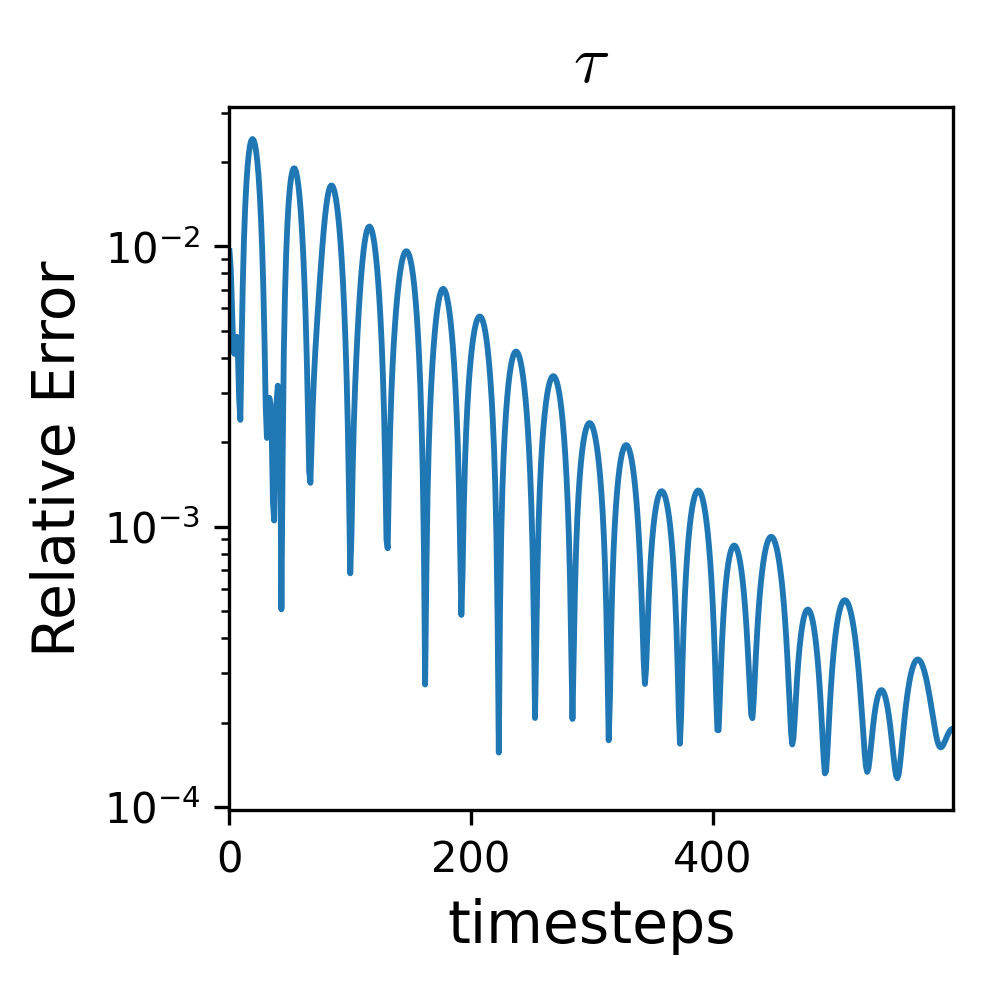}
        \caption{        }
        \label{fig:control_learning error}
    \end{subfigure}
    \hfill
    \begin{subfigure}[c]{0.20\textwidth}
        \includegraphics[width=\linewidth]{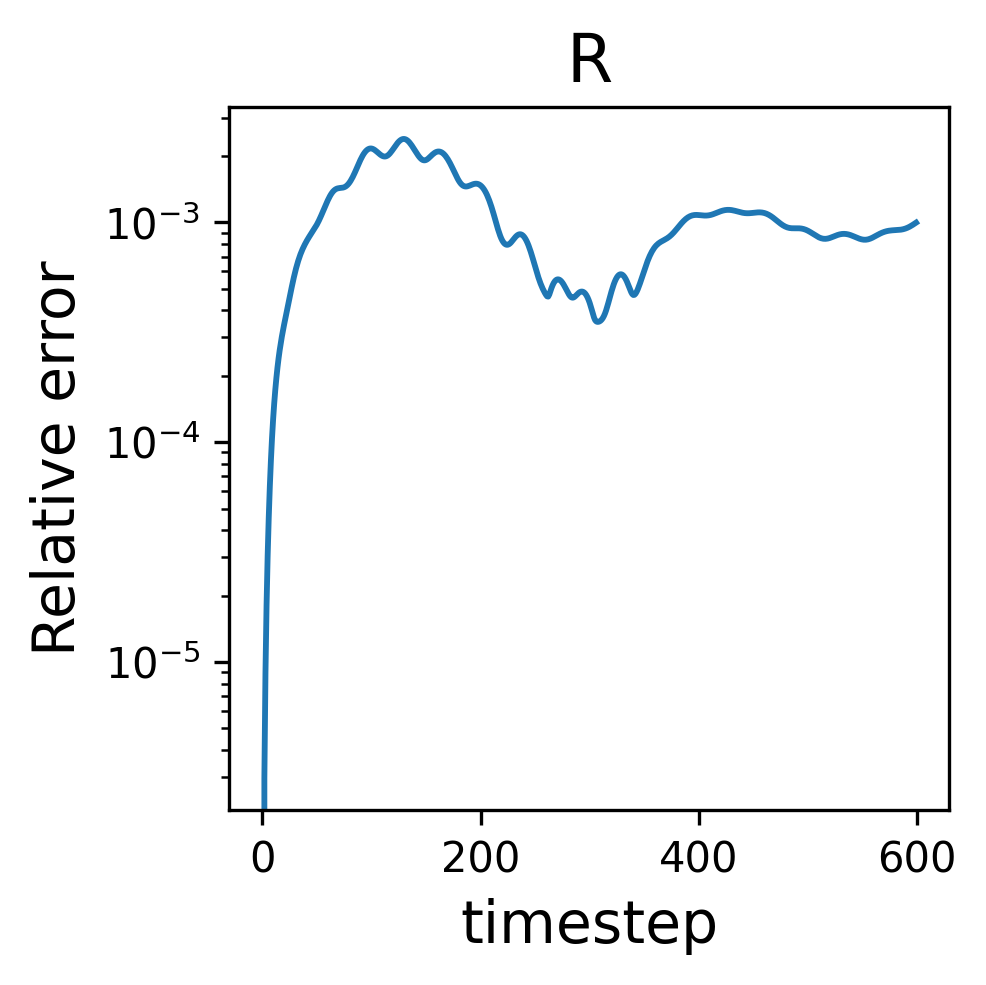}
        \caption{
        }
        \label{fig:control_learning rotation error}
    \end{subfigure}
    \caption{
    Identifying a PD controller on $SO(3)$ with a known Lagrangian. Results are obtained from a single 600-step trajectory. (\Cref{fig:control_learning components}, \Cref{fig:control_learning error}): the learned torque $\tau^\theta$ (solid) accurately recovers the true PD control signal (dashed) along the observed trajectory. (\Cref{fig:control_learning rollout}, \Cref{fig:control_learning rotation error}): rollout of the identified system showing faithful reproduction of the stabilizing motion toward the target $R_\mathrm{f}$, with errors remaining stable over the long-term horizon.
    }
    \label{fig:control_learning}
\end{figure}

\paragraph{Results and discussion.}
After training, the learned system is rolled out from the true initial pair $(R_0, R_1)$ by solving the discrete controlled Euler-Lagrange equations \eqref{eq:DEL2-control} forward in time with the known Lagrangian and the learned control $\tau^\theta$, as described in the inference procedure. \Cref{fig:control_learning} summarizes the results.

The learned control $\tau^\theta$ matches the true PD control signal closely (\Cref{fig:control_learning components}), capturing both the large transient torques at the start of the motion and the decay as the error $e_R$ vanishes. The relative control error (\Cref{fig:control_learning error}) remains low throughout the sequence, peaking slightly during the rapid initial transient.

The resulting rollout (\Cref{fig:control_learning rollout}) follows the ground-truth stabilizing trajectory on the sphere, maintaining a low relative rotation error over the full 600-step horizon (\Cref{fig:control_learning rotation error}).

Two aspects of this experiment are worth emphasizing. First, the discrete variational structure successfully disentangles the unknown control from the known conservative dynamics using only position data. The ability to recover the control signal without observing torques or velocities confirms the strength of the LieDFLNN as a tool for system identification in closed-loop settings.
Second, in line with \Cref{remark: exp5 identifiability}, we evaluate the identifiable product $C^\theta u^\theta$ rather than its factors. The factorization \eqref{eq: exp5 learned control} nevertheless provides the interface through which a known actuation structure could be imposed by fixing $C$, which we regard as a natural extension toward underactuated systems.

\section{CONCLUSIONS}
\label{sec:conclusion LieDFLNN}
In this work we have introduced a framework for learning mechanical systems using discrete forced Euler-Lagrange equations on Lie groups. By embedding the geometric structure of $SO(3)$ and $SE(3)$ directly into the variational integrator, we eliminate the coordinate singularities and geometric drift common in Euclidean baselines. Our experiments demonstrate that the LieDFLNN maintains high-fidelity predictions across large-angle rotations and long-term rollouts. The model recovers complex dynamics from position data alone and scales to high-dimensional multibody systems like human motion. Furthermore, the framework is modular, capable of identifying full dynamics or isolating specific components---such as unknown control laws---when the Lagrangian is known.

\section{ACKNOWLEDGMENTS}
The authors would like to express gratitude to Benjamin Kwanen Tapley and Jacob Goodman for their valuable discussions during the preparation of this paper. This research was supported by EU through MSCA-SE: REMODEL (Project ID: 101131557), and by the Research Council of Norway through PhysML (No. 338779). DMdD  acknowledges financial support from the Spanish Ministry of Science and Innovation under grants PID2022-137909NB-C21, PCI2024-155047-2 and from the Severo Ochoa Programme for Centres of Excellence in R\&D (CEX2023-001347-S). The data used in this project was obtained from mocap.cs.cmu.edu. The database was created with funding from NSF EIA-0196217.


\appendix

\section{BASIC NOTIONS ABOUT $SO(3)$ AND $SE(3)$}
\label{appendix:basic notions about SO3 and SE3}
\subsection{The Lie group $SO(3)$}
Consider the 3D rotation group $SO(3) = \{ R\in\mathbb{R}^{3\times 3} : R^\top R = R R^\top = I, \, \det(R) =1\} $. Its Lie algebra $\mathfrak{so}(3)=\{\Omega \in \mathbb{R}^{3\times 3} : \Omega^\top = -\Omega  \}$ can be identified with $\mathbb{R}^3$ via the hat map 
\begin{equation*}
    \hat{}:\mathbb{R}^3 \to \mathfrak{so}(3),\qquad
    \widehat{(a,b,c)}=\begin{pmatrix}
    0&-c&b\\
    c&0&-a\\
    -b&a&0
    \end{pmatrix},
\end{equation*}
and we let $\vex: \mathfrak{so}(3) \to \mathbb{R}^3 $ be its inverse. 

\paragraph{Left and Right multiplications in $SO(3)$.}
Let $R\in SO(3)$. 
The Left and Right multiplications by $R$ on $SO(3)$, $\mathcal{L}_R, \mathcal{R}_R : SO(3) \to SO(3)$ are
\begin{equation*}
    \mathcal{L}_{R}(Q) = RQ, \qquad
    \mathcal{R}_{R}(Q) = QR.
\end{equation*}
The tangent lifts at the identity of the Left and Right multiplications $T\mathcal{L}_R, T\mathcal{R}_R :\mathfrak{so}(3) \to T_R SO(3)$ are given by
\begin{equation*}
    T \mathcal{L}_R (\hat{\omega}) = \frac{d}{dt}\Bigg|_{t=0} \mathcal{L}_{R} Q(t) =  R\hat{\omega},
    \qquad
    T \mathcal{R}_R (\hat{\omega}) = \frac{d}{dt}\Bigg|_{t=0} \mathcal{R}_{R} Q(t) = \hat{\omega} R
\end{equation*}
where $Q(t)$ is a curve on $SO(3)$ such that $Q(0) = I_3$ and  $\Dot{Q}(0)= \hat{\omega} \in \mathfrak{so}(3)$. The cotangent lifts of the Left and Right multiplications $\mathcal{L}_R^*, \mathcal{R}_R^*: (T_R SO(3))^* \to \mathfrak{so}(3)^*$ are the maps such that 
\begin{equation*}
    \langle \mathcal{L}_R^* (\Pi), \hat{\omega}  \rangle =  \langle  \Pi, T \mathcal{L}_R (\hat{\omega})  \rangle,
    \qquad 
    \langle \mathcal{R}_R^* (\Pi), \hat{\omega}  \rangle =  \langle  \Pi, T \mathcal{R}_R (\hat{\omega})  \rangle
\end{equation*}
for any $\hat{\omega}\in \mathfrak{so}(3)$. Define the pairing $\langle \Pi, \hat{\omega} \rangle = \trace(\Pi^\top \hat{\omega})$, then one has
\begin{equation*}
    \mathcal{L}_{R}^* (\Pi) = R^\top \Pi, \qquad
    \mathcal{R}_{R}^* (\Pi) = \Pi R^\top.
\end{equation*}

\paragraph{Exp and Log maps in $SO(3)$.}
The exponential map $\exp^{SO} : \mathfrak{so}(3) \to SO(3)$ is given by
\begin{equation*}
    R = \exp^{SO}(\widehat{\omega}) := I + \frac{\sin\theta}{\theta} \widehat{\omega} + \frac{1-\cos\theta}{\theta^2} \widehat{\omega}^2, \qquad \theta := \| \omega \|.
\end{equation*}
Taking the trace on both sides
\begin{align*}
    \trace(R) = 3 + \frac{1-\cos\theta}{\theta^2} 2 \theta^2 
    \quad \Rightarrow \quad
    \theta = \cos^{-1} \left( \frac{\trace(R)-1}{2} \right).
\end{align*}
Also
\begin{align*}
    R - R^\top = \frac{2 \sin\theta}{\theta} \widehat{\omega}
    \quad \Rightarrow \quad
    \widehat{\omega} = \frac{\theta}{2\sin\theta} (R-R^\top),
\end{align*}
so $\log^{SO}: SO(3) \to \mathfrak{so}(3)$ is given by
\begin{equation*}
    \widehat{\omega} = \log^{SO}(R) := \frac{\theta}{2\sin\theta} (R-R^\top), \qquad 
    \theta := \cos^{-1} \left( \frac{\trace(R)-1}{2} \right).
\end{equation*}

\subsection{The Lie group $SE(3)$}
Consider the group of 3D rotations and translations $SE(3) \simeq SO(3) \times \mathbb{R}^3$ and its Lie algebra $\mathfrak{se}(3) \simeq \mathbb{R}^3 \times \mathbb{R}^3$. Let $g\in SE(3)$, $\xi\in \mathfrak{se}(3)$, then
\[ 
g=\begin{bmatrix}
    R & p \\ 0 & 1
\end{bmatrix}, \quad g^{-1}=\begin{bmatrix}
    R^\top & -R^\top p \\ 0 & 1
\end{bmatrix}, \quad\xi=\begin{bmatrix}
    \hat{\omega} & v \\ 0 & 0
\end{bmatrix}, \qquad R\in SO(3), \hat{\omega} \in \mathfrak{so}(3), p, v\in\mathbb{R}^3.
\]

\paragraph{Left and Right multiplications in $SO(3)$.}
The Left and Right multiplications by $g=(R, p)$ on $SE(3)$, $\mathcal{L}_g, \mathcal{R}_g : SE(3) \to SE(3)$ are
\begin{equation*}
    \mathcal{L}_{(R,p)}(Q,y) = (RQ, p+Ry), \qquad
    \mathcal{R}_{(R,p)}(Q,y) = (QR, y+Qp).
\end{equation*}
The tangent lifts of the Left and Right multiplications $T \mathcal{L}_g, T \mathcal{R}_g:\mathfrak{se}(3) \to T_g SE(3)$ are given by
\begin{equation*}
    T \mathcal{L}_g (\hat{\omega}, v) = \frac{d}{dt}\Bigg|_{t=0} \mathcal{L}_{(R, p)} (Q(t), y(t)) = 
    \frac{d}{dt}\Bigg|_{t=0} (RQ(t), p+ Ry(t)) = (R\hat{\omega}, Rv)
\end{equation*}
and
\begin{equation*}
    T \mathcal{R}_g (\hat{\omega}, v) = \frac{d}{dt}\Bigg|_{t=0} \mathcal{R}_{(R, p)} (Q(t), y(t)) = 
    \frac{d}{dt}\Bigg|_{t=0} (Q(t)R, Q(t)p+ y(t)) = (\hat{\omega} R, \hat{\omega} p + v)
\end{equation*}
where $(Q(t), y(t))$ is a curve on $SE(3)$ such that $(Q(0), y(0)) = (I_3,0)$ and  $(\Dot{Q}(0), \Dot{y}(0)) = (\hat{\omega},v) \in \mathfrak{se}(3)$. The cotangent lifts of the Left and Right multiplications $\mathcal{L}_g^*, \mathcal{R}_g^*: (T_g SE(3))^* \to \mathfrak{se}(3)^*$ are the maps such that 
\begin{equation*}
    \langle \mathcal{L}_g^* (\tau), \xi  \rangle =  \langle  \tau, T \mathcal{L}_g (\xi)  \rangle,
    \qquad 
    \langle \mathcal{R}_g^* (\tau), \xi  \rangle =  \langle  \tau, T \mathcal{R}_g (\xi)  \rangle.
\end{equation*}
for any $\xi\in \mathfrak{se}(3)$. For $\tau = (\Pi, F) \in \mathfrak{so}(3)^* \times \mathbb{R}^3$ and $\xi = (\hat{\omega}, v) \in \mathfrak{so}(3) \times \mathbb{R}^3$, let us define the pairing
\begin{equation*}
    \langle (\Pi, F), (\hat{\omega}, v) \rangle = \trace(\Pi^\top \hat{\omega}) + F^\top v.
\end{equation*}
One can then compute that
\begin{equation*}
    \mathcal{L}_{(R,p)}^* (\Pi, F) = (R^\top \Pi, R^\top F), \qquad
    \mathcal{R}_{(R,p)}^* (\Pi, F) = \left(\Pi R^\top - \frac{1}{2} \widehat{F \times p} , F \right).
\end{equation*}
Above we have used
\begin{equation*}
    F^\top (\hat{\omega} p ) = F \cdot (\omega \times p) = \omega \cdot (p \times F) = - \frac{1}{2} \trace ( \widehat{F \times p}^\top \hat{\omega} ).
\end{equation*}

\paragraph{Exp and Log maps in $SE(3)$.}
The exponential map is given by
\begin{equation*}
    \exp^{SE} : \mathfrak{se}(3) \to SE(3), \qquad
    (\omega, v) \mapsto (R, Av)
\end{equation*}
where, for $\theta := \| \omega \|$,
\begin{equation*}
    R = I + \frac{\sin\theta}{\theta} \widehat{\omega} + \frac{1-\cos\theta}{\theta^2} \widehat{\omega}^2, \qquad
    A = I + \frac{1-\cos{\theta}}{\theta^2} \hat{\omega} + \frac{\theta - \sin{\theta}}{\theta^3} \hat{\omega}^2.
\end{equation*}
Its inverse, namely the logarithmic map, is given by
\begin{equation*}
    \log^{SE} : SE(3) \to \mathfrak{se}(3), \qquad
    (R, p) \mapsto (\omega, A^{-1}p)
\end{equation*}
where, for $\theta := \cos^{-1} \left( \frac{\trace(R)-1}{2} \right)$,
\begin{equation*}
    \widehat{\omega} = \log^{SO}(R) := \frac{\theta}{2\sin\theta} (R-R^\top), \qquad 
    A^{-1} = I - \frac{1}{2} \hat{\omega} + \left( \frac{1}{\theta^2} - \frac{1+\cos{\theta}}{2\theta \sin{\theta}} \right) \hat{\omega}^2
\end{equation*}

\begin{remark}
    \label{remark: approximation of log map in SE3}
    In numerics, when $\theta \to 0$ one encounters division by zero. It is then desirable to replace the above close forms with e.g. Taylor approximations. More precisely, for small $\theta$ one has
    \begin{align*}
        \frac{\sin{\theta}}{\theta} &\approx 1 - \frac{\theta^2}{6}, \quad
        \frac{1-\cos{\theta}}{\theta^2} \approx \frac{1}{2} - \frac{\theta^2}{24} ,\quad
        \frac{\theta - \sin{\theta}}{\theta^3} \approx \frac{1}{6} - \frac{\theta^2}{120}, \quad
        \frac{1}{\theta^2} - \frac{1+\cos{\theta}}{2\theta\sin{\theta}} \approx \frac{1}{12} + \frac{\theta^2}{720},
    \end{align*}
    hence
    \begin{equation*}
        \exp^{SE}(\omega, v) \approx \left(  I + \hat{\omega} + \frac{1}{2} \hat{\omega}^2, \left( I + \frac{1}{2} \hat{\omega} + \frac{1}{6} \hat{\omega}^2 \right) v \right)
    \end{equation*}
    and
    \begin{equation*}
        \log^{SE}(R,p) \approx ({\omega}, v) = \left( \frac{3}{6-\theta^2} (R-R^\top), \left(  I - \frac{1}{2} \hat{\omega} + \frac{1}{12} \hat{\omega}^2 \right) p \right).
    \end{equation*}
    In particular, using the first order approximation for the exp map, we obtain 
    \begin{equation*}
        (W, z ) = \exp^{SE}(h \omega, h v) \approx \left( I + h \hat{\omega}, \left(I + \frac{1}{2} h \hat{\omega} \right) h v \right) 
    \end{equation*}
    from which one can compute 
    \begin{equation*}
        \hat{\omega} \approx \frac{W - I}{h}, \qquad
        v \approx \left(I + \frac{1}{2} h \hat{\omega} \right)^{-1} \frac{z}{h} \approx \left(I - \frac{1}{2} h \hat{\omega} \right) \frac{z}{h}
    \end{equation*}
    where we have used the Neumann series to approximate the inverse of a matrix. To make sure that $\hat{\omega}$ is skew symmetric one should take the skew symmetric part of $\frac{W-I}{h}$, leading to
    \begin{equation*}
        \hat{\omega} \approx \mathrm{skew} \left(\frac{W - I}{h}\right) = \frac{1}{2h} (W - W^\top).
    \end{equation*}
    The same is obtained using the first order approximation of the log map:
    \begin{equation*}
        (\hat{\omega}_k, z_k ) = \frac{1}{h} \log^{SE}(W_k, z_k) \approx \left( \frac{1}{2h} (W_k - W_k^\top), \left(I - \frac{1}{2} h \hat{\omega}_k \right) \frac{z_k}{h} \right).
    \end{equation*}
\end{remark}

\section{DETAILED DERIVATION OF THE DISCRETE EULER-LAGRANGE EQUATIONS  (DEL)}\label{appendix: detailed derivation of DEL}

\subsection{Discrete forced rigid body equations on $SO(3)$}
\label{appendix: DEL SO3}
Let $\eta\in\mathbb{R}^3$. 
The discrete Euler-Lagrange equation \eqref{eq:DEL2} rewrites
\begin{align*}
    \langle{\mathcal L}^*_{R_k} \frac{\partial \bar{L}_{d,k}}{\partial R_k}, \hat{\eta} \rangle &- \langle{\mathcal R}^*_{W_k} \frac{\partial \bar{L}_{d,k}}{\partial W_k}, \hat{\eta} \rangle + \langle{\mathcal L}^*_{W_{k-1}} \frac{\partial \bar{L}_{d,k-1}}{\partial W_{k-1}}, \hat{\eta} \rangle + \langle {\tau_{\mathrm{ext}}}^-_{d,k} , \eta \rangle + \langle {\tau_{\mathrm{ext}}}^+_{d,k-1}, \eta \rangle \\
    &= \trace\left( \left( R_k^\top \frac{\partial \bar{L}_{d,k}}{\partial R_k} -  \frac{\partial \bar{L}_{d,k}}{\partial W_k} W_k + W_{k-1} \frac{\partial \bar{L}_{d,k}}{\partial W_{k-1}} 
    + \frac{1}{2} (\widehat{{\tau_{\mathrm{ext}}}^-_{d,k}}+ \widehat{{\tau_{\mathrm{ext}}}^+_{d,k-1}})
    \right)^\top \hat{\eta}
    \right) \\
    &=: \trace ( A^\top \hat{\eta}) = 0.
\end{align*}
The above holds for any $\eta$ if and only if $A = A^\top$, that is \eqref{eq: DEL SO3 version with A}.

\subsection{Discrete forced rigid body equations on $SE(3)$}
\label{appendix: DEL SE3}
Let $\xi=(\hat{\eta},\mu)\in\mathfrak{se}(3)$. The discrete Euler-Lagrange equation \eqref{eq:DEL2} rewrites
\begin{align*}
    \langle{\mathcal L}^*_{(R_k,p_k)} & \left( \frac{\partial \bar{L}_{d,k}}{\partial R_k}, \frac{\partial \bar{L}_{d,k}}{\partial p_k} \right), \xi\rangle - \langle{\mathcal R}^*_{(W_k,z_k)} \left( \frac{\partial \bar{L}_{d,k}}{\partial W_k}, \frac{\partial \bar{L}_{d,k}}{\partial z_k} \right), \xi\rangle + \langle{\mathcal L}^*_{(W_{k-1},z_{k-1})} \left( \frac{\partial \bar{L}_{d,k-1}}{\partial W_{k-1}}, \frac{\partial \bar{L}_{d,k-1}}{\partial z_{k-1}} \right), \xi\rangle& \\
    =& \, \trace\left( \left( R_k^\top \frac{\partial \bar{L}_{d,k}}{\partial R_k} - \frac{\partial \bar{L}_{d,k}}{\partial W_k} W_k^\top + \frac{1}{2} \left( \frac{\partial \bar{L}_{d,k}}{\partial z_k} \times z_k \right)^\wedge + W_{k-1}^\top \frac{\partial \bar{L}_{d,k-1}}{\partial W_{k-1}} 
    + \frac{1}{2} (\widehat{{\tau_\mathrm{ext}}^-_{d,k}}+ \widehat{{\tau_\mathrm{ext}}^+_{d, k-1}})
    \right)^\top \hat{\eta}
    \right) \\
    & + \left(  R_k^\top \frac{\partial \bar{L}_{d,k}}{\partial p_k} - \frac{\partial \bar{L}_{d,k}}{\partial z_{k}} + W_{k-1}^\top \frac{\partial \bar{L}_{d,k-1}}{\partial z_{k-1}} 
    + ({f_\mathrm{ext}}^-_{d,k}+{f_\mathrm{ext}}^+_{d,k-1})
    \right)^\top \mu \\
    =:& \, \trace(A^\top \hat{\eta}) + b^\top \mu = 0.
\end{align*}
The above equation holds true for any $(\hat{\eta}, \mu)$ if and only if $A=A^\top$ and $b=0$, that is,
\begin{align*}
    R_k^\top \frac{\partial \bar{L}_{d,k}}{\partial R_k} - \frac{\partial \bar{L}_{d,k}}{\partial R_k}^\top R_k + W_k \frac{\partial \bar{L}_{d,k}}{\partial W_k}^\top - \frac{\partial \bar{L}_{d,k}}{\partial W_k} W_k^\top + \left( \frac{\partial \bar{L}_{d,k}}{\partial z_k} \times z_k \right)^\wedge  & \nonumber \\
    + W_{k-1}^\top \frac{\partial \bar{L}_{d,k-1}}{\partial W_{k-1}} - \frac{\partial \bar{L}_{d,k-1}}{\partial W_{k-1}}^\top W_{k-1} 
    + (\widehat{{\tau_\mathrm{ext}}^-_{d,k}}+ \widehat{{\tau_\mathrm{ext}}^+_{d, k-1}}) &= 0, \\
    R_k^\top \frac{\partial \bar{L}_{d,k}}{\partial p_k} - \frac{\partial \bar{L}_{d,k}}{\partial z_{k}} + W_{k-1}^\top \frac{\partial \bar{L}_{d,k-1}}{\partial z_{k-1}}
    + ({f_\mathrm{ext}}^-_{d,k}+{f_\mathrm{ext}}^+_{d,k-1}) &= 0,
\end{align*}
which correspond to equations \eqref{eq: DEL SE3 version with A}-\eqref{eq: DEL SE3 version with b}.

\begin{example}[Rigid body]
    \label{example: L=T-V in SE3}
    Let us consider the Lagrangian of a rigid body
    \begin{equation*}
        \bar{L} (R, p, \omega, v) = T(\omega, v) - V(R, p) 
        = \frac{1}{2} \omega^\top J \omega + \frac{1}{2} m (v + \hat{\omega} 
        \rho_{COM})^\top (v + \hat{\omega} 
        \rho_{COM}) - V(R, p).
    \end{equation*}
    where $J$ is the inertia matrix for vector representation, $m>0$ the mass, $\rho_{COM}\in\mathbb{R}^3$ the center of mass vector in the body frame, $v +\hat{\omega}\rho_{\mathrm{COM}} = \frac{\partial p}{\partial t} + \frac{\partial R}{\partial t}\rho_{\mathrm{COM}}$ the velocity of the center of the mass in the global frame ($p + R\rho_\text{COM}$ is its position).
    The discrete Lagrangian is
    \begin{align*}
        \bar{L}_{d,k} =& \, \bar{L}_d(R_k,p_k,W_k,z_k) = h \bar{L} \left(R_k, p_k, \frac{W_k - I}{h}, \frac{z_k
        }{h} \right)  \nonumber \\
        =&\, \frac{1}{h} (\trace(J_d) - \trace(W_k J_d)) + \frac{1}{2h} m (z_k + (W_k - I)\rho_{\mathrm{COM}})^\top  (z_k + (W_k - I)\rho_{\mathrm{COM}})
        - h V(R_k, p_k) \nonumber \\
    \end{align*}
    where $J_d$ satisfies $J = \trace({J_d}) I - J_d$ as noted in \cite{lee2005lie}.
    The partial derivatives of $\bar{L}_d$ with respect to $R$, $p$, $W$ and $z$ are
    \begin{align*}
         \frac{\partial \bar{L}_{d}}{\partial R} &= - h \frac{\partial V}{\partial R} , & \frac{\partial \bar{L}_{d}}{\partial W} &= - \frac{1}{h} J_d + \frac{1}{h} m z_k \rho_{COM}^\top -\frac{1}{h} m \rho_{COM} \rho_{COM}^\top,\\
         \frac{\partial \bar{L}_{d}}{\partial p} &= - h \frac{\partial V}{\partial p} , &
        \frac{\partial \bar{L}_{d}}{\partial z} & = \frac{1}{h} m z + \frac{1}{h} m (W_k -I)\rho_{COM}
    \end{align*}
    and we can replace them in  \eqref{eq: DEL SE3 version with A}-\eqref{eq: DEL SE3 version with b} to find
    \begin{align}
        \label{eq: DEL SE3 version with A example}
        -h  \left(R_k^\top \frac{\partial V_{k}}{\partial R}  - \left(\frac{\partial V_{k}}{\partial R}\right)^\top  R_k \right) 
        + \frac{1}{h} \left( J_d W_k^\top - W_k J_d \right)  
        + \frac{1}{h} \left( J_d W_{k-1} - W_{k-1}^\top J_d \right) & \nonumber   \\
         + \frac{1}{h} m \left( W_k \rho_{\mathrm{COM}} z_k^\top - z_k \rho_{\mathrm{COM}}^\top W_k^\top 
        + W_{k-1}^\top z_{k-1} \rho_{\mathrm{COM}}^\top - \rho_{\mathrm{COM}} z_{k-1}^\top W_{k-1} \right) & \nonumber \\
        + \frac{1}{h} m \left( \rho_{\mathrm{COM}} \rho_{\mathrm{COM}}^\top (W_k^\top + W_{k-1}) - (W_k + W_{k-1}^\top) \rho_{\mathrm{COM}} \rho_{\mathrm{COM}}^\top \right) & \nonumber \\
        + \frac{1}{h} m \left( (W_k - I) \rho_{\mathrm{COM}} \times z_k \right)^\wedge + (\widehat{{\tau_\mathrm{ext}}^-_{d,k}}+ \widehat{{\tau_\mathrm{ext}}^+_{d, k-1}}) &=0 \\
        \label{eq: DEL SE3 version with b example}
        - h R_k^\top \frac{\partial V_{k}}{\partial p} + \frac{1}{h} m \left( W_{k-1}^\top z_{k-1} - z_k + 2 I \rho_{\mathrm{COM}} - (W_{k-1}^\top + W_k) \rho_{\mathrm{COM}} \right) + ({f_\mathrm{ext}}^-_{d,k}+{f_\mathrm{ext}}^+_{d,k-1}) &= 0.
    \end{align}
    with $R_{k+1} = R_k W_k$, $p_{k+1} = p_k + R_k z_k $.
\end{example}

\subsection{Discrete forced equations on $SE(3) \times SO(3)$}
\label{appendix: DEL SE3 x SO3}
We consider two bodies and choose $SE(3) \times SO(3)$ as configuration space. 
The state and velocity of body 1 at timestep $k$ is given by $(R_{1,k}, p_{1,k}, \omega_{1,k}, v_{1,k}) \in SE(3) \times \mathfrak{se}(3)$, while for body 2 we have $(R_{2,k}, \omega_{2,k}) \in SO(3) \times \mathfrak{so}(3)$. The discrete Lagrangian $\bar{L}_d:SE(3) \times SE(3) \times SO(3) \times SO(3) \to \mathbb{R}$ is 
\begin{align*}
    \bar{L}_{d,k} &= \bar{L}_{d,1,k} + \bar{L}_{d,2,k} -  h V_{\mathrm{int},k}  \\
    &=
    \bar{L}_{d,1} (R_{1,k}, p_{1,k}, W_{1,k}, z_{1,k}) + \bar{L}_{d,2} (R_{2,k}, W_{2,k})
    - h V_\mathrm{int} (R_{1,k}, p_{1,k}, R_{2,k})
\end{align*}
where $\bar{L}_{d,1}: SE(3)\times SE(3) \to \mathbb{R}$ is the discrete Lagrangian for body $1$, $\bar{L}_{d,2}: SO(3)\times SO(3) \to \mathbb{R}$ is the discrete Lagrangian for body $2$, and we have assumed that the interaction potential $V_{\mathrm{int}} : SE(3) \times SO(3) \to \mathbb{R} $ depends on positions only. For readability, we also assume that there are no external forces. The system of equations corresponding to the Euler-Lagrange equation \eqref{eq:DEL2} is then
\begin{align}
    \label{eq: general discrete EL equations in SE(3)xSO(3) A}
    R_{1,k}^\top \left( \frac{\partial \bar{L}_{d,1,k}}{\partial R_{1,k}} + \frac{\partial V_{\mathrm{int},k}}{\partial R_{1,k}} \right) - \left( \frac{\partial \bar{L}_{d,1,k}}{\partial R_{1,k}}  + \frac{\partial V_{\mathrm{int},k}}{\partial R_{1,k}} \right)^\top R_{1,k}  + \left( \frac{\partial \bar{L}_{d,1,k}}{\partial z_{1,k}} \times z_{1,k} \right)^\wedge & \nonumber \\
    + W_{1,k} \frac{\partial \bar{L}_{d,1,k}}{\partial W_{1,k}}^\top - \frac{\partial \bar{L}_{d,1,k}}{\partial W_{1,k}} W_{1,k}^\top + W_{1,k-1}^\top \frac{\partial \bar{L}_{d,1,k-1}}{\partial W_{1,k-1}} - \frac{\partial \bar{L}_{d,1,k-1}}{\partial W_{1,k-1}}^\top W_{1,k-1} &= 0, \\
    \label{eq: general discrete EL equations in SE(3)xSO(3) B}
    R_{2,k}^\top \left( \frac{\partial \bar{L}_{d,2,k}}{\partial p_{2,k}} + \frac{\partial V_{\mathrm{int},k}}{\partial p_{2,k}} \right)- \frac{\partial \bar{L}_{d,2,k}}{\partial z_{2,k}} + W_{2,k-1}^\top \frac{\partial \bar{L}_{d,2,k-1}}{\partial z_{2,k-1}} &= 0, \\
    \label{eq: general discrete EL equations in SE(3)xSO(3) C}
     R_{2,k}^\top \left( \frac{\partial \bar{L}_{d,2,k}}{\partial R_{2,k}} - \frac{\partial \bar{L}_{d,2,k}}{\partial R_{2,k}}^\top \right) - 
     \left( \frac{\partial \bar{L}_{d,2,k}}{\partial W_{2,k}} - \frac{\partial \bar{L}_{d,2,k}}{\partial W_{2,k}}^\top \right) W_{2,k} 
     + W_{2,k-1} \left( \frac{\partial \bar{L}_{d,2,k}}{\partial W_{2,k-1}} - \frac{\partial \bar{L}_{d,2,k}}{\partial W_{2,k-1}}^\top \right) &= 0,
\end{align}
with $R_{1,k+1} = R_{1,k} W_{1,k} $, $p_{1,k+1} = p_{1,k} + R_{1,k} z_{1,k} $ and $R_{2,k+1} = R_{2,k} W_{2,k} $.

\section{REGULAR LAGRANGIAN}
\label{Appendix: regular Lagrangian}
\paragraph{Discrete regular Lagrangian.}
Let us consider the discrete Lagrangian of a rigid body evolving on $SE(3)$ (see \Cref{example: L=T-V in SE3})
\begin{align*}   \bar{L}_d (R_k, p_k, W_k, z_k) &= hL \left(R_k, p_k, \frac{W_k -I}{h}, \frac{z_k}{h} \right) \\
    &= \frac{1}{h} \trace((I-W_k)J_d) + \frac{m}{2h} \| X_k \|^2 - hV_d(p_k,R_k),
\end{align*}
where $X_k = z_k + (W_k-I)\rho_{\mathrm{COM}}$, $J_d$ is such that $J = \trace(J_d)I - J_d$, $J$ is the inertia matrix, $m$ is the mass and $\rho_\mathrm{COM}$ the position of the center of mass ($\mathrm{COM}$). 
Let $g_k=(R_k,p_k)$ and $P_k=(W_k,z_k)=g_k^{-1}g_{k+1}$, so that $L_d(g_k,g_{k+1})=\bar L_d(g_k,P_k)$. By \Cref{section: Lie group DFLNN}, $L_d$ is regular if the mixed Hessian $\mathbb H(L_d)(g_k,g_{k+1})$ is non-singular. The potential term depends on $g_k$ only and does not contribute to a mixed derivative. Moreover, $(g_k\exp^{SE} (s\xi))^{-1}g_{k+1}=\exp^{SE}(-s\xi)P_k=P_k\exp^{SE} (-s\,\mathrm{Ad}_{P_k^{-1}}\xi)$, hence
\begin{equation*}
    \mathbb H(L_d)(g_k,g_{k+1}) = -\,\mathbb A_k^\top\,\mathbb H_{\mathrm{inc}},
    \qquad
    (\mathbb H_{\mathrm{inc}})_{ij}=\left.\frac{\partial^2}{\partial s\,\partial t}\right|_{s=t=0}
    \bar L_d\big(g_k,\,P_k\exp^{SE}(s e_i)\exp^{SE}(t e_j)\big),
\end{equation*}
where $\mathbb A_k$ is the matrix of $\mathrm{Ad}_{P_k^{-1}}$ in the basis $\{e_i\}$. Since $\det\mathbb A_k=1$, we have $|\det\mathbb H(L_d)|=|\det\mathbb H_{\mathrm{inc}}|$, so it suffices to compute $\mathbb H_{\mathrm{inc}}$, the Hessian of $\bar L_d(g_k,\cdot)$ with respect to the increments, taken along left-invariant directions of $SE(3)$. Up to transposition, which does not affect the determinant, it is the block matrix
\begin{equation*}
    \mathbb{H} = \begin{bmatrix}
        \mathbb{H}_{zz} & \mathbb{H}_{zW} \\ \mathbb{H}_{Wz} & \mathbb{H}_{WW}
    \end{bmatrix}
\end{equation*}
computed below.

\paragraph{First variations (discrete momenta).} 
The discrete momenta are the first variation of $\bar{L}_d$ with respect to a variation of $P_k$ along a left-invariant direction of $SE(3)$, namely $P_k\mapsto P_k\exp^{SE}(\epsilon(\hat\eta,\mu))$ with $\eta,\mu\in\mathbb R^3$. To first order,
\begin{equation*}
    (W_k, z_k) \mapsto \left(W_k \exp^{SO}(\epsilon\hat{\eta}),\; z_k+\epsilon W_k\mu\right).
\end{equation*}
For the variation with respect to $\mu$ we obtain
\begin{equation*}
    \delta_z \bar{L}_d = \frac{\partial \bar{L}_d}{\partial z_k}\cdot (W_k\mu)
    = M_{k,z}\cdot\mu, \qquad
    M_{k,z} := W_k^\top \frac{\partial \bar{L}_d}{\partial z_k} = \frac{m}{h} W_k^\top X_k .
\end{equation*}
For the variation with respect to $\eta$, $z_k$ is fixed and $W_k$ is perturbed as
\begin{equation*}
    W_k (\epsilon) = W_k \exp^{SO}(\epsilon\hat{\eta}).
\end{equation*}
The first variation (discrete momentum) is given by
\begin{align*}
    \delta_W \bar{L}_d &=
    \left. \frac{d}{d\epsilon} \right|_{\epsilon=0}  \bar{L}_d (R_k, p_k, W_k \exp^{SO}(\epsilon \hat{\eta}), z_k) \\
    & = \left. \frac{d}{d\epsilon} \right|_{\epsilon=0} \frac{1}{h} \trace((I - W_k \exp^{SO}(\epsilon\hat{\eta})) J_d) +
    \left. \frac{d}{d\epsilon} \right|_{\epsilon=0} \frac{m}{2h} \| z_k + (W_k\exp^{SO}(\epsilon \hat{\eta})-I) \rho_{\mathrm{COM}} \|^2
\end{align*}
For the first term we have
\begin{align*}
    \left. \frac{d}{d\epsilon} \right|_{\epsilon=0} \frac{1}{h} \trace((I - W_k \exp^{SO}(\epsilon\hat{\eta})) J_d) & = -\frac{1}{h} \trace (W_k \hat{\eta} J_d)
    = -\frac{1}{h} \trace (J_d W_k \hat{\eta} ) \\
    & =  \frac{1}{h} \vex ( J_d W_k - W_k^\top J_d) \cdot \eta =: M_{k,\mathrm{rot}} \cdot \eta
\end{align*}
where we have used the property $\left. \frac{d}{d\epsilon} \right|_{\epsilon=0} \exp^{SO}(\epsilon \hat{\eta}) = \hat{\eta}$ and the trace identity $\trace(A \hat{b}) = \vex(A^\top -A) \cdot b$ with $\vex$ the inverse of the hat operator.\\
For the second term
\begin{align*}
    \left. \frac{d}{d\epsilon} \right|_{\epsilon=0} \frac{m}{2h} \| z_k + (W_k\exp^{SO}(\epsilon \hat{\eta})-I) \rho_{\mathrm{COM}} \|^2 =
    \frac{m}{h} X_k \cdot (W_k \hat{\eta} \rho_\mathrm{COM}) = - \frac{m}{h} X_k \cdot (W_k \hat{\rho}_\mathrm{COM} \eta)
\end{align*}
where we have used $\hat{v}w = v \times w = - w \times v = - \hat{w}v$. Now
\begin{equation*}
    X_k \cdot (W_k \hat{\rho}_\mathrm{COM} \eta) = (W_k \hat{\rho}_\mathrm{COM} \eta) \cdot X_k
    = (W_k \hat{\rho}_\mathrm{COM} \eta)^\top X_k = \eta^\top  \hat{\rho}_\mathrm{COM}^\top W_k^\top X_k = - \eta \cdot ( \hat{\rho}_\mathrm{COM} W_k^\top X_k)
\end{equation*}
hence we get
\begin{equation*}
    \delta_W \bar{L}_d = \left( M_{k,\mathrm{rot}} + \frac{m}{h} \hat{\rho}_\mathrm{COM} W_k^\top X_k \right) \cdot \eta =: \left( M_{k,\mathrm{rot}} + M_{k,\mathrm{COM}} \right) \cdot \eta.
\end{equation*}

\paragraph{Second variations.}
To find the components of $\mathbb{H}$ we need the variations of the momenta $M_{k,z}$ and $M_{k,W}:=M_{k,\mathrm{rot}}+M_{k,\mathrm{COM}}$ with respect to a second variation $(\mu, \zeta)$, $\mu,\zeta \in \mathbb{R}^3$, of the same type as above. For the variation of $M_{k,z}$ with respect to $\mu$, and of $M_{k,W}$ with respect to $\mu$ (note that $M_{k,\mathrm{rot}}$ does not depend on $z_k$), we have
\begin{align*}
    \mathbb{H}_{zz}\mu &= \left. \frac{d}{d\epsilon} \right|_{\epsilon=0} \frac{m}{h} W_k^\top \big(X_k+\epsilon W_k\mu\big)
    = \frac{m}{h} \mu, \\
    \mathbb{H}_{Wz}\mu &= \left. \frac{d}{d\epsilon} \right|_{\epsilon=0} \frac{m}{h} \hat{\rho}_\mathrm{COM} W_k^\top \big(X_k+\epsilon W_k\mu\big) = \frac{m}{h} \hat{\rho}_\mathrm{COM} \mu,
\end{align*}
that is, $\mathbb{H}_{zz}=\frac{m}{h} I_{3\times3}$ and $\mathbb{H}_{Wz}=\frac{m}{h} \hat{\rho}_\mathrm{COM}$.
For the variation with respect to $\zeta$ (that is, $W_k\mapsto W_k\exp^{SO}(\epsilon\hat\zeta)$ with $z_k$ fixed) we proceed as earlier. For $\mathbb{H}_{zW}$ we compute
\begin{align*}
    \left. \frac{d}{d\epsilon} \right|_{\epsilon=0} \frac{m}{h} \exp^{SO}(-\epsilon\hat{\zeta}) W_k^\top \big(z_k + (W_k\exp^{SO}(\epsilon \hat{\zeta})-I)\rho_{\mathrm{COM}}\big)
    &= \frac{m}{h} \big(-\hat{\zeta} W_k^\top X_k + \hat{\zeta}\rho_{\mathrm{COM}}\big) \\
    &= \frac{m}{h} \big(\widehat{W_k^\top X_k} - \hat{\rho}_{\mathrm{COM}}\big)\zeta =: \mathbb{H}_{zW} \zeta .
\end{align*}
For $\mathbb{H}_{WW}$ we need
\begin{align*}
    \delta_W M_{k,\mathrm{rot}} &= \left. \frac{d}{d\epsilon} \right|_{\epsilon=0}  \frac{1}{h} \vex \left( J_d W_k \exp^{SO}(\epsilon\hat{\zeta}) - (W_k \exp^{SO}(\epsilon\hat{\zeta} ) )^\top J_d \right) \\
    &= \frac{1}{h} \vex \left( J_d W_k \hat{\zeta} + \hat{\zeta} W_k^\top J_d \right)
    = \frac{1}{h} (\trace( W_k^\top J_d) I - W_k^\top J_d) \zeta =: \frac{1}{h}\mathbb{H}_{\mathrm{rot}} \zeta
\end{align*}
where we have used the identity $\vex(A^\top \hat{\zeta} + \hat{\zeta}A) = (\trace(A) I - A)\zeta$, and
\begin{align*}
    \delta_W M_{k,\mathrm{COM}} &= \left. \frac{d}{d\epsilon} \right|_{\epsilon=0} \frac{m}{h} \hat{\rho}_\mathrm{COM} (W_k \exp^{SO}(\epsilon\hat{\zeta}))^\top (z_k + (W_k \exp^{SO}(\epsilon\hat{\zeta}) - I) \rho_{\mathrm{COM}}) \\
    & = \frac{m}{h} \hat{\rho}_\mathrm{COM} (-\hat{\zeta} W_k^\top X_k + \hat{\zeta} \rho_{\mathrm{COM}}) = \frac{m}{h} \hat{\rho}_\mathrm{COM} (\widehat{W_k^\top X_k} - \hat{\rho}_{\mathrm{COM}}) \zeta  =: \frac{1}{h} \mathbb{H}_{\mathrm{COM}} \zeta,
\end{align*}
so that $\mathbb{H}_{WW} =\frac{1}{h} ( \mathbb{H}_{\mathrm{rot}} + \mathbb{H}_{\mathrm{COM}}) $.

\paragraph{Condition for regularity.} 
We now impose $\det(\mathbb{H}) \ne 0$. We use the Schur complement for a block matrix $\begin{bmatrix} A & B \\ C & D \end{bmatrix}$, so that
\begin{equation*}
    \det(\mathbb{H}) = \det(A) \det(D-CA^{-1}B).
\end{equation*}
In our case we have
\begin{equation*}
    \det(A) = \det(\mathbb{H}_{zz}) = \left(\frac{m}{h}\right)^3 \ne 0
\end{equation*}
as long as $m>0$. Then
\begin{equation*}
    CA^{-1}B = \mathbb{H}_{Wz} \mathbb{H}_{zz}^{-1} \mathbb{H}_{zW} =
    \left( \frac{m}{h} \hat{\rho}_\mathrm{COM} \right) \frac{h}{m} I \left( \frac{m}{h} \big(\widehat{W_k^\top X_k} - \hat{\rho}_{\mathrm{COM}}\big) \right)
    = \frac{1}{h}\mathbb{H}_{\mathrm{COM}},
\end{equation*}
so that
\begin{equation*}
    D - CA^{-1}B = \mathbb{H}_{WW} - \frac{1}{h}\mathbb{H}_{\mathrm{COM}} = \frac{1}{h} \mathbb{H}_{\mathrm{rot}}.
\end{equation*}
The final condition is then
\begin{equation*}
    \det(\mathbb{H}) = \frac{m^3}{h^6}\,\det\left( \trace( W_k^\top J_d) I - W_k^\top J_d \right) \ne 0.
\end{equation*}

\begin{remark}
The regularity condition does not depend on $\rho_\mathrm{COM}$, on $z_k$ or on the potential $V$. It reduces to $\det(\trace(W_k^\top J_d) I - W_k^\top J_d)\neq 0$, which is also the condition obtained when the frame is located at the center of mass, where the discrete Lagrangian simplifies to
\begin{align*}
    \bar{L}_d (R_k, p_k, W_k, z_k) = \frac{1}{h} \trace((I-W_k)J_d) + \frac{m}{2h} \|z_k\|^2 - h V(R_k, p_k).
\end{align*}
\end{remark}

\section{DETAILED EXPERIMENTS SET-UP}

\subsection{Implementation and training details}
\label{appendix: implementation details}

All models are implemented in PyTorch~\cite{paszke2019pytorch} and trained on CPU. The neural networks parameterizing the discrete Lagrangian and
forces are multilayer perceptrons with GELU activations, their widths and depths
are listed in \Cref{tab:hyperparams-arch}, and training parameters in \Cref{tab:hyperparams-train}. All results presented for LieDFLNN have been trained using the non-smooth regularization option $\mathrm{loss}_\mathrm{reg1}$.

\begin{table}[ht]
    \centering
    \small
    \setlength{\tabcolsep}{4pt}
    \caption{Network architecture for the proposed model and all baselines.
    ``MLP'' gives the width\,$\times$\,depth of the learned networks (the potential
    $V^\theta$, control $u^\theta$, general force $F^\theta_\mathrm{NN}$, nonlinear
    dissipation force $F^\theta_\mathrm{NL}$, or the Neural ODE vector field
    $f^\theta$). All MLPs use GELU activations. ``Reg.'' denotes the number of regularization points used when relevant. }
    \label{tab:hyperparams-arch}
    \vspace{1mm}
    \begin{tabular}{*{6}{|c}|}
        \hline
        Exp. & Model & Input & Learned ($\theta$) & MLP & Reg. \\
        \hline
        \multirow{2}{*}{1}
          & LieDFLNN (ours) & $R$ (9) & $J_d^\theta,D^\theta,V^\theta$ & $64\times3$ & 10\\
          & Euclidean DFLNN w/Euler-angles      & $(\phi,\theta,\psi)$ (3) & $J_d^\theta,D^\theta,V^\theta$ & $64\times3$ & 100\\
        \hline
        \multirow{4}{*}{2}
          & LieDFLNN (ours) & $R$ (9)   & $J_d^\theta,D^\theta$ & -- & 10 \\
          & GLNN w/RK4             & $R,\omega$ & $J_d^\theta,D^\theta$ & -- & --  \\
          & GLNN w/Lie-Midpoint    & $R,\omega$ & $J_d^\theta,D^\theta$ & -- & --  \\
          & GLNN w/Lie-Heun        & $R,\omega$ & $J_d^\theta,D^\theta$ & -- & -- \\
        \hline
        \multirow{3}{*}{3}
          & LieDFLNN (ours) & $(R,p)$ (12) & $J_d^\theta,D^\theta,\rho^\theta,V^\theta$ & $32\times3$ & 10\\
          & Euclidean DFLNN       & $(R,p)$ (12) & $J_d^\theta,D^\theta,\rho^\theta,V^\theta$ & $32\times3$ & 10\\
          & Neural ODE             & $(R,p)$ (12) & $f^\theta$                                 & $64\times3$ & -- \\
        \hline
        4 & LieDFLNN (ours) & joints &
          \makecell[l]{$J_d^\theta,D^\theta,\rho^\theta,V^\theta,$\\$V_\mathrm{int}^\theta,F^\theta_\mathrm{NN},F^\theta_\mathrm{NL}$}
          & $32\times3$ & 10\\
        \hline
        5 
            & LieDFLNN (ours) & $R$ (9) & $C^\theta,u^\theta$ & $64\times3$ & --\\
        \hline
    \end{tabular}
\end{table}

\begin{table}[ht]
    \centering
    \small
    \setlength{\tabcolsep}{4pt}
    \caption{Training settings for the proposed model and all baselines.
    ''Window'' is the number of poses per backpropagated segment, ''Traj.'' the full data-sequence length. All models are trained with the SOAP optimizer at learning rate
    $3\times10^{-3}$.}
    \label{tab:hyperparams-train}
    \vspace{1mm}
    \begin{tabular}{*{6}{|c}|}
        \hline
        Exp. & Model & Batch & Window & Traj. & Epochs \\
        \hline
        \multirow{2}{*}{1}
          & LieDFLNN (ours) & 1024 & 3 & 100 & $2.500$ \\
          & Euclidean DFLNN w/Euler-angles     & 1024 & 3 & 100 & $2.500$ \\
        \hline
        \multirow{4}{*}{2}
          & LieDFLNN (ours) & 256/512/1024/1024 & 3    & 10/40/70/100 & $1000$ \\
          & GLNN w/RK4             & 8/16/32/32        & full & 10/40/70/100 & $2.000$ \\
          & GLNN w/Lie-Midpoint    & 256/512/1024/1024 & 3    & 10/40/70/100 & $1.000$ \\
          & GLNN w/Lie-Heun        & 8/16/32/32        & full & 10/40/70/100 & $500$ \\
        \hline
        \multirow{3}{*}{3}
          & LieDFLNN (ours) & 1024 & 3    & 32 & $5.000$ \\
          & Euclidean DFLNN        & 1024 & 3   & 32     & $5.000$ \\
          & Neural ODE             & 1024 & 3 & 32     & $5.000$ \\
        \hline
        4 & LieDFLNN (ours) & 63 & 3 & 2 seq. & $10.000$ \\
        \hline
        5 
            & LieDFLNN (ours) & 128 & 3 & 600 & $2.000$ \\
        \hline
    \end{tabular}
\end{table}

\paragraph{Optimization.}
All models are trained with the SOAP optimizer~\cite{vyas2024soap} at a learning rate of $3\times10^{-3}$. Batch size, number of epochs, and data window length are reported in \Cref{tab:hyperparams-train}. The proposed model (and baseline Euclidean DFLNN) minimize the discrete forced Euler-Lagrange residual together with a regularity penalty (see \Cref{Appendix: regular Lagrangian}), combined with weights $\omega_\mathrm{DEL}=\omega_\mathrm{reg}=0.5$, and threshold exponent $\epsilon=10$ (Experiments~1, 3), $\epsilon=100$ (Experiment~2) or $\epsilon=10^{5}$ (Experiment~4). The Neural ODE has no variational structure and is trained with a pure trajectory mean-squared error. In Experiment~5, the Lagrangian is fixed, so the loss reduces
to the controlled residual alone ($\omega_\mathrm{DEL}=1$,
$\omega_\mathrm{reg}=0$) and no regularity penalty is needed.

\paragraph{Window length.}
We distinguish the \emph{trajectory length} (timesteps per data sequence) from the \emph{window length} (consecutive poses in a single training segment, i.e.\
the window backpropagated through), since the latter is the unit counted by the batch size. The proposed model and the discrete baselines (Euclidean DFLNN, GLNN
w/Lie-Midpoint) operate on pose triplets (sample length $3$), whereas the rollout-based GLNN w/RK4, GLNN w/Lie-Heun and the Neural ODE backpropagate through the full integrated trajectory. 

\paragraph{Inference.}
Forward prediction integrates the learned discrete system by solving the implicit update at each step with a damped Newton iteration (tolerance $10^{-5}$, at most $10$ iterations), as described in \Cref{subsection: inference routine}.

\subsection{Rigid body on $SO(3)$ in a gravitational field with Rayleigh dissipation}
\label{appendix: Rigid body on SO3 in a gravitational field with Rayleigh dissipation}
The continuous system describing a single rigid body evolving on $SO(3)$ (\Cref{example: L=T-V in SO3}) in the presence of an external gravitational field and frictional damping torque is
\begin{align*}
    \dot{\omega} &= J^{-1} ( \tau - \omega \times (J \omega))\\ 
    \dot{R} & = R \hat{\omega} 
\end{align*}
where $J\in\mathbb{R}^{3\times3}$ is the symmetric positive definite inertia matrix for vector representation, and $\tau = \tau_\mathrm{cons} + \tau_\mathrm{ext} $ is the sum of the conservative (gravitational) torque and the external torque.
We consider the gravitational potential
\begin{equation*}
    V:SO(3) \to \mathbb{R}, \qquad V(R) = m g e_3^\top R \rho_{COM} 
\end{equation*}
where $m>0$ is the mass and $\rho_{COM}\in \mathbb{R}^3$ is the center of mass vector in the body frame. The corresponding conservative torque is given by the negative gradient of the potential
\begin{equation*}
    \tau_\mathrm{cons} = -\grad_R V(R).
\end{equation*}
Since $\grad_R V(R)\in T_R SO(3)$ and by left trivialization $T_R SO(3) \simeq R \, \mathfrak{so}(3)$, we have
\begin{equation*}
    \tau_\mathrm{cons} = -\frac{1}{2} \vex \left(R^\top \frac{\partial V}{\partial R} - \left( R^\top \frac{\partial V}{\partial R} \right)^\top \right)
    = - mg \rho_{COM} \times (R^\top e_3).
\end{equation*}
For the external torque we can use a Rayleigh dissipative load of the form
\begin{equation*}
    \tau_\mathrm{ext} (\omega) = - D \omega, \qquad 
\end{equation*}
where $D\in\mathbb{R}^{3\times3}$ is a symmetric positive semidefinite damping matrix.

\subsection{Rigid body on $SE(3)$ in a gravitational field with Rayleigh dissipation}
\label{appendix: Rigid body on SE3 in a gravitational field with Rayleigh dissipation}
The continuous system describing a single rigid body evolving on $SE(3)$ (\Cref{example: L=T-V in SE3}) in the presence of an external gravitational field and frictional damping force is 
\begin{align*}
    \dot \omega & = J^{-1}\!\left( \tau
    - \omega \times \big(J\omega + \rho_{\mathrm{COM}} \times P\big)
    - v \times P - \rho_{\mathrm{COM}} \times \big(f - \omega \times P\big) \right) \\
    \dot v &= \frac{1}{m}\big(f - \omega \times P\big)
    - \dot{\omega} \times \rho_{\mathrm{COM}} \\
    \dot R &=R\hat{\omega} \\
    \dot p &= Rv
\end{align*}
where $P := m\big(v + \omega \times \rho_{\mathrm{COM}}\big)$, $J\in\mathbb{R}^{3\times 3}$ is the symmetric positive definite inertia matrix for vector representation, $m>0$ is the mass, $\rho_{COM}\in\mathbb{R}^3$ is the center of mass vector in the body frame, $\tau = \tau_\mathrm{cons} + \tau_\mathrm{ext}$ is the sum of the conservative (gravitational) torque and the external torque, and $f = f_\mathrm{cons} + f_\mathrm{ext}$ is the sum of the conservative (gravitational) translational force and the external translational force. 
We consider the gravitational potential
\begin{equation*}
    V:SE(3) \to \mathbb{R}, \qquad V(R,p) = mge_3^\top(p + R \rho_{\mathrm{COM}}).
\end{equation*}
The corresponding conservative torque and translational force read
\begin{equation*}
    \tau_\mathrm{cons} = - m g \rho_{\mathrm{COM}} \times (R^\top e_3 ), \qquad 
    f_\mathrm{cons} = -m g R^\top e_3.
\end{equation*}
For the external force we can use a Rayleigh dissipative load of the form
\begin{equation*}
    \begin{pmatrix}
    \tau_{\mathrm{ext}}  (\omega, v) \\
    f_{\mathrm{ext}} (\omega, v)
    \end{pmatrix}
    = - D \begin{pmatrix}
    \omega \\
    v
    \end{pmatrix}
    = - \begin{pmatrix}
    D_{\omega\omega}\omega + D_{\omega v} v \\
    D_{v\omega}\omega + D_{vv} v
    \end{pmatrix},
\end{equation*}
where $\tau_{\mathrm{ext}} \in \mathbb{R}^3$ is the rotational damping torque (e.g. rotational drag, internal friction, bearing or joint damping), $f_{\mathrm{ext}} \in \mathbb{R}^3$ is the translational damping force (e.g. air drag, fluid resistance, translational viscous friction) and $D\in\mathbb{R}^{6\times6}$ is a symmetric positive semidefinite damping matrix
\begin{equation*}
    D =
    \begin{pmatrix}
    D_{\omega\omega} & D_{\omega v}\\
    D_{v\omega} & D_{vv}
    \end{pmatrix},
    \qquad
    D_{\bullet\bullet} \in \mathbb{R}^{3\times 3}.
\end{equation*}

\subsection{2-body system}
\label{appendix: 2 body system}
Consider a 2-body system evolving on $SE(3)\times SO(3)$ as in \Cref{appendix: DEL SE3 x SO3}. For body $i$, $i=1,2$, let $J_i \in\mathbb{R}^{3\times 3}$ be the symmetric positive definite inertia matrix for vector representation, $m_i>0$ the mass, and $\rho_{{COM}_i}\in\mathbb{R}^3$ the center of mass vector in the body frame. 
The continuous equations describing the system in the presence of an external gravitational field are
\begin{align}
    \label{eq:2_body_ode_gravitational_potential omega1}
    \dot \omega_1 & = J_1^{-1}\!\left(
    \tau_1 - \omega_1 \times \big(J_1\omega_1 + \rho_{\mathrm{COM},1} \times P\big)
    - v_1 \times P - \rho_{\mathrm{COM},1} \times \big(f_1 - \omega_1\times P\big)
    \right)\\
    \label{eq:2_body_ode_gravitational_potential v1}
    \dot v_1 &= \frac{1}{m_1}\big(f_1 - \omega_1 \times P_1 \big)
    - \dot{\omega}_1 \times \rho_{\mathrm{COM},1} \\
    \label{eq:2_body_ode_gravitational_potential R1}
    \dot R_1 &=R\hat{\omega}_1 \\
    \label{eq:2_body_ode_gravitational_potential p1}
    \dot p_1 &= Rv_1 \\
    \label{eq:2_body_ode_gravitational_potential omega2}
    \dot{\omega} &= J_2^{-1} ( \tau_2 - \omega_2 \times (J_2 \omega_2))\\ 
    \label{eq:2_body_ode_gravitational_potential R2}
    \dot{R} & = R \hat{\omega}_2
\end{align}
where $P:= m_1\big(v_1 + \omega_1 \times \rho_{\mathrm{COM},1}\big)$, $\tau_i$ is the conservative (gravitational) torque acting on body $i$, and $f = f_{\mathrm{cons}} + f_{\mathrm{ext}}$ is the conservative (gravitational) translational force acting on body 1. We have
\begin{align*}
    \tau_i = - \vex \left( \mathrm{skew} \left ( R_i^\top \left( \frac{\partial V_i}{\partial R_i} + \frac{\partial V_{\mathrm{int}}}{\partial R_i} \right) \right) \right), \qquad
    f = - R_1^\top \left( \frac{\partial V_1}{\partial p_1} + \frac{\partial V_{\mathrm{int}}}{\partial p_1} \right)
\end{align*}
where $V_\mathrm{int}: SE(3) \times SO(3) \to \mathbb{R}$ is the interaction potential, and the gravitational potentials are given by
\begin{align*}
    V_1:SE(3) &\to \mathbb{R}, \qquad &V_1(R_1,p_1) &= m_1 g e_3^\top(p_1 + R_1 \rho_{\mathrm{COM},1}) \\
    V_2:SO(3) &\to \mathbb{R}, \qquad &V_2(R_2) &= m_2 ge_3^\top R_2 \rho_{\mathrm{COM},2}.
\end{align*}
The continuous Lagrangian corresponding to the system \eqref{eq:2_body_ode_gravitational_potential omega1}-\eqref{eq:2_body_ode_gravitational_potential R2} is
\begin{align*}
    L(R_1, p_1, \omega_1, v_1, R_2, \omega_2) & = \frac{1}{2} \omega_1^\top J_1 \omega_1 
    + \frac{1}{2} \omega_2^\top J_2 \omega_2 
    + \frac{1}{2} m_1 (v_1 + \hat{\omega}_1 \rho_{\mathrm{COM},1})^\top (v_1 + \hat{\omega}_1
    \rho_{\mathrm{COM},1})  \\
    &  - V_1(R_1, p_1) -  V_2(R_2) - V_{\mathrm{int}} (R_1, p_1, R_2).
\end{align*}

\subsubsection{Examples of interaction potentials}\label{appendix: examples of interaction potentials}
Here we provide some choices of explicit interaction potentials.
\begin{enumerate}
    \item \textbf{Alignment potential.} If the two bodies are connected by a spring or a flexible beam with stiffness $\mu$ at a point $r_1$ (in body 1's frame) and $r_2$ (in body 2's frame). The global position of these points are $X_1 = p_1 + R_1 r_1$ and $X_2 = p_2 + R_2 r_2$,
    hence the interaction potential is
    \begin{equation*}
        V_{\mathrm{int}}(R_{1}, p_{1}, R_{2}, p_{2}) = 
        \frac{\mu}{2} \| (p_{1} + R_{1} r_1) - (p_{2} + R_{2} r_2) \|^2.
    \end{equation*}
    If the configuration manifold is $SO(3)\times SO(3)$, we need to penalize the relative rotation between them. The interaction potential is then
    \begin{equation*}
        V_{\mathrm{int}}(R_1, R_2) = \frac{1}{2} \mu \| \log^{SO}(R_1^\top R_2) \|^2
    \end{equation*}
    or, for a computationally cheaper version
    \begin{equation*}
        V_{\mathrm{int}}(R_1, R_2) = \mu (3-\trace(R_1^\top R_2)).
    \end{equation*}
    \item \textbf{Gravitational Interaction.} If the two bodies are massive, then they exert a mutual gravitational pull on each other. The interaction potential is the standard Newtonian gravity between the centers of mass
    \begin{equation*}
        V_{\mathrm{int}}(p_1, p_2) = - g \frac{m_1 m_2}{\| (p_2 +R_2 \rho_{\mathrm{COM}, 2}) - (p_1 +R_1 \rho_{\mathrm{COM}, 1}) \|}.
    \end{equation*}
    \item \textbf{Magnetic dipole-dipole interaction.} Assume each body has an internal magnet located at positions $r_1$ and $r_2$, with $\mu_1$ and $\mu_2$ the magnetic moment vectors (in the local body frames). Let $r = X_2 - X_1 = (p2 + R_2 r_2) - (p1 + R_1 r_1) $ be the distance vector between the magnets in the global frame. The interaction potential is
    \begin{equation*}
        V_{\mathrm{int}}(R_1, p_1, R_2, p_2) = \frac{\mu_0}{4\pi\|r\|^3} \left( (R_1 \mu_1) \cdot (R_2 \mu_2) - \frac{3}{\|r\|^2} (R_1 \mu_1 \cdot r) (R_2 \mu_2 \cdot r) \right).
    \end{equation*}
\end{enumerate}

\subsection{PD control in $SO(3)$}
\label{appendix: pd control in SO3}
Consider a control-affine system in $SO(3)$ as in \cite{lee2012robust}. Let $R_{\mathrm{f}}$ be the desired attitude, and consider the PD controller
\begin{equation*}
    \tau_{PD}(R, \omega) = -K_R e_R (R, R_\text{f}) - K_\omega e_\omega (\omega) + \vex({\tau_{\mathrm{ff}}}),
\end{equation*}
where $K_R, K_\omega >0$ are gain constants, and the attitude and angular velocity error vectors are
\begin{align*}
    e_R(R, R_\mathrm{f}) &= \frac{1}{2} \vex(R_\mathrm{f}^\top R - R^\top R_\mathrm{f}), \\
    e_\omega (\omega) &= \omega.
\end{align*}
To compensate for gravitational effects, we include the feed-forward torque $\tau_{\mathrm{ff}}$ derived from the potential $V$: 
\begin{equation*}
    {\tau_{\mathrm{ff}}} = \frac{1}{2} \left( R^\top \frac{\partial V}{\partial R} - \left(\frac{\partial V}{\partial R}\right)^\top R \right).
\end{equation*}
This control low drives the orientation $R$ toward $R_\mathrm{f}$ by canceling the conservative torque and providing dissipative-restorative feedback.
One could further choose $K_R, K_\omega \in \mathbb{R}^{3\times 3}$ as symmetric positive definite matrices to tune the response along specific body axes, and/or $K_R=K_R(R)$ to be a function of $R$ with larger magnitude when $R$ is far from $R_\mathrm{f}$.

\end{document}